\documentclass{amsart}
\usepackage{fullpage}

\usepackage{amstext,amsmath,amssymb,natbib,graphicx,url}
\usepackage[colorlinks=true, linkcolor=blue, citecolor=green, urlcolor=cyan]{hyperref}

\usepackage{tikz}
\usetikzlibrary{mindmap}

\usepackage{subfigure}

\usepackage{ulem}

\usepackage[ruled]{algorithm2e}

\usepackage{tikz}
\usetikzlibrary{mindmap}

\usepackage{pgfplots}

\newtheorem{theor}{Theorem}[section]
\newtheorem*{theorem*}{Theorem}

\newtheorem{lemma}[theor]{Lemma}

\newtheorem{corollary}[theor]{Corollary}

\newtheorem{definition}[theor]{Definition}

\newtheorem{proposition}[theor]{Proposition}

\theoremstyle{definition}
\newtheorem{rem}[theor]{Remark}
\newtheorem{remark}[theor]{Remark}

\theoremstyle{plain}

\newtheorem{assumption}{Assumption}

\newcommand{\R}{\mathbb{R}}

\def\Prob{{\mathbb P}}

\usepackage{xcolor}
\definecolor{b}{HTML}{4472c4}
\definecolor{o}{HTML}{ED7D31}
\definecolor{g}{HTML}{70ad47}
\definecolor{t}{RGB}{40,154,150}

\newcommand{\sign}{\mathbf{Sign}}

\def\dist{{\rm dist}}

\def\R{{\mathbb R}}

\def\sign{{\rm sign}}

\def\Prob{{\mathbb P}}

\def\d{{\rm dist}}

\def\dist{{\rm dist}}

\def\diam{\operatorname{diam}}

\def\cal{\mathcal}
\def\tt{\mathtt}

\def\eU{{\mathcal U}}
\newcommand{\pd}{{\rm d}}
\def\rv{\mu_{\min}  }
\newcommand{\rp}{{\rm p}}
\newcommand{\rK}{{\rm K}}

\newcommand{\cdr}{C_{\rm gap}}

\newcommand{\tp}{{\tt T}}
\newcommand{\sd}{{\bf d}}
\newcommand{\gc}[1]{{#1}_{\rm gc}}

\newcommand{\TM}{H_{i_0}}
\newcommand{\rr}{{\rm r}}
\newcommand{\de}{\sd_{\rm euc}}
\newcommand{\iW}{i_{W'}}
\newcommand{\iWs}{i^\sharp }
\newcommand{\varu}{\varpi}

\newcommand{\inx}{i_{\rm nx}}

\newcommand{\bnet}{\hyperref[alg:buildNet]{\texttt{\rm buildNet}}}
\newcommand{\gclu}{\hyperref[alg:cluster]{\texttt{\rm generateCluster}}}

\newcommand{\nc}{\hyperref[alg:net-check]{\texttt{\rm netCheck}}}
\newcommand{\wg}{\hyperref[alg:weightedGraph]{\texttt{\rm weightedGraph}}}

\newcommand{\nocolorref}[2]{\hypersetup{linkcolor=black}\hyperref[#1]{#2}\hypersetup{linkcolor=blue}}
\newcommand{\Eclu}{\nocolorref{defi clu}{{\mathcal E}_{\rm clu}}}
\newcommand{\Enavi}{\nocolorref{eq: E-navi}{\mathcal{E}_{\rm navi}}}
\newcommand{\Enet}{\nocolorref{eq:Enet}{\mathcal{E}_{\rm net}}}
\newcommand{\Ecn}{\nocolorref{eq: Ecommonneighbor}{\mathcal{E}_{\rm cn}}}

\newcommand{\Eao}{\nocolorref{eq: Eao}{{\cal E}_{\rm ortho}}}
\newcommand{\Enx}{\nocolorref{eq: Enx}{{\cal E}_{\rm next}}}
\newcommand{\Enc}{\nocolorref{eq: Enc}{{\cal E}_{\rm netCheck}}}

\newcommand{\defn}[1]{{\bf \textcolor{blue}{#1}}}

\newcommand{\bbR}{\mathbb{R}}
\newcommand{\calM}{\mathcal{M}}

\newcommand{\vol}{\operatorname{vol}}

\definecolor{color2}{RGB}{118,128,97}

\definecolor{pumpkin}{RGB}{255,117,24}

\begin{document}

\title{Reconstructing the Geometry of Random Geometric Graphs}
\author{Han Huang, Pakawut Jiradilok, and Elchanan Mossel}
\date{\today}

\address{Department of Mathematics, University of Missouri, Columbia, MO 65203}
\email[H.~Huang]{hhuang@missouri.edu}

\address{Department of Mathematics, Massachusetts Institute of Technology, Cambridge, MA 02139}
\email[P.~Jiradilok]{pakawut@mit.edu}

\address{Department of Mathematics, Massachusetts Institute of Technology, Cambridge, MA 02139}
\email[E.~Mossel]{elmos@mit.edu}

\begin{abstract}
Random geometric graphs are random graph models defined on measure metric spaces. Such a  model is defined by first sampling points from a metric space and then connecting each pair of sampled points with probability that depends on their distance, independently among pairs.

In this work we show how to efficiently reconstruct the geometry of the underlying space from the sampled graph under the {\em manifold} assumption, i.e., assuming that the underlying space is a low dimensional manifold and that the connection probability is a strictly decreasing function of the Euclidean distance between the points in a given embedding of the manifold in $\mathbb{R}^N$.

\end{abstract}

\maketitle
\thispagestyle{empty}

\section{Introduction}

Random graphs provide a versatile language for modeling large networks. The Erdős–Rényi model \(G(n,p)\) is a foundational example in which edges appear independently with probability \(p\). Its analytic clarity has shaped much of modern probabilistic combinatorics and continues to offer valuable benchmarks for phenomena such as connectivity, typical distances, and the emergence of large components. We refer to the classical monographs~\cite{bollobas2011random,janson2011random} for a comprehensive overview of the subject; see also \cite{frieze2015introduction} for a more recent treatment. In many applications, however, one expects interactions to be modulated by spatial or geometric considerations. In such settings, it is natural to augment the probabilistic framework with a latent metric structure.

Random geometric graphs (RGGs) formalize the idea of geometry-driven connectivity by assigning latent positions to vertices and letting distance modulate adjacency. In the classical \emph{hard-disc} model, two vertices $v$ and $w$ are connected whenever ${\rm dist}(X_v, X_w) \le r$, where $X_v$ and $X_w$ denote their latent positions. The \emph{soft} (or random-connection) model generalizes this by introducing a monotone connection function $\rho({\rm dist}(X_v, X_w))$, so that edges appear independently with probability $\rho({\rm dist}(X_v, X_w))$, reflecting fading or noisy interactions typical of wireless and other spatial systems~\cite{Pen03,dettmann2016random}.

In the hard-disc model, it was recently studied whether one can distinguish a spherical RGG from an Erd\H{o}s--R\'enyi graph with the same edge density. In the dense regime, it was first shown that a sharp detectability threshold occurs at latent dimension $d \asymp n^3$~\cite{BDEM16}. Subsequent work extended these results to the sparse regime $p = c/n$, where indistinguishability now holds down to polylogarithmic dimensions (up to polynomial factors), nearly confirming the conjectures of~\cite{BDEM16}~\cite{LMSY22,BBN20}. Another line of research studies nonparametric inference for translation-invariant RGGs (TIRGGs) on $S^{d-1}$, in which edges are generated according to an unknown function $p(\langle x, y\rangle)$~\cite{ADC19,EMP22,DDC23}. The goal in this setting is to recover the $p$ from a single observed graph. We refer to the survey~\cite{DDC23} for an overview of recent progress in this area.

This paper takes a different viewpoint, where we assume the underlying metric space is unknown. Suppose latent points are sampled from a probability measure \(\mu\) supported on a \(d\)-dimensional submanifold \(M\subset\mathbb{R}^N\), and edges are sampled independently with probability \(\rp(\|X_v-X_w\|)\) for a strictly decreasing \(\rp\).
We ask the following question:
\medskip

\noindent \textbf{Question A.} \, {\em To what extent can one infer geometric information about \((M,\mu)\) from a single realization of the graph?}

\noindent \textbf{Question B.} \, {\em If the estimation is possible, is there an efficient algorithm to do so?}

\medskip
To the best of our knowledge, these questions have not been studied in the literature.
\subsection{Our Results}
We answer both Questions A and B affirmatively. Our results show that, under natural regularity assumptions on \(M\), \(\mu\), and \(\rp\), it is possible to recover both intrinsic (geodesic) and extrinsic (ambient Euclidean) distances between the sampled points up to a vanishing error, and to approximate the underlying metric–measure structure.

We now formalize the model \(G(n,M,\mu,\rp)\) and present our main theorems.
We consider the case where $M$ is a smooth compact connected $d$-dimensional manifold embedded in a Euclidean space.
Our results will depend on a bound on the second fundamental form of $M$,
which is a measure of how curved the manifold is.
They will also depend on a {\em repulsion property},
informally requiring that if $p,q \in M$ are sufficiently close in the Euclidean distance,
then they are also close in the geodesic distance.
As for $\mu$, we assume a lower bound of order $r^d$ on the probability measure of any ball of radius $r$ for $r$ close to $0$. Finally, we assume that $\rp$ is a strictly monotone decreasing continuous function,
with some quantitative bound on the Lipschitzness of $\rp$: for any $0 \le a, b \le \sup_{p,q \in M} \|p-q\|$, we have
\[
\ell_{\rp} |a-b| < |\rp(b)-\rp(a)| < L_{\rp}|a-b|,
\]
for some positive constants $\ell_\rp \le L_\rp$.

See Assumptions~\ref{assump: M-and-pm} and \ref{assump: pdf} for the exact formulation.

In our main results we prove that the geometry can be recovered from the sampled graph. There are two different notions of geometric recovery stated in Theorems~\ref{thm: geodInformal} and \ref{thm: geodGHInformal} below, where
$\sd_{\rm gd}$ denotes the geodesic distance.

\begin{theor}  \label{thm: geodInformal}
There exists $c>0$ so that the following holds.
Suppose $(M,\mu,\rp)$ satisfies Assumptions~\ref{assump: M-and-pm} and \ref{assump: pdf}. Then there exists a deterministic algorithm that takes $G=G(n,M,\mu,\rp)$ as input, with running time $O(n^3)$ outputs a weighted graph $\Gamma$ on $[n]$ and a metric $\de$ on $[n]$, such that, with probability $1-n^{-\omega(1)}$,
\begin{itemize}
\item for every $p \in M$, there exists $v$ such that
$$
    \|X_v-p\| \le \sd_{\rm gd}(X_v,p) \le C(M,\mu,\rp)n^{-c/d}.
$$
\item
    for every pair of vertices $v,w$,
    $$
        \left|\sd_{\rm gd}(X_v,X_w) - \sd_{\Gamma}(v,w) \right|
    \le
        C(M,\mu,\rp)n^{-c/d},
    $$
    where $\sd_{\rm gd}$ denotes the geodesic distance on $M$, and $\sd_{\Gamma}$ denotes
    the path metric on $\Gamma$.
\item
for every pair of vertices $v,w$,
    $$
        \big| \| X_v - X_w \| - \de(v,w) \big|
    \le
        C(M,\mu,\rp)n^{-c/d},
    $$
where $\| \cdot \|$ denotes the Euclidean norm.
\end{itemize}
Here, $C(M,\mu,\rp) \ge 1$ is a constant which depends on $M$, $\mu$, and $\rp$. \end{theor}

Informally this theorem says that we can recover both the intrinsic and the extrinsic distances of the sampled points on the manifold.

\begin{remark}
    An extended abstract of this work is accepted for presentation at the Conference on Learning Theory (COLT) 2024. \cite{HJM:24}
\end{remark}
\begin{remark}[On the necessity of $\rp$ beyond step functions]
    \label{rem:hard}
    In many studies of random geometric graphs, the function $\rp$ is typically chosen as a step function, while the underlying manifolds are usually restricted to specific classes, such as spheres $S^{m-1}$ with unknown dimension $m$. In comparison, our $\rp$ appears to encode more information. Regarding the manifold reconstruction problem, we provide examples demonstrating that step functions are \emph{insufficient} for reconstruction in general. Specifically, we exhibit examples of two random geometric graphs generated from the same $\rp$ but arising from distinct manifolds, yet the two graphs have identical distributions, see Proposition \ref{prop:two-helices-m-n}: In short, the manifolds we consider is a helix wrapping around a flat torus embedded in $\mathbb{R}^4$, i.e., $S^1\times S^1 \subseteq \mathbb{R}^4$ equipping with the uniform measure, for any step function $\rp = p {\bf 1}_{[0,r]}$, there exists another such helix with the uniform measure such that they differ both in terms of the extrinsic and intrinsic distance, yet the resulting random graphs are identically distributed, showing that the step function alone does not distinguish the underlying manifold structure.
\end{remark}

\noindent
\underline{Connection to Manifold Learning.}
The manifold assumption in machine learning is a popular assumption postulating that many models of data arise from distributions over manifolds, see e.g.~\cite{DDC23,ADC19,FILLN21,FILN20,FILN23,ADL23} among many others. A major problem studied in this  is the inference problem of estimating an unknown manifold given data sampled from the manifold. Some of the foundational work in this area shows that in many situations given the (possibly noisy) distances between sampled points it is possible to estimate the unknown manifold.

In our setup, points are again sampled from a manifold but now the data is a graph. In the graph each sampled point corresponds to a vertex and an edge is included with probability that depends in a strictly monotone fashion on the (embedded) distance of the corresponding endpoints. Reconstructing a manifold from a graph is a more difficult question than manifold learning from exact distances between the sampled points as for every pair of point we only have access to a binary variable whose sampling probability depends on the distance rather to the distance itself.

Combining our result with the main results of~\cite{FIKLN20,FILLN21},
it is possible to recover a manifold that is close to the one that the data is sampled from.
\begin{corollary}\label{cor:get-manifold}
Assume the same settings as in Theorem~\ref{thm: geodInformal}. There exist an absolute constant $c > 0$ and a deterministic polynomial time algorithm that takes $G=G(n,M,\mu,\rp)$ as input, and outputs, with probability $1 - n^{-\omega(1)}$, a smooth Riemannian manifold $(\widehat{M}, \widehat{g})$ which is diffeomorphic to $(M,g)$, together with a diffeomorphism
\[
F: \widehat{M} \to M
\]
such that, for any $x, y \in \widehat{M}$,
\[
\frac{1}{L} \le \frac{\sd_{{\rm gd}_M}(F(x), F(y))}{\sd_{{\rm gd}_{\widehat{M}}}(x,y)} \le L,
\]
where $L = 1 + C(M,\mu,\rp) \cdot n^{-c/d}$, and where $C(M,\mu,\rp) \ge 1$ is a constant which depends on $M$, $\mu$, and $\rp$.
\end{corollary}
\begin{proof}
This corollary essentially follows from~\cite[Thm.~1.2~(2)]{FILLN21}. To apply this result, we need to check that the manifold in our setting satisfies the required conditions. The bounds on the diameter, the injectivity radius, and the sectional curvature are given under Assumptions~\ref{assump: M-and-pm} and~\ref{assump: pdf}. Theorem~\ref{thm: geodInformal} implies that we obtain a finite $\varepsilon_0$-net with $\varepsilon_0 \ll_{M,\mu,\rp} n^{-c/d}$ on the whole manifold $M$, and we also obtain {\em distance vector data} (see \cite{FILLN21}) with distance noise upper bound $\varepsilon_1 \ll_{M,\mu,\rp} n^{-c/d}$ between any two points of the net. Thus, \cite{FILLN21} gives an algorithmic reconstruction of the desired smooth Riemannian manifold together with the desired diffeomorphism.
\end{proof}

Our second main result involves recovering the manifold $M$ together with the measure $\mu$ in a version of Gromov--Hausdorff distance for metric measure spaces.
For more background on metric measure spaces and Gromov--Hausdorff distances, see e.g.~\cite{Gro99,Shi16}.
\begin{theor}\label{thm: geodGHInformal}
Suppose $(M,\mu,\rp)$ satisfies Assumptions \ref{assump: M-and-pm} and \ref{assump: pdf}. Then there exists $C = C(M,\mu,\rp) \ge 1$
    so that the following holds. Consider the graph $G=G(n,M,\mu,\rp)$. There exists a deterministic algorithm that takes $G$ as input. With probability $1-n^{-\omega(1)}$, it returns  $(\widetilde{\Gamma}, \nu,\de)$ with running time $O(n^3)$ such that
    \begin{itemize}
        \item $\widetilde{\Gamma}$ is a weighted graph whose vertex set is a subset $V'$ of the vertices of $G$,
        \item $\nu$ is a probability measure on $V'$,
        \item $\de$ defines a metric space on $V'$,
        \item there exists a coupling $\pi$ of $\mu$ and $\nu$ so that,
         for two independent copies $(X,u),(X',u') \sim \pi$,
    \begin{align*}
    \mathbb{P}\big\{
            \big|\sd_{\rm gd}(X,X') - \sd_{\widetilde{\Gamma}}(u,u')\big| > Cn^{-c/d}
        \big\}
    \le
        C n^{-1/4},
    \end{align*}
    and
        \begin{align*}
    \mathbb{P}\big\{
            \big|\| X - X' \| - \de(u,u')\big| > Cn^{-c/d}
        \big\}
    \le
        C n^{-1/4},
    \end{align*}
        \item for any vertex $u$ of $\widetilde{\Gamma}$ and $t \ge 0$,
    \begin{align*}
        \mu(B_{\rm gd}(X_u,t - C&n^{-c/d})) -
        C n^{-1/4} \\
    &\le  \nu(B_{\widetilde{\Gamma}}(u,t)) \\
    &\le
        \mu(B_{\rm gd}(X_u,t + Cn^{-c/d})) +
        C n^{-1/4},
    \end{align*}

    and

        \begin{align*}
        \mu(B_{\| \cdot \|}(X_u,t - C&n^{-c/d})) -
        C n^{-1/4} \\
    &\le  \nu(B_{\de}(u,t)) \\
    &\le
        \mu(B_{\| \cdot \|}(X_u,t + Cn^{-c/d})) +
        C n^{-1/4},
    \end{align*}

    where $B_{\rm gd}(x,t)$ (resp., $B_{\| \cdot \|}(x,t)$, $B_{\widetilde{\Gamma}}(x,t)$, and $B_{\sd_{\rm euc}}(x,t)$) denotes the open ball of radius $t$ around $x$ in geodesic distance (resp., Euclidean distance in the ambient space $M$ is embedded in, path metric on $\widetilde{\Gamma}$, and the metric $\sd_{\rm euc}$ on $\widetilde{\Gamma}$).
    \end{itemize}
\end{theor}

The theorem shows that $\widetilde{\Gamma}$ is a good approximation to $M$ as a metric measure space.
The coupling $\pi$  ``matches" the two spaces. Under this matching distances match:
the intrinsic distance on the original manifold matches the graph distance on $\widetilde{\Gamma}$ and the embedded distance on the original manifold matches the distance $\sd_{\rm euc}$.

The proofs of both theorems essentially use the following result which is the main technical result
of the paper. It shows how to extract  a ``net of clusters"  from the random geometric graph.

\begin{theor}\label{thm: NetInformal}
Suppose that $(M,\mu,\rp)$ satisfies Assumptions~\ref{assump: M-and-pm} and \ref{assump: pdf}.
There exist constants $C=C(M,\mu,\rp)$,\, $c\in (0,1)$,\, and  an  algorithm~\bnet, which takes $G = G(n,M,\mu,\rp)$ as input so that the following holds.
Let
$$
    \eta = C n^{-c/d}\,.
$$
With probability $1-n^{-\omega(1)}$ on $G$,
the output of \bnet~ is a collection of pairs $\{ (U_\alpha, u_\alpha) \}_{\alpha \in [\ell]}$ with $u_\alpha \in U_\alpha \subseteq [n]$ which form a {\em Cluster-Net} in the following sense.

The output of the algorithm satisfies
    \begin{enumerate}
        \item For each $\alpha \in [\ell]$, $|U_\alpha| \ge n^{1/2}$ and for each $v \in U_\alpha$, $\|X_{u_\alpha} - X_{v}\| \le \eta$.
        \item For each $p \in M$, there exists $\alpha \in [\ell]$ such that $\|p - X_{u_\alpha}\| \le 1000\sqrt{d}\eta$.
    \end{enumerate}
    Further, the running time of the algorithm is $O(n^3)$.
\end{theor}

\subsection{Related Work}
Our work connects to two large bodies of works:  Random Geometric Graphs and Manifold Learning.

For surveys on Random Geometric Graphs, see e.g.~\cite{Pen03,DDC23}.

Perhaps the closest to our work is the work of \cite{ADC19}. They consider random geometric graphs generated from latent points on the sphere $\mathbb{S}^{d-1}$ with the probability of having an edge between two latent points given by the value of the probability ``link'' function evaluated at the distance between the two points.
Similar to our result, their work gives a way to approximate the distances between the latent points from the random geometric graph. The setup of \cite{ADC19} crucially relies on the assumption of uniform sampling from the unit sphere. This allows them to use spectral/harmonic analysis on the sphere to obtain fast algorithms and good rates of convergence.
This also allows them to reconstruct graphs in the sparse regime, which we do not.
See also the follow up work~\cite{EMP22}, and see also~\cite{STP12} for related work in a similar setup.
Since we are considering general manifolds, such techniques cannot be applied.
This may also explain why our algorithms for the general case are less efficient and have worse error rates.

Works of Fefferman, Ivanov, Kurylev, Lassas, Lu, and Narayanan
(some with/without Kurylev and some with/without Lu)~\cite{FIKLN20,FILN20,FILLN21,FILN23} consider the problem of manifold learning as part of a more general manifold extension problem.
The main difference in perspective is that the interests in these lines of work are finding a bona fide manifold and moreover, points are given with (approximate) distances between them.
As mentioned earlier we can leverage this line of work together with our results to recover manifolds in our setting as well. (See Corollary~\ref{cor:get-manifold} above.)
Another line of research related to manifold learning is due to Aizenbud and Sober~\cite{JMLR:v26:25-0183, aizenbud2021non}, who consider noisy observations in a tubular neighborhood of a smooth manifold and develop methods that, for a given query point, estimate its projection onto the manifold together with the local tangent space, with provable consistency and rates.

There are many other problems that have been studied in random geometric graphs, including those of finding the location of points given noisy distances~\cite{OMK10,JM13}, questions relating to testing geometry vs.\ non-geometric random graphs, especially in high dimensions e.g.~\cite{BDEM16,LMSY22,ADL23}, and questions related to geometric block models, e.g.~\cite{LS23}.

\subsection{Practical inspiration for the Model}

We now present an illustrative example regarding tracking devices that may provide inspiration for
the type of questions studied in this paper.

Some tracking devices, such as Apple’s AirTag$^{\text{\textregistered}}$,\footnote{AirTag$^{\text{\textregistered}}$ is a registered trademark of Apple Inc.} do not have an internal GPS navigation system. Instead, their location is determined through Bluetooth communication with nearby Apple devices. This enables the owner to track the AirTag$^{\text{\textregistered}}$’s position by leveraging the network of connected devices. Now, consider a network of devices where any two local devices can establish a Bluetooth connection with a probability that depends on their distance. A natural way to model such a network is through a Random Geometric Graph. Interpreted from this context, our question can be phrased as:

\begin{center}
    \it
Given the connection network of a distributed collection of Bluetooth devices, can we reconstruct the shape of the underlying space (without GPS) and the density of the devices?
\end{center}

Of course, our model is far from modeling all aspects of this particular example.
For one, the network does have some GPS information for some of the points in the networks.
Also, in practice, the probability of connection between two devices depends not only on their distance but also on various factors such as transmission power, physical obstacles, and environmental conditions. For instance, higher transmission power can extend the effective range of Bluetooth signals, while obstacles like walls and furniture can attenuate signals and reduce connectivity. It is not clear if the probability of connection can be modeled in terms of probability kernel on some manifold. Instead,
our model represents an idealized situation where connection probability is purely determined by distance.\\
\subsection{Proof Sketch of Theorem~\ref{thm: NetInformal}}
\label{sec: ProofIdea}
\begin{figure}[ht]
    \centering
    \includegraphics[width=0.5\textwidth]{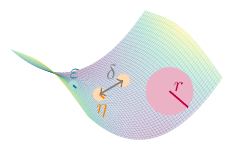}
    \caption{Illustration of parameters \(\varepsilon < \eta < \delta < r\).}
    \label{fig:parameter}
\end{figure}

We now outline the key ideas behind the algorithm \(\bnet\) and the proof of Theorem~\ref{thm: NetInformal}.
To facilitate our discussion, we introduce a sequence of increasing radii:
\[
    \varepsilon < \eta < \delta < r
\]
(see Figure~\ref{fig:parameter} for illustration), where $\eta$ is the parameter in Theorem~\ref{thm: NetInformal}.

Instead of giving the explicit definition of \(\varepsilon, \eta, \delta, r\), we first discuss their intended geometric roles. First, we assume the latent points in \(M\) are sufficiently dense, ensuring that every ball of radius \(\varepsilon\) contains many such points. Using the algorithm \(\bnet\), we form clusters of latent points, each fitting within a ball of radius at most \(\eta\). The algorithm seeks to form a net of clusters such that each point in \(M\) lies within latent Euclidean distance at most \(\delta\) of at least one cluster, and pair of clusters are separated by at least \(c\delta\) for some small constant \(c>0\). Finally, each ball of radius \(r\) within \(M\) is locally well approximated by a \(d\)-dimensional affine plane.

\smallskip

\noindent
\textit{Clusters reveal distance.}
Let us begin by explaining why we want clusters of points. Given two vertices \(u, v\), the presence or absence of an edge between them is indeed only weakly associated with the distance between \(X_u\) and \(X_v\).
However, suppose there exists a cluster \(U\) surrounding a vertex \(u\)—a pair we denote by \((U,u)\) for convenience of later reference—meaning that for every \(w\) in \(U\),
\[
    \|X_w - X_u\| \le \eta,
\]
for a small radius \(\eta\).
In this scenario, the number of edges between vertices in \(U\) and another vertex \(v\) can reliably reveal the latent Euclidean distance between $v$ and to the cluster, due to concentration properties of sums of independent Bernoulli random variables. Therefore, the presence of a cluster allows us to estimate its latent Euclidean distance to every other vertex in the graph.

\medskip

\noindent
\textit{Generating Clusters via Maximal Common Neighbors.}
The question of how to generate a cluster at a given vertex \(v\) is not trivial.
Intuitively, vertices with small latent Euclidean distance to \(v\) should have many common neighbors with \(v\).
Thus, one might think of simply grouping vertices based on the size of their common neighborhood with $v$. To make this intuition precise, consider the \defn{common neighbor probability}:

\begin{definition}\label{def:K}
    For $x,y \in M$, we define the \defn{common neighbor probability} of $x$ and $y$ as
    \[
    \rK (x,y) := \mathbb{E}_{Z\sim \mu} {\rm p}(\|x-Z\|){\rm p}(\|y-Z\|).
    \]
\end{definition}
For three different vertices $v,w,u \in {\bf V}$, conditioned on $X_v = x_v$ and $X_w = x_w$, the probability of $u \in N(v)\cap N(w)$ is $\rK(x_v,x_w)$.
This explains the name of $\rK(x,y)$ above. Specifically, when the number of vertices $n$ is large, $N(v) \cap N(w)$ is concentrated around $(n-2)\rK(X_v,X_w)$.

Returning to the discussion above, our intuition might be stated as
\begin{align*}
    \forall x \in M, \quad {\rm argmax}_{y\in M} \rK(x,y) = x.
\end{align*}

Unfortunately, this claim does {\em not} hold in general: counterexamples arise, for instance, due to curvature effects in $M$ or specific distributions $\mu$ that skew neighbor distributions, causing some points $y \neq x$ to have a larger $\rK(x,y)$ than $\rK(x,x)$.

Nevertheless, the underlying idea is not entirely wrong and it can still be exploited.

Indeed, by the Cauchy--Schwarz inequality:
\begin{align}
\label{eq: KmaxIntro2}
    \rK(x,y) \le \sqrt{\rK(x,x)\rK(y,y)} \le \max\{\rK(x,x),\rK(y,y)\},
\end{align}
we see that the global maximum $\rK(x,y)$ must occur along the diagonal \(x = y\). Interpreting this at the graph level, the pair $(v,w)$ among all pairs in the graph with the largest common neighborhood is likely to have very close latent Euclidean distance.

Following this intuition, we introduce the algorithm \gclu~which generates clusters by identifying vertex pairs that share a maximal number of common neighbors.

\smallskip

\noindent
\textit{Encoding Geometric Information into Algorithm \gclu.}
Algorithm \gclu~extracts clusters effectively but does not inherently provide control over the latent geometric location of the resulting clusters. To address this limitation, we notice that the Cauchy–Schwarz-based inequality \eqref{eq: KmaxIntro2} holds for any pair $(X_v,X_w)$, implying that it is not necessary to consider all vertices simultaneously.

Specifically, suppose we select a subset of vertices $W \subseteq {\bf V}$ satisfying the following condition:
\begin{align}\label{eq: introWcondition}
&\mbox{\it each vertex $w \in W$ has many other vertices in $W$ situated} \\
&\mbox{\it within a small latent Euclidean distance (of order $\eta$).} \notag
\end{align}
Under this condition, the critical approximation restricted to $W$ holds:
\[
    \max_{w_1,w_2 \in W} \rK(X_{w_1},X_{w_2}) \simeq \max_{w_1 \in W} \rK(X_{w_1},X_{w_1})\,.
\]
This approximation implies that identifying a vertex pair $(v,w)$ with maximal common neighbors within $W$ will likely yield vertices whose latent Euclidean distance is small. Thus, by restricting Algorithm \gclu~to the subset $W$, we extract a cluster $(V,v)$ with $V \subseteq W$. Therefore, a careful choice of $W$ allows us to effectively ``encode'' geometric information into the input of Algorithm \gclu, granting some control over the latent positioning of the resulting cluster.

To illustrate how geometric information can be encoded into the choice of the vertex subset $W$, consider the following scenario. Suppose we already have a cluster $(U,u)$ and select two radii $0 < a < b$. Define $W$ as the set containing all vertices $w$ satisfying the condition:
\[
    \rp^{-1}(b) \le \frac{|N(w) \cap U|}{|U|} \le \rp^{-1}(a)\,.
\]

Since, for large $n$, we have the approximation
\[
    \frac{|N(w) \cap U|}{|U|} \approx \rp(\|X_w - X_u\|),
\]
it follows that each vertex $w \in W$ likely satisfies
\[
    a \lesssim \|X_w - X_u\| \lesssim b.
\]

If $W$ satisfies the condition \eqref{eq: introWcondition}, then by running Algorithm \gclu~restricted to this set $W$, we extract a new cluster whose latent Euclidean distance from $X_u$ is controlled, specifically falling within the range $[a,b]$. Thus, the subset $W$ explicitly encodes distance constraints relative to the existing cluster $(U,u)$.

\medskip

\begin{figure}[htbp]
    \centering
    \includegraphics[width=0.6\textwidth]{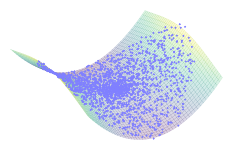}
    \caption{If $W$ is a vertex set corresponding to latent points in $M$ such that for every vertex $w \in W$ there exist many vertices $w’ \in W$ within a small distance from $w$, then the algorithm \gclu~can extract a cluster $(V,v)$ from $W$. We emphasize that the set of latent points does not need to be spread across the entire manifold.}
    \label{fig:gclu}
\end{figure}

\begin{figure}[ht]
    \centering
    \includegraphics[width=0.5\textwidth]{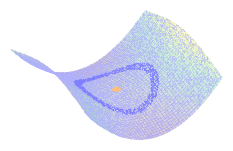}
    \caption{When a cluster of vertices $(U,u)$ is found, we can identify the set of vertices $W$ whose latent Euclidean distance from $(U,u)$ is say between $a$ and $b$. Then running Algorithm \gclu will extract a new cluster $U'$ whose latent Euclidean distance from $U$ is between $a$ and $b$.}
\end{figure}

\medskip

\noindent
\textit{Finding a Cluster Near a Target Vertex} Now we are ready to illustrate how to locate a new cluster near a target vertex $v$.
Suppose we have identified a vertex $v$ at a latent Euclidean distance approximately $\delta$ from an existing cluster $(U,u)$. We now describe how to locate a new cluster $(V',v')$ near the target vertex $v$. This construction relies on the fact that locally the manifold behaves approximately as a flat $d$-dimensional space.

The construction proceeds as follows. First, we apply Algorithm \gclu~to a set $W_1$, defined by vertices whose latent Euclidean distances to $U$ are approximately $r$. This yields a new cluster $(U_1,u_1)$ at distance roughly $r$ from $(U,u)$. Next, we repeat the process: for each subsequent cluster $(U_i,u_i)$ (with $2 \leq i \leq d$), we apply \gclu~to a vertex set $W_i$ constrained so that vertices within it have latent Euclidean distances approximately equal to $r$ from $(U,u)$, but approximately equal to $\sqrt{2}\,r$ from all previously chosen clusters $(U_j,u_j)$ for $j < i$.

\begin{figure}[ht]
    \centering
    \includegraphics[width=\textwidth]{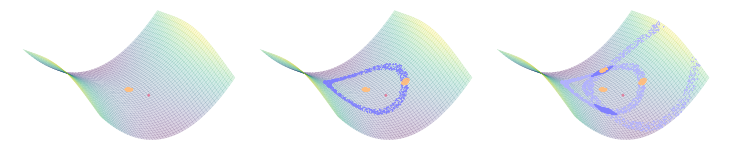}
    \caption{Suppose we have a cluster $(U,u)$ and $v$ is a vertex whose latent Euclidean distance is not far from $U$. Our goal is to find a cluster around $v$. The direct application of \gclu~may not work. Here we rely on the fact that $v$ is not far from $U$, and locally metric space is close enough to a flat plane. Instead, we can find a set of vertices $W_1$ whose latent Euclidean distance from $U$ is roughly $r$. Then, we can apply \gclu~to find a cluster $(U_1,u_1)$ whose latent Euclidean distance from $U$ is roughly $r$. We can repeat this process to find a set of clusters $U_1, U_2, \ldots, U_d$ whose latent Euclidean distances from $U$ are roughly $r$ and the vectors $\{X_{u_i} - X_u\}_{i \in [d]}$ form an approximate orthogonal basis of vectors locally for $M$ at $X_u$.}
    \label{fig:proof_idea_orthogonal}
\end{figure}

The choice of the distance $\sqrt{2} r$ ensures that $(X_{u_i}-X_u)_{i \in [d]}$ form a ``roughly orthogonal'' frame locally. Indeed, when points $X_{u_1}, \dots, X_{u_d}$ each have distance approximately $r$ from $X_u$ and pairwise distances approximately $\sqrt{2} r$ from each other, the resulting configuration is nearly orthonormal. (See Figure~\ref{fig:proof_idea_orthogonal}.)

\begin{figure}[ht]
    \centering
    \includegraphics[width=\textwidth]{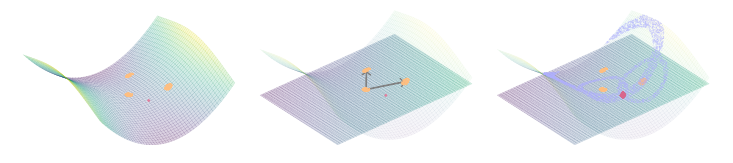}
    \caption{Given the fact that it is locally looks like a plane.
    Then the distance from $v$ to each cluster $U,V_1,\dots, V_d$ should pinpoint the location of $v$ in the manifold. This allow us to extract a vertex set $W$ whose latent Euclidean distance from $v$ is of order $\eta$. Then, running Algorithm \gclu~will extract a new cluster $U'$ whose latent Euclidean distance from $v$ is of order $\eta$.}
    \label{fig:proof_idea_next_cluster}
\end{figure}

Finally, the latent position of the target vertex $v$ can be pinpointed using the set of clusters we've constructed. Specifically, we apply Algorithm \gclu~once more, this time restricted to a carefully chosen vertex subset $W_{d+1}$. Each vertex $w \in W_{d+1}$ is selected so that its latent Euclidean distances to all previously formed clusters $(U_1,u_1),\dots,(U_d,u_d)$ closely match the corresponding latent Euclidean distances from the target vertex $v$ to these clusters. Thus, the resulting cluster $(V',v')$ obtained from $W_{d+1}$ will have its center located near the latent position of the target vertex $v$, effectively pinpointing its geometric location within the manifold. (See Figure~\ref{fig:proof_idea_next_cluster}.)
Overall, the algorithm \bnet~ keeps applies Algorithm \gclu~to carefully chosen subsets of vertices, and each time it extracts a cluster whose latent Euclidean distance not far from one of the previous cluster. Gradually, it constructs a net of clusters that covers the manifold.

\smallskip

\noindent
\textit{Technical Challenges: Quantitative justification}
The core intuition behind extracting clusters via maximal common neighbors is valid even without a manifold assumption. However, providing quantitative guarantees at multiple iterative steps of the algorithm crucially relies on the manifold’s locally Euclidean structure. Without careful quantification, the cumulative error after polynomially many iterations could become prohibitively large—potentially growing super-polynomially in  $n$.

Specifically, to form a new cluster at each step, we apply Algorithm \gclu~to subsets $W\subseteq{\bf V}$ defined by certain “fuzzy” geometric constraints—for example, vertices whose distances to a known cluster lie within some range $(a,b)$. Although the geometric constraints are stated approximately (hence, “fuzzy”), the selected set $W$ must satisfy a quantitative version of \eqref{eq: introWcondition} for the cluster extraction process to succeed quantitatively. If the range of $(a,b)$ is not within the local regime, verifying the condition for $W$ becomes challenging or even infeasible.

Hence, a substantial portion of the technical analysis is dedicated to rigorously verifying that each candidate subset W, derived from approximate latent Euclidean distance constraints, indeed meets a precise, quantitative version of \eqref{eq: introWcondition}. A typical challenging scenario arises when clusters are positioned to form an approximate orthogonal basis in latent space. When we select $W$ based on simultaneous distance constraints to these clusters, the shape of the resulting subset resembles a high-dimensional “polytope” with fuzzy boundaries. Vertices lying near the “corners” of this polytope are particularly delicate cases: we must show rigorously that even these vertices still have many nearby neighbors within W. Doing so requires leveraging both the local flatness of the manifold and additional regularity conditions on the map
\[
x \mapsto \frac{1}{|U|}\sum_{u{\prime}\in U} \rp(\|X_{u{\prime}} - x\|)
\]
for a given cluster $(U,u)$.

Thus, while the fundamental geometric intuition of the algorithm is straightforward, the primary technical challenge lies in carefully bounding the propagation and accumulation of errors at every step, thereby ensuring that each “fuzzy” vertex set $W$ satisfies the necessary geometric conditions for successful cluster extraction.

\subsection{Structure of the paper}
The rest of the paper is structured as follows.
\begin{enumerate}
    \item In Section~\ref{sec: RGM}, we introduce the random graph model and the notations used throughout the paper, including definitions of parameters and standard events.

    \item In Section~\ref{sec: Kxy}, we introduce the function $\rK(x,y)$ and discuss its properties.
    \item In Section~\ref{sec: good-pairs}, we introduce the Algorithm \gclu\, and discuss necessary prerequisites for the algorithm to successfully identify a valid cluster.
    \item In Section~\ref{sec: navi}, we discuss some regularity property of the function $M \ni p \mapsto \sum_{u \in V} \rp(\|p-X_v\|)$ for a given cluster $(V,v)$.
    \item In Sections~\ref{sec: orthogonalClusters} and \ref{sec: next-cluster}, we present two technical parts of the proof: we prove that under suitable constraints, one can produce new cluster with suitable geometric constraints.
    \item In Section~\ref{sec: netCheck}, we introduce a simple algorithm to either verify a given set of clusters form a cluster net or return a vertex which will be used as a pivot to build the next cluster.
    \item In Section~\ref{sec:proof-main}, we outline the main algorithm \bnet\, and prove Theorem \ref{thm: NetInformal}.
    \item In Section~\ref{sec: net-to-GH}, we prove Theorems~\ref{thm: geodInformal} and~\ref{thm: geodGHInformal}, based on Theorem \ref{thm: NetInformal}.
    \item In Appendix~\ref{sec: ERM}, we provide a necessary background on differential geometry.
    \item In Appendix \ref{appx: basic result proof}, we include proofs of auxiliary lemmas used throughout the paper.
    \item In Appendix~\ref{sec: GO}, we discuss about how the parameters in the algorithm can be chosen.
    \item In Appendix~\ref{appx: step-function}, we provide the construction of the two random geometric graphs on distinct manifolds that are identically distributed when generated from the same step function $\rp$, as stated in Proposition \ref{prop:two-helices-m-n}.
    \item In Appenix \ref{sec: CS}, we provide one page list of all important parameters and events used throughout the paper for easy reference.
\end{enumerate}

\section*{Acknowledgments}
Han Huang and Pakawut Jiradilok and Elchanan Mossel were supported Bush Faculty Fellowship ONR-N00014-20-1-2826 and by Elchanan Mossel's Simons Investigator award (622132). Elchanan Mossel was also supported by awards
ARO MURI W911NF1910217,
NSF DMS-2031883,
and NSF award CCF 1918421.

Part of this work was done when Jiradilok was at Mahidol University International College (MUIC), Nakhon Pathom, Thailand. Jiradilok would like to express his gratitude towards MUIC, especially its Science Division, for its hospitality.

We would like to thank Anna Brandenberger, Tristan Collins, Dan Mikulincer, Ankur Moitra, Greg Parker, and Konstantin Tikhomirov for helpful discussions.

\section{Definitions and Parameters}
\label{sec: RGM}
\subsection{Notations}
In this subsection, we define a few notations so we can use them throughout the paper and also the appendix. We fix a positive integer $N$. For $p \in \R^N$ and $r \ge 0$, let
\[
B(p,r) := \{ x \in \R^N \,:\, \|x-p\|<r\},
\]
where throughout this paper we use $\| \cdot \|$ to denote the Euclidean norm in $\mathbb{R}^N$. We use $\mathbb{S}^{N-1}$ to denote the set of all unit vectors in $\mathbb{R}^N$, namely, $w \in \mathbb{R}^N$ with $\|w\| = 1$. For any (linear or affine) subspace $H \subseteq \mathbb{R}^N$, we let $P_H$ be the orthogonal projection from $\mathbb{R}^N$ to $H$.

\bigskip

\begin{definition}\label{def:dist-btwn-d-spaces}
Let $H_1,H_2 \subseteq \mathbb{R}^N$ be linear subspaces of dimension $d \ge 1$. Define
\begin{align} \label{eq:dH1H2}
\sd(H_1,H_2) := \max_{v\in H_2, \, \|v\| = 1} \|P_{H_1^\perp}v\|.
\end{align}
We also extend the definition to affine subspaces as follows. For any affine subspaces $A_1, A_2 \subseteq \mathbb{R}^N$ of dimension $d \ge 1$, take any points $a_1 \in A_1$ and $a_2 \in A_2$, and define
\[
\sd(A_1, A_2) := \sd(A_1 - a_1, A_2 - a_2).
\]
\end{definition}

\begin{remark}\label{rem: Gr-N-d}
When we consider only linear subspaces, the function $\sd: {\rm Gr}(N,d) \times {\rm Gr}(N,d) \to \mathbb{R}_{\ge 0}$, where ${\rm Gr}(N,d)$ denotes the Grassmannian of $d$-spaces in $\mathbb{R}^N$, is a distance function. However, one needs to be careful when extending the definition of $\sd$ to affine spaces, where $\sd$ is no longer a distance function, as $\sd(A,A')$ vanishes whenever $A$ and $A'$ are parallel.
\end{remark}

\subsection{Random Graph Model} \label{sub: RGM}
This subsection describes the random graph model.

\subsubsection{Manifold and probability measure}\label{subsubs: M-and-pm}
\begin{assumption}[Manifold and probability measure]\label{assump: M-and-pm}
Let $d$ be a fixed positive integer. Let $M$ be a \( d \)-dimensional compact complete connected smooth manifold embedded in $\R^{N}$ with its Riemannian metric induced from $\mathbb{R}^{N}$. Let $\mu$ be a probability measure on $M$. We make the following definitions and assumptions.
\begin{enumerate}
\item[(i)] Let
\[
\kappa := \max_{p \in M} \, \max_{\substack{u,v\in T_pM, \\ \|u\|=\|v\|=1 }} \| {\rm II}(u,v)_p \|,
\]
where ${\rm II}$ denotes the {\it second fundamental form} of $M$ (see the appendix).
For readers unfamiliar with the second fundamental form, we note that $\kappa$ provides a measure of how curved the manifold $M$ is. In particular, if $M$ is a sphere of radius $R$, then $\kappa = 1/R$.

\item[(ii)] Let $r_{M,0}$ be the largest value such that for every point $p \in M$ and for every real number $r \in (0, r_{M,0})$, the intersection $B(p,r) \cap M$ is connected. Here, we assume
$$
    r_{M,0} >0.
$$
and we allow $r_{M,0} = \infty$.
\item[(iii)] Define $\mu_{\min}: [0, \infty) \to [0,1]$ by
\begin{equation} \label{eq:muAssumption}
\rv(r) := \min_{p\in M}  \mu( B(p,r)\cap M),
\end{equation}
for each $r \ge 0$. We assume that there exists a pair of positive values $(r_{M,1},c_\mu)$ such that
 $$\inf_{r \in (0, r_{M,1}]} \mu_{\min}(r) \cdot r^{-d} > c_\mu$$
for some constant $c_{\mu} >0$.

\item[(iii)] Let
\begin{equation} \label{eq:r_M}
r_M := 0.01 \cdot \min\{ 1 /\kappa, r_{M,0} ,r_{M,1}\}.
\end{equation}
Note that since $\kappa$ is strictly positive, we have $r_M < \infty$.
\end{enumerate}
\end{assumption}

\begin{rem}
We remark that the uniform distribution on $M$ satisfies Assumption~\ref{assump: M-and-pm}(iii), since the volume of a ball is exactly proportional to $r^d$ for small $r$. Moreover, this condition is one-sided. For instance, the probability measure $\mu$ may be taken as a mixture of the uniform distribution with discrete distributions supported on finitely many points of $M$.
\end{rem}

\subsubsection{Distance-probability function}\label{subsubs: Pdf}
\begin{assumption}[distance-probability function]\label{assump: pdf}
Let $\rp: [0,\infty) \to [0,1]$ be a smooth monotone strictly decreasing function such that $\rp'(t) = \frac{\pd}{\pd t} \rp(t) < 0$ for every $t \ge 0$, and assume that it is Lipschitz continuous with a bounded Lipschitz constant denoted by $L_{\rp} < \infty$.

Let $\ell_{\rp}>0$ be the largest positive value such that
\begin{equation}\label{eq: lp}
\forall \, 0\le a\le b \le 2{\rm diam}_{\rm euc}(M), \qquad \rp(a)-\rp(b) \ge \ell_{\rp} |b-a|.
\end{equation}
Equivalently,
\[
\ell_\rp := \min_{x \in [0,2{\rm diam}_{\rm euc}(M)]} |\rp'(x)| > 0.
\]
\end{assumption}

\subsubsection{Graph Model}\label{subsubs: GM}
Let ${\bf V}$ be a nonempty finite index set. We construct the random (simple) graph $G=G({\bf V},M,\mu,\rp)$ in the following way. The graph $G$ has vertex set ${\bf V}$. Let $\{X_i\}_{i\in {\bf V}}$ be i.i.d. random points in $M$ generated according to the probability measure $\mu$. Each pair of distinct vertices $i$ and $j$ are connected by an edge in $G$ with probability $\rp(\|X_i-X_j\|)$ independently. We give the following formal definition of the graph.

\begin{definition} \label{def: graphModel}
    Given an embedded manifold $M$ in a Euclidean space, a probability measure $\mu$ on $M$, and a distance-probability function $\rp:[0,\infty) \to [0,1]$, the {\em random geometric graph} $G = G( {\bf V}, M, \mu, \rp)$ with vertex set ${\bf V}$ is the random graph with adjacency matrix $A = (a_{i,j})_{i,j\in {\bf V}}$
    given by
    \begin{align} \label{eq: adjacencyMatrix}
    a_{i,j} = \begin{cases}
        {\bf 1} \big( \eU_{i,j}  \le \rp( \|X_i-X_j\|)\big) & i\neq j, \\
        0 & i=j,
    \end{cases}
    \end{align}
    where
    \begin{itemize}
        \item $ X = \{X_i\}_{i \in {\bf V}}$ are i.i.d. random points in $M$ generated according to $\mu$, and
        \item $ \eU = \{\eU_{i,j}\}_{i,j \in {\bf V}}$ are i.i.d. random variables uniformly distributed over the interval $[0,1]$, subject to the constraints that $\eU_{i,j}=\eU_{j,i}$, and $\eU_{i,i} = 0$.
    \end{itemize}
    \end{definition}
Furthermore, we write $G(n',M,\mu,\rp)$ to denote such a random graph with a given vertex set of size $n'$ without specifying the vertex set itself.

\medskip

For any subsets $V, U\subseteq {\bf V}$, let $G_V$ be the induced subgraph of $G$ with vertex set $V$. We use $N(i)$ and $N_V(i)$ to denote the neighbors of $i$ in ${\bf V}$ and the neighbors of $i$ in $V$, respectively. For example, $N_V(i)$ can be observed from $G_{V \cup \{i\}}$.
Furthermore, let
\begin{align} \label{eq: XVSV}
   X_V:= \{X_i\}_{i\in V} \qquad \mbox{and} \qquad \eU_{V,U}:= \{\eU_{i,j}\}_{v \in V, u \in U}.
\end{align}
For simplicity, we also use $\eU_V := \eU_{V,V}$.

\subsubsection{Remark on the choice of $\ell_\rp$ with dependency on $M$.}
\label{sec: lp}
\begin{figure}
\centering

\begin{tikzpicture}[scale = 1.5]
\def\hs{2.5} 
\def\a{0.5}
\def\b{2}
\def\c{1}
\def\d{-0.5}

\begin{scope}[shift={(-\hs,0)}]
\draw[rounded corners, thick] (0,0) -- (\b,0) -- (\b,3) -- (0,3) -- (0,2) -- (+\a,2) -- (+\a,1) -- (0,1) -- cycle;
\node at (\c,\d) {$M_a$};
\end{scope}

\begin{scope}[shift={(+\hs,0)}]
\draw[rounded corners, thick] (0,0) -- (\b,0) -- (\b,3) -- (0,3) -- (0,2) -- (-\a,2) -- (-\a,1) -- (0,1) -- cycle;
\node at (\c,\d) {$M_b$};
\end{scope}
    
\end{tikzpicture}
\caption{Two closed curves.}\label{fig:2cc}
\end{figure}

Let \(M_a\) and \(M_b\) denote two embedded manifolds within \(\mathbb{R}^2\), characterized by the configurations depicted in Figure \ref{fig:2cc}. It is evident that an isometry \(F\) exists, respecting the intrinsic metric, such that \(F\) reflects the protruding segments of \(M_a\) onto the corresponding indented segments of \(M_b\) and acts identically elsewhere. A key observation is the existence of a constant \(t > 0\) ensuring that, for any points \(p, q \in M_a\) satisfying \(\|p - q\| \leq t\), the equation \(\|F(p) - F(q)\| = \|p - q\|\) holds.

Consider \(\mu_a\) and \(\mu_b\) as the uniform measures on \(M_a\) and \(M_b\), respectively. Notice that \(\mu_b\) is also the pushforward measure of \(\mu_a\) under \(F\). Assume a function \(\rp: [0, \infty) \to [0, 1]\) that is constant for \(x \geq t\). Now, let us couple the two graphs \(G([n], M_a, \mu_a, \rp)\) and \(G([n], M_b, \mu_b, \rp)\): Let \(X_1, \dots, X_n\) represent independent and identically distributed random points on \(M_a\), based on the distribution \(\mu_a\), and let \({\cal U}_{i,j}\) denote independent and identically distributed uniform random variables on \([0,1]\) for \(i, j \in [n]\), with the stipulation that \({\cal U}_{i,j} = {\cal U}_{j,i}\). A graph is then formed with vertex set \([n]\), connecting vertices \(i\) and \(j\) if \({\cal U}_{i,j} \leq \rp(\|X_i - X_j\|)\), yielding a graph distribution equivalent to \(G([n], M_a, \mu_a, \rp)\). Analogously, using the same vertex set but connecting \(i\) and \(j\) if \({\cal U}_{i,j} \leq \rp(\|F(X_i) - F(X_j)\|)\) results in a graph with the distribution of \(G([n], M_b, \mu_b, \rp)\).

Given that \(\rp(\|F(X_i) - F(X_j)\|) = \rp(\|X_i - X_j\|\) whether \(\|X_i - X_j\| \leq t\) (by virtue of the properties of \(M_a\) and \(M_b\)) or \(\|X_i - X_j\| > t\) (as \(\rp\) is constant beyond \(t\)), the resultant random graphs are indistinguishable. Consequently, it is impossible to recover the Euclidean structure from the graphs.

\subsection{Parameters and Events }
\label{sec: parameterAssumption}
In this subsection, we introduce variables (``parameters'') along with relations between them (``parameter assumptions''), which we frequently refer to throughout the paper. We also define some basic events closely tied to these parameters and the proof outline of Theorem~\ref{thm: NetInformal}.

{\it Since we frequently refer to these parameters and events, we have collected them in a single-page cheatsheet provided in Appendix~\ref{sec: CS}. For readers who wish to go over the paper in detail, we recommend printing the cheatsheet and keeping it handy for quick reference.}

We introduce a minor modification to simplify the exposition. Specifically, instead of considering a graph with exactly $n$ vertices as stated in the theorem, we consider a slightly larger graph whose vertex set size is explicitly chosen as $n \cdot (d+2) \cdot \lceil n^{\varsigma} \rceil$, where $\varsigma$ is a fixed constant satisfying
\begin{align}
\label{eq: varsigma}
0 < \varsigma < 1/4.
\end{align}
 We partition these vertices into batches of size $n$. Later, we will introduce an algorithm which will extract a cluster of latent points in $M$ from each batch.  Although this modification does not alter the statement of the theorem, it does affect certain constants by a factor of up to $2$.

From now on, we fix $\varsigma \in (0,1/4)$ and let $n$ be a positive integer. We define a finite vertex set ${\bf V}$ whose size is given explicitly by
\begin{align} \label{eq: Vsize}
    |{\bf V}| = n \cdot (d+2) \cdot \lceil n^{\varsigma} \rceil.
\end{align}
This formulation divide ${\bf V}$ into $(d+2) \cdot \lceil n^{\varsigma} \rceil$ batches, each of size $n$.

\begin{definition}\label{def:K}
For $x,y \in M$, we define the \defn{common neighbor probability} of $x$ and $y$ as
\[
\rK (x,y) := \mathbb{E}_{Z\sim \mu} {\rm p}(\|x-Z\|){\rm p}(\|y-Z\|).
\]
\end{definition}

\begin{remark}
For three different vertices $i,j,k \in {\bf V}$, conditioned on $X_i = x_i$ and $X_j = x_j$, the probability of $k \in N(i)\cap N(j)$ is $\rK(x_i,x_j)$. This explains the name of $\rK(x,y)$ above.
\end{remark}

We will introduce a sequence of increasing parameters
$$  \varepsilon <  \eta < \delta < r.$$
See Figure \ref{fig:parameter} for an intuitive illustration for their roles and a detailed explantion of these parameters are discussed below, in Section~\ref{sec: ProofIdea}.

Let $\varepsilon >0$ be a parameter such that
\begin{align}\label{eq: CondVarepsilon}
\rv ( \varepsilon / 6 ) \ge 2n^{-\varsigma}
\qquad \mbox{and} \qquad
\varepsilon \ge \frac{1}{\sqrt{\| \rK \|_\infty} L_\rp} n^{-1/2+ \varsigma},
\end{align}
where
$\|\rK\|_{\infty} := \max_{p,q \in M} \rK(p,q)$.
Due to $\mu_{\min}(r) \ge c_\mu r^d$ for $r \in (0,r_M]$ from Assumption~\ref{assump: M-and-pm} and $\varsigma < 1/4$ from \eqref{eq: varsigma}, the conditions in \eqref{eq: CondVarepsilon} implies that
\begin{align}
    \varepsilon = O(n^{-\varsigma/d})
\end{align}
when $n$ is large enough.

Let $\cdr := 2^{10}$. Let $c_1 \in (0,1)$, $C_2 \ge 1$, and $c_3 \in (0,1)$ be parameters such that
\begin{equation}\label{ineq:Cond-c1}
0 < c_1 \le \frac{1}{800} \cdot \ell_\rp^2 \cdot \mu_{\min}(r_M/4),
\end{equation}
\begin{equation}\label{ineq:Cond-C}
C_2 \ge 4 \cdot \frac{\|\rK\|_{\infty}^{1/4} C_{\rm gap}^{1/2} d^{1/2} L_{\rp}}{\ell_\rp^{1/2} c_1^{1/2}},
\end{equation}
and
\begin{equation}\label{ineq:Cond-c3}
0 < c_3 \le \frac{\ell_\rp}{\cdr \sqrt{d}}.
\end{equation}

Let $\eta>0$ be a parameter such that
\begin{equation}
\label{eq: CondEta}
\eta \ge \max\Big\{ C_2 \cdot \varepsilon^{1/2}, \frac{L_\rp^2}{c_3 \ell_\rp \sqrt{d}} \cdot \varepsilon \Big\},
\end{equation}
which is the parameter $\eta$ that shows up in Theorem \ref{thm: NetInformal}.

Let $\delta$ and $r$ be the parameters given by
\begin{equation}\label{eq: CondDeltaR}
\delta := \cdr \sqrt{d} \cdot \eta
\qquad \mbox{and} \qquad
r := \cdr d^2 \cdot \delta.
\end{equation}

Finally we assume that the choice of $r$ satisfies
\begin{align}\label{eq: parameterFeasible}
r_M \ge 2^4 \cdr d^2 r
\qquad \mbox{and} \qquad
n \ge 100.
\end{align}

We remark that a choice of $\varepsilon, \eta, \delta, r$ which satisfies these parameter assumptions always exists, when $n$ is large enough.

Notice that
\begin{align} \label{eq: deltaetaOrder}
    \delta  \simeq \eta \simeq \varepsilon^{1/2} \simeq n^{-\varsigma/2d},
\end{align}
when $n$ is large enough, and $a \simeq b$ means there exists $C = {\rm poly}(L_{\rp}, 1/\ell_{\rp},1/\|\rK\|_\infty, d, \mu_{\min}(r_M/4))$ such that $ \frac{1}{C}b \le a \le Cb$.

\begin{remark}
In Appendix~\ref{sec: GO} below, we discuss practical use of our results in this paper. There we will be explicit about what the graph observer is given and what the graph observer can compute to obtain ``graph-observer-accessible'' versions of parameters. Furthermore, Proposition~\ref{prop: graph-obs} in the appendix gives an explicit test, which can be carried out by the graph observer, to check whether the parameters are feasible.
\end{remark}

\subsection{Basic Events}
\begin{definition}\label{def:cn-net}
For any $W \subseteq {\bf V}$ with $|W|=n$, the \defn{common neighbor event} of $W$ is
\begin{equation} \label{eq: Ecommonneighbor}
\Ecn(W) := \left\{\forall \{i,j\} \in {W \choose 2},\, \Big| |N_W(i) \cap N_W(j)|/n - \rK(X_i,X_j) \Big| \le n^{-1/2+\varsigma}\right\},
\end{equation}
and the \defn{net event} of $W$ is
\begin{equation} \label{eq:Enet}
\Enet(W) :=
\bigg\{
\forall p \in M, \, \big| \left\{ i \in W \, : \, X_i \in B(p, \varepsilon) \right\} \big| \ge \rv(2\varepsilon/3) \cdot \frac{n}{2}
\bigg\}.
\end{equation}
\end{definition}

The term ``cluster'' which we use throughout this paper is defined as follows.
\begin{definition}\label{def:cluster}
 A \defn{cluster} is a pair $(V,i)$ for $i \in V \subseteq {\bf V}$ which satisfies the following two properties:
\begin{itemize}
    \item[(i)]  for each $j \in V$, $\|X_j - X_i\|< \eta$.
    \item[(ii)] $|V| \ge n^{1-\varsigma}$.
\end{itemize}
Further, we say that $(V,i)$ is a $t$-\defn{cluster} for $t>0$ if condition $(i)$ is replaced by
$\|X_j - X_i\|<t$ for $j \in V$.
\end{definition}

\begin{definition}
    For a finite index set $A$ and a collection of pairs $\{(V_\alpha, i_\alpha)\}_{\alpha \in A}$ with $i_\alpha \in V_\alpha \subseteq {\bf V}$, the \defn{cluster event} of the collection $\{(V_\alpha, i_\alpha)\}_{\alpha \in A}$ is defined as
    \begin{align}
        \label{defi clu}
    \Eclu(\{(V_\alpha, i_\alpha)\}_{\alpha \in A}) := \big\{ \text{for each } \alpha \in A, \, \text{the pair } (V_\alpha, i_\alpha) \text{ is a cluster} \big\}.
    \end{align}
\end{definition}

Suppose we have a collection of clusters $\{(V_\alpha, i_\alpha)\}_{\alpha \in A}$ and a given vertex set $W$. If the number of edges from each cluster to vertices in $W$ is close to its expected value, we can then determine the distances from each vertex $w \in W$ to all clusters. Formally, we define the corresponding event as follows.

\begin{definition}\label{def:navi}
    Suppose that $A$ is an index set. For a collection $\{V_\alpha\}_{\alpha \in A}$ of nonempty subsets of ${\bf V}$ and for a subset $W \subseteq {\bf V}$, the \defn{navigation event} of $(\{V_\alpha\}_{\alpha \in A}, W)$ is
    \begin{equation}\label{eq: E-navi}
    \Enavi(\{V_\alpha\}_{\alpha \in A}, W) := \bigg\{
    \forall i \in W,
    \forall \alpha \in A,
    \bigg| \frac{|N(i) \cap V_\alpha|}{ |V_\alpha|} -
    \sum_{j\in V_\alpha} \frac{\rp(\|X_i-X_j\|)}{|V_\alpha|}
    \bigg| \le n^{-1/2 + \varsigma}
    \bigg\}.
    \end{equation}
\end{definition}

\subsection{Additional Notation and Tools}
For each $i \in {\bf V}$, let
\begin{align}
\label{eq: H_i}
   H_i := T_{X_i}M
\end{align}
be the tangent plane of $M$ at the point $X_i$. Let ${\rm diam}(M)$ and ${\rm diam}_{\rm gd}(M)$ be the diameters of $M$ with respect to the Euclidean distance and the geodesic distance, respectively.

Let us also recall the standard Hoeffding's inequality for the sum of bounded independent random variables.

\begin{lemma}[Hoeffding's inequality]
For any $n \ge 1$, let $Y_1, Y_2, \dots, Y_n$ be independent random variables such that $Y_i \in [a_i, b_i]$ for each $i \in [n]$. Then for any $t > 0$, we have
\[
    \mathbb{P}\Big\{ \Big|  \sum_{i \in [n]} Y_i - \mathbb{E}\sum_{i \in [n]}Y_i \Big| \ge t \Big\} \le 2 \exp\left( - \frac{2t^2}{\sum_{i=1}^n (b_i-a_i)^2} \right).
\]
\end{lemma}
In particular, if $\{Y_i\}_{i \in [n]}$ are i.i.d. ${\rm Bern}(p)$ random variables, we have
\[
    \mathbb{P}\Big\{ \Big|  \sum_{i \in [n]} Y_i - \mathbb{E}\sum_{i \in [n]}Y_i \Big| \ge t \Big\} \le 2 \exp\left( - \frac{2t^2}{n} \right).
\]

\section{Common neighbor probability $\rK(x,y)$}\label{sec: Kxy}
Our focus of this section is on the \defn{common neighbor probability} function $\rK$ introduced in Definition \ref{def:K}:
\[
\rK (x,y) := \mathbb{E}_{Z\sim \mu} {\rm p}(\|x-Z\|){\rm p}(\|y-Z\|).
\]
The main goal in this section is to derive the following property of $\rK(x,y)$.

\begin{lemma}\label{lem:KxyKmax}
For any $x, y \in M$, we have
\[
\rK(x,y) \le \frac{ \rK(x,x)+\rK(y,y)}{2} - c_1 \cdot \min\{\|x-y\|^2,r_M^2\},
\]
where $c_1$ is the parameter described in Section~2.
\end{lemma}

An immediate consequence of the lemma is
\begin{align}
\label{eq: rK(x,y)}
\rK(x,y) \le \|\rK\|_\infty - c_1 \cdot \min\{\|x-y\|^2,r_M^2\},
\end{align}
where $\|K\|_\infty = \max_{x,y \in M} \rK(x,y)$.

Suppose $|{\bf V}|$ is sufficiently large so that $|N(i)\cap N(j)| \simeq (|{\bf V}|-2)\rK(X_i,X_j)$ for each $\{i,j\} \in { {\bf V} \choose 2}$. Then, by comparing $|N(i)\cap N(j)|$ with $\max_{i',j' \in {\bf V}} |N(i')\cap N(j')|$, we can use \eqref{eq: rK(x,y)} to derive information about the Euclidean distance $\|X_i-X_j\|$.

Let us begin with some standard properties about ${\bf K}(x,y)$.

\begin{remark} \label{rem: Lipschitz K}
$\rK(x,y)$ is Lipschitz continuous in each variable with Lipschitz constant $\le L_{\rp}\sqrt{\|\rK\|_\infty}$. To see this, observe that
\begin{align} \label{eq:LipK}
|\rK(x,y)- \rK(x',y)|
& \le \int_M  \big|\rp(\|x-z\|)-\rp(\|x'-z\|)\big| \, \rp(\|y-z\|) \, \pd \mu(z) \\
\nonumber
& \le L_{ \rp} \|x-x'\|   \int_M \rp(\|y-z\|) \pd \mu(z) \\
\nonumber
& \le L_{\rp }\|x-x'\| \sqrt{\|\rK\|_\infty}.
\end{align}
The last inequality follows from Cauchy--Schwarz:
\[
\left( \int_M \rp(\|y-z\|) \, \pd \mu(z) \right)^2 \le \int_M \rp(\|y-z\|)^2 \, \pd \mu(z) = \rK(y,y) \le \|\rK\|_{\infty}.
\]
\end{remark}

Using the formula $(a-b)^2 = a^2-2ab +b^2$, we have
\begin{equation}\label{eq:Kdiagonal}
\rK(x,y) = \frac{ \rK(x,x)+\rK(y,y)}{2} - \frac{1}{2}\int_M (\rp(\|x-z\|) - \rp(\|y-z\|))^2 \, \pd \mu(z).
\end{equation}
Since the integral on the right-hand side of the equation above is non-negative, it follows that
\[
\rK(x,y) \le \frac{ \rK(x,x)+ \rK(y,y)}{2}.
\]
Recalling that $\rK$ is continuous and $M$ is compact, we find that the maximum of $\rK$ is attained at $\rK(x,x)$ for some (not necessarily unique) point $x \in M$.

\begin{remark}\label{rem: Lp-rK-gtr-0}
We remark that $L_\rp\sqrt{\|\rK\|_\infty} \ge |\rp'(0)| \cdot \rp({\rm diam}(M)) > 0$
is strictly positive.
\end{remark}

Let us now give a quick useful lemma about the function $\rp$ as follows.

\begin{proof}[Proof of Lemma \ref{lem:KxyKmax}]
It suffices to prove that there exists a point $z_0 \in B(x,\frac{3}{4} r_M) \cap M$ such that every point $z \in B(z_0, \frac{1}{4} r_M) \cap M$ satisfies
\[
\Big| \|z-x\| - \|z-y\| \Big| \ge \frac{1}{20} \min\{\|x-y\|,r_M\}.
\]
Suppose such point $z_0$ exists. Consider $z \in B(z_0, \frac{1}{4} r_M) \cap M$. From triangle inequality we have
$$ \|z-x\| \le \|z-z_0\| + \|z_0 - x\|\le r_M.$$
Then, by the monotonicity of $\rp$,
$$
   \big| \rp(\|z-x\|) - \rp(\|z-y\|) \big|
\ge
    \max_{s \in [0,r_M]} \rp(s) - \rp\Big(s+\frac{1}{20}\min\{ \|x-y\|, r_M\} \Big)
\ge
    \frac{\ell_\rp}{20} \min\{ \|x-y\|, r_M\},
$$
where the last inequality follows from the definition of $\ell_\rp$. Hence,
\begin{align*}
\int_M \left( \rp(\|x-z\|) - \rp(\|y-z\|)\right)^2 \, \pd\mu(z) &\ge \int_{B(z_0, r_M/4) \cap M} \left| \rp(\|x-z\|) - \rp(\|y-z\|) \right|^2 \, \pd \mu(z) \\
&\ge \mu_{\min}\!\left( \frac{r_M}{4} \right) \cdot \frac{\ell_{\rp}^2}{400} \cdot \min \left\{ \|x-y\|, r_M \right\}^2 \\
&\ge 2c_1 \cdot \min\left\{ \|x-y\|^2 , r_M^2 \right\}.
\end{align*}

Here, we capture the term $\mu_{\min}\!\left( \frac{r_M}{4} \right) \cdot \frac{\ell_{\rp}^2}{400}$ and introduce the parameter $c_1$ with the constraint that
$$
    c_1
\le
    \frac{1}{2} \mu_{\min}\!\left( \frac{r_M}{4} \right) \cdot \frac{\ell_{\rp}^2}{400},
$$
to simplify our expression, where the additional factor of $\frac{1}{2}$ was included to compensate for cancellation in the derivation later.

With the above bound and Equation~\eqref{eq:Kdiagonal}, the lemma immediately follows.

\medskip

Now we prove that $z_0$ exists. If $\|x-y\|\ge r_M$, then we can simply take $z_0=x$. In this case, for every $z \in B(x, \frac{1}{4} r_M) \cap M$, we have
\[
\|z-y\| -\|z-x\| \ge \|x-y\| - \|z-x\| -\|z-x\| \ge r_M/2.
\]
It thus suffices to consider $\|x-y\| < r_M$. Note that the desired inequality holds trivially when $x = y$, so let us assume $0 < \|x-y\| < r_M$.

With a shift and a rotation, we may assume $x=\vec{0}$ and $T_{\vec{0}}M = \R^d$. From Proposition~\ref{prop: SecondFundamentalForm} in the appendix, we know that there exists a map $\phi: B(\vec{0},r_M) \cap \R^d \to M$  such that $\phi \circ P\vert_{ B(\vec{0},r_M)\cap M} = {\rm Id}_{ B(\vec{0},r_M)\cap M}$, where $P$ is the orthogonal projection from $\R^N$ to $\R^d$. Furthermore, for any $v \in B(\vec{0},r_M) \cap \mathbb{R}^d$, we can express $\phi(v) = v+\tilde \phi(v)$, where $\tilde \phi(v) \in T_{\vec{0}}M^\perp$ and $\|\tilde \phi(v)\|\le \kappa \|v\|^2$.

Now, we express $y= Py + y^{\perp}$, where $y^\perp = y - Py =  \tilde \phi(Py)$. Note that since $y \neq \vec{0}$, we know by Corollary~\ref{cor:PH-diffeo} in the appendix, that $Py \neq \vec{0}$, from which it makes sense to consider
$$
    u
:=
    -\frac{Py}{\|Py\|}
\quad \text{and take} \quad
z_0 := \phi( \frac{r_M}{2}u) \in M.
$$

First, with $ \|Pz_0\| = \left\| \frac{r_M}{2} u \right\| = \frac{r_M}{2} < r_M \le \frac{0.01}{\kappa}$,
\[
\|z_0\| \le \sqrt{\left\| \frac{r_M}{2} u \right\|^2 + \kappa^2 \left\| \frac{r_M}{2} u \right\|^4} < \frac{3}{4} r_M,
\]
and hence ${\cal B} := B( z_0 , r_M/4) \cap M \subseteq B(\vec{0},r_M) \cap M$.  Therefore,
\[
    {\cal B}
 =
    \phi \circ P ( {\cal B})
\subseteq
    \phi\Big(B\Big( \frac{r_M}{2}u,r_M/4\Big) \cap \R^d\Big),
\]
since the projection $P$ is a contraction.

Next, for any $z \in {\cal B}$, we can decompose $z= au + w$ with $w\perp u$. We will estimate $\|z-y\| -\|z-\vec{0}\|$ from below.
First,
\begin{equation} \label{eq: KxyKmax00}
\|z-y\| - \|z - \vec{0}\| = \frac{\|z-y\|^2 - \|z \|^2}{\|z-y\| + \|z\|}.
\end{equation}

To estimate the numerator, we find by decomposing $\|z-y\|^2$ that
\begin{align}
\nonumber
         \|z-y\|^2
& =
         \langle z-y, u \rangle^2 +  \|w- y^\perp\|^2\\
\nonumber
& =
        a^2  + 2a\|Py\| + \|Py\|^2
        + \|w\|^2 - 2 \langle w, y^\perp\rangle + \|y^\perp\|^2 \\
\nonumber
& \ge
         a^2  + 2a\|Py\| + \|w\|^2- 2 \|w\| \cdot \|y^\perp\| \\
\label{eq: KxyKmax01}
& =
         \|z\|^2  + 2a\|Py\| - 2 \|w\| \cdot \|y^\perp\|.
\end{align}
Now we want to bound the second and the third summands of the above equation.
First, $a = \langle z,u \rangle = \langle Pz,u\rangle$ and $Pz \in
B(\frac{r_M}{2}u,r_M/4) \cap \R^d$, we have $$a \ge \frac{r_M}{2} - \frac{r_M}{4} \ge  \frac{r_M}{4}.$$
Second, $$\|w\| \le \|z\| \le \|z-z_0\| + \|z_0\| \le \frac{1}{4}r_M + \frac{3}{4}r_M \le r_M.$$
Last, with $\|Py\| \le \|y\| <r_M$, we can apply the property of $\tilde \phi$ to get
that
\begin{align}
\label{eq: KxyKmax02}
 \|y^\perp\| = \|\tilde\phi(Py)\| \le \kappa \|Py\|^2 \le \kappa r_M  \|Py\| \le 0.01 \|Py\|.
\end{align}

With the above three estimates substituted into \eqref{eq: KxyKmax01}, we have
\begin{align*}
    \|z-y\|^2
& \ge
\|z\|^2 + (0.5-0.02)r_M\|Py\|.
\end{align*}

Together with $\|y\|^2 = \|Py\|^2 + \|y^\perp\|^2 \overset{\eqref{eq: KxyKmax02}}{\le} 1.01 \|Py\|^2$,
the numerator of the right-hand side of Equation~\eqref{eq: KxyKmax00} can be bounded below by
\[
\|z-y\|^2 - \|z\|^2 \ge  \frac{0.5-0.02}{1.01} r_M\|y\|.
\]
With the estimate $\|z-y\| + \|z\| \le 3r_M$ for the denominator in \eqref{eq: KxyKmax00}, we conclude
\begin{align*}
   \|z-y \| - \|z - \vec{0}\| \ge \frac{1}{20} \|y\|,
\end{align*}
and the proof is completed.
\end{proof}

\section{The Cluster-Finding Algorithm and Good Pairs}\label{sec: good-pairs}
In this section, we describe our main algorithm \gclu~ which we use multiple times in this paper to find clusters. (See Definition~\ref{def:cluster} for the definition of a cluster.) We introduce in Definition~\ref{def:good-pair} the notion of a ``good pair.'' The upshot is that if we input a good pair into Algorithm~\gclu, then under certain assumptions, the algorithm gives a cluster (Lemma~\ref{lem: good-pair-gives-cluster}). The main achievement of this section is Proposition~\ref{prop: 1cluster}, in which we obtain the first cluster using the algorithm.

The following is the algorithm \gclu\footnote{See Section~\ref{sec: GO} for how the graph observer can practically use the algorithms in this paper. All of the four algorithms in this paper are practical.}  which we use to find clusters in the paper.

\medskip

\begin{algorithm}
\caption{\gclu}
\label{alg:cluster}

\SetKwInOut{Input}{Input}
\SetKwInOut{Output}{Output}

\Input{
$V_1 \subseteq {\bf V}$, a subset of vertices of size $n$. \\
$V_2 \subseteq {\bf V}$, a subset of at least $2$ vertices. \\
}
\Output{
    $W_{\rm gc} \subseteq V_2$, a subset of vertices. \\
    $i_{\rm gc} \in W_{\rm gc}$, a vertex.
}
{\em Step 1.} Sort \(\{i,j\} \in { V_2 \choose 2 }\) according to $|N_{V_1}(i)\cap N_{V_1}(j)|$ from the largest to the smallest, and return a list $\{i_1, j_1\}$, $\{i_2, j_2\}$, $\ldots$, $\{i_L, j_L\}$, where $L = \binom{|V_2|}{2} = \big| \binom{V_2}{2} \big|$.

{\em Step 2.} Let $m$ be the largest positive integer such that
\[
|N_{V_1}(i_m)\cap N_{V_1}(j_m)|/n \ge  |N_{V_1}(i_1)\cap N_{V_{1}}(j_1)|/n - \frac{1}{2} c_1 \eta^2.
\]

{\em Step 3.} Consider a graph
$\gc{G}$
with vertex set $ \gc{V} := \big( \bigcup_{k \in [m]} \{i_k\}\big) \cup \big( \bigcup_{k \in [m]} \{j_k\} \big)$ and edge set $\gc{E} := \left\{ \{i_k,j_k\} \, : \, k \in [m] \right\}$.

{\em Step 4.} Take a pair $(\gc{W}, \gc{i})$ where $\gc{i} \in \gc{V}$ maximizes the size of neighbors in $\gc{G}$ and
$\gc{W} = \left\{j \in \gc{V} \, : \, \{\gc{i}, j\} \in \gc{E} \right\} \cup \{\gc{i}\}$.

\Return{$( \gc{W}, \gc{i} )$}
\end{algorithm}
\paragraph{Complexity.} Algorithm~\ref{alg:cluster} runs in
\begin{align}
    \label{eq: gclu-runningtime}
    O(n^3)\,.
\end{align}
The dominant cost is computing $\bigl|N_{V_1}(i)\cap N_{V_1}(j)\bigr|$ for every ${i,j}\in\binom{V_2}{2}$, which costs $O\bigl(\binom{n}{2},n\bigr)=O(n^{3})$.\
The subsequent sorting step costs $O\bigl(\binom{n}{2}\log\binom{n}{2}\bigr)=O(n^{2}\log n)$, which is asymptotically smaller.

\medskip

Notice that by construction the output $( \gc{W}, \gc{i})$ always satisfies $ \gc{i} \in \gc{W} \subseteq V_2$.

\begin{definition}\label{def:good-pair}
A pair $(W_1, W_2)$ of subsets of ${\bf V}$ is said to be a \defn{good pair} if the following conditions are satisfied:
\begin{itemize}
    \item $\varnothing \neq W_2 \subseteq W_1$,
    \item $|W_1| = n$,
    \item for every $i \in W_2$, there exists $p \in M$ such that
    \begin{itemize}
        \item[(i)] $\|X_i - p\| < \cdr d \frac{L_\rp}{\ell_\rp} \varepsilon$, and
        \item[(ii)] $\{j \in W_1 \, : \, \|p - X_j\| < \varepsilon\} \subseteq W_2$.
    \end{itemize}
\end{itemize}
\end{definition}

\begin{lemma}[Good pair lemma]\label{lem: good-pair-gives-cluster}
Assume that the parameters are feasible.
If $(W_1, W_2)$ is a good pair
and $W_1$ is a sample within the events $\Ecn(W_1)$ and $\Enet(W_1)$, then the output $( \gc{W}, \gc{i} )$ of Algorithm~\gclu~ with input $V_1 = W_1$ and $V_2 = W_2$ is a cluster with $\gc{W} \subseteq W_2$.
\end{lemma}

\begin{proof}
Before even discussing the output $(\gc{W}, \gc{i})$, let us begin by checking that $(W_1, W_2)$ is an appropriate input for the algorithm. Namely, we argue why $W_2$ has at least two elements, and along the way we introduce a set $S$ which we will use in later parts of this proof.

Since $(W_1, W_2)$ is a good pair, we know that $W_2$ is nonempty. Thus, Lemma~\ref{lem:KxyKmax} implies that there exists $i_0 \in W_2$ such that
\[
\rK(X_{i_0}, X_{i_0}) = \max_{i,j \in W_2} \rK(X_i, X_j).
\]

Consider the set
\[
S := \left\{ j \in W_2 : \|X_j - X_{i_0}\| < 2\cdr d \frac{L_\rp}{\ell_\rp} \varepsilon \right\}.
\]
Since $(W_1, W_2)$ is a good pair and $i_0 \in W_2$, there exists $p \in M$ such that $\|X_{i_0} - p\| < \cdr d \frac{L_\rp}{\ell_\rp} \varepsilon$, and
\[
S' := \left\{ j \in W_1 \, : \, \|p - X_j\| < \varepsilon \right\} \subseteq W_2.
\]
Using the triangle inequality together with the above inclusion, we obtain
\[
S' \subseteq S \subseteq W_2.
\]
We know from the event $\Enet(W_1)$ that
\[
|S'| \ge \mu_{\min}\!\left( \frac{2\varepsilon}{3}\right) \cdot \frac{n}{2} \ge \mu_{\min}\!\left( \frac{\varepsilon}{6}\right) \cdot \frac{n}{2}
\overset{\eqref{eq: CondVarepsilon}}{\ge} 2n^{-\varsigma} \cdot \frac{n}{2} = n^{1-\varsigma},
\]
which implies that
\[
|W_2| \ge |S| \ge |S'| \ge n^{1-\varsigma}.
\]
In particular, by the feasibility assumption, we know $n \ge 100$, and thus $|W_2| \ge 100^{3/4} > 31$. Hence, we have established that $(W_1, W_2)$ is an appropriate input for Algorithm~\gclu.

\medskip

Now we will show $(\gc{W}, \gc{i})$ is a cluster starting with condition (i) from Definition \ref{def:cluster}. For any pair $\{s_1, s_2\} \in \binom{S}{2}$, the event $\mathcal{E}_{\rm cn}(W_1)$ implies that
\[
\frac{|N_{W_1}(s_1) \cap N_{W_1}(s_2)|}{n} \ge \rK(X_{s_1}, X_{s_2}) - n^{-\frac{1}{2} + \varsigma}.
\]
By the definition of $S$, both $\|X_{s_1}-X_{i_0}\|$ and $\|X_{s_2}-X_{i_0}\|$ is bounded by  $2\cdr d \frac{L_\rp}{\ell_\rp} \varepsilon$.

Together with the Lipschitz constant of $\rK$ is bounded above by $L_\rp \sqrt{\|\rK\|_\infty}$ (See Remark \ref{rem: Lipschitz K}),
we have
\[
\rK(X_{s_1}, X_{s_2}) \ge \rK(X_{i_0}, X_{i_0}) - 2L_{\rp} \sqrt{\|\rK\|_{\infty}} \cdot 2 \cdr d \frac{L_{\rp}}{\ell_{\rp}} \varepsilon.
\]
Combining the two inequalities above yields
\begin{equation}\label{ineq:any-s1-s2}
\frac{|N_{W_1}(s_1) \cap N_{W_1}(s_2)|}{n} \ge \rK(X_{i_0}, X_{i_0}) - 2L_{\rp} \sqrt{\|\rK\|_{\infty}} \cdot 2 \cdr d \frac{L_{\rp}}{\ell_{\rp}} \varepsilon - n^{-\frac{1}{2} + \varsigma},
\end{equation}
for any pair $\{s_1, s_2\} \in \binom{S}{2}$. Since $S$ has at least $n^{1-\varsigma} > 2$ elements and because $S \subseteq W_2$, we have the following bound:
\begin{equation}\label{ineq:C1-1}
\max_{\{i',j'\} \in \binom{W_2}{2}} \frac{|N_{W_1}(i') \cap N_{W_1}(j')|}{n} \ge \rK(X_{i_0}, X_{i_0}) - 4 \sqrt{\|\rK\|_{\infty}} \cdr d \frac{L_{\rp}^2}{\ell_{\rp}} \varepsilon - n^{-\frac{1}{2} + \varsigma}.
\end{equation}

Now consider the graph $\gc{G}$ from Algorithm~\gclu. Observe that a pair $\{i,j\} \in \binom{W_2}{2}$ appears as an edge in $\gc{G}$ if and only if
\begin{equation}\label{ineq:C1-2}
\frac{|N_{W_1}(i) \cap N_{W_1}(j)|}{n} \ge \max_{\{i',j'\} \in \binom{W_2}{2}} \frac{|N_{W_1}(i') \cap N_{W_1}(j')|}{n} - \frac{1}{2} c_1 \eta^2.
\end{equation}
Consider any pair $\{i,j\} \in \binom{W_2}{2}$ which satisfies the above inequality. We claim $\|X_i - X_j\| < \eta$.

Suppose, for the sake of contradiction, that $\|X_i - X_j\| \ge \eta$. Then by $\mathcal{E}_{\rm cn}(W_1)$ and Lemma~\ref{lem:KxyKmax}, we find
\begin{align}
\frac{|N_{W_1}(i) \cap N_{W_1}(j)|}{n} &\le \rK(X_i, X_j) + n^{-\frac{1}{2} + \varsigma}
\le \rK(X_{i_0}, X_{i_0}) - c_1 \eta^2 + n^{-\frac{1}{2} + \varsigma}, \label{ineq:C1-3}
\end{align}
where we also use that $r_M \ge \eta$ in the last inequality. This follows from the feasibility assumption: $r_M \ge 2^4 \cdr d^2 r > r > \delta > \eta$. Combining \eqref{ineq:C1-1}, \eqref{ineq:C1-2}, and \eqref{ineq:C1-3}, we find the following inequality
\[
4 \sqrt{\|\rK\|_{\infty}} \cdr d \frac{L_{\rp}^2}{\ell_{\rp}} \varepsilon + 2n^{-\frac{1}{2} + \varsigma} \ge \frac{1}{2} c_1 \eta^2,
\]
which will be a contradiction if
\begin{align}
\label{eq: goodpair00}
  \frac{1}{4}c_1\eta^2  \ge   4 \sqrt{\|\rK\|_{\infty}} \cdr d \frac{L_{\rp}^2}{\ell_{\rp}} >  2n^{-\frac{1}{2} + \varsigma} .
\end{align}

This is precisely the reason why we impose the condition
$ \eta \ge  4\frac{\|\rK\|_{\infty}^{1/4} C_{\rm gap}^{1/2} d^{1/2} L_{\rp}}{\ell_\rp^{1/2} c_1^{1/2}}\varepsilon^{1/2}$ and
 $\varepsilon \ge \sqrt{\|\rK\|_\infty} L_\rp n^{-1/2+\varsigma}$. With these two conditions and
 $ \cdr d \frac{L_\rp}{\ell_\rp} >1$,  \eqref{eq: goodpair00} holds and hence a contradiction follows.

The argument above shows that every edge $\{i,j\}$ in $\gc{G}$ satisfies $\|X_i - X_j\| < \eta$. Since for each $i \in \gc{W} - \{\gc{i}\}$, the pair $\{\gc{i}, i\}$ is an edge in $\gc{G}$ by construction, we have established (i).

\medskip

It remains to establish (ii). Consider any pair $\{i', j'\} \in \binom{W_2}{2}$. Within the event $\mathcal{E}_{\rm cn}(W_1)$, we have
\[
\frac{|N_{W_1}(i') \cap N_{W_1}(j')|}{n} \le \rK(X_{i'}, X_{j'}) + n^{-\frac{1}{2} + \varsigma} \le \rK(X_{i_0}, X_{i_0}) + n^{-\frac{1}{2} + \varsigma}.
\]
This shows that
\[
\rK(X_{i_0}, X_{i_0}) \ge \max_{\{i',j'\} \in \binom{W_2}{2}} \frac{|N_{W_1}(i') \cap N_{W_1}(j')|}{n} - n^{-\frac{1}{2} + \varsigma}.
\]
Using the above estimate with \eqref{ineq:any-s1-s2}, we find that every pair $\{s_1, s_2\} \in \binom{S}{2}$ satisfies
\begin{align*}
\frac{|N_{W_1}(s_1) \cap N_{W_1}(s_2)|}{n}
&\ge \max_{\{i',j'\} \in \binom{W_2}{2}} \frac{|N_{W_1}(i') \cap N_{W_1}(j')|}{n} - 4 \sqrt{\|\rK\|_{\infty}} \cdr d \frac{L_{\rp}^2}{\ell_{\rp}} \varepsilon - 2n^{-\frac{1}{2} + \varsigma} \\
&> \max_{\{i',j'\} \in \binom{W_2}{2}} \frac{|N_{W_1}(i') \cap N_{W_1}(j')|}{n} - \frac{1}{2} c_1 \eta^2,
\end{align*}
where the last inequality follows from the parameter assumptions. This implies that every pair $\{s_1, s_2\} \in \binom{S}{2}$ is an edge in $\gc{G}$. Therefore, the maximum degree in $\gc{G}$ is at least $|S| - 1 \ge n^{1-\varsigma} - 1$, and thus $|\gc{W}| \ge n^{1-\varsigma}$. We have completed the proof.
\end{proof}

As a consequence of the lemma above, we can use Algorithm~\gclu~to find the first cluster.

\begin{proposition}\label{prop: 1cluster}
Assume that the parameters are feasible. For any subset $W \subseteq {\bf V}$ with $|W|=n$, within the events $\Enet(W)$ and $\Ecn(W)$, Algorithm~\gclu~ with input $V_1=W$ and $V_2=W$ returns a cluster $(\gc W,\gc i)$ with $\gc i\in \gc W \subseteq W$.
\end{proposition}
\begin{proof}
It is straightforward to check that $(W,W)$ is a good pair. This proposition therefore follows immediately by applying Lemma~\ref{lem: good-pair-gives-cluster}.
\end{proof}

\section{Regularity of $\rp_\alpha$}
\label{sec: navi}
In the previous section, we described how we can use Algorithm~\gclu~ to generate a cluster. When we have a collection of clusters $(V_\alpha, i_\alpha)$, indexed by $\alpha$, we can define a corresponding collection of functions $\rp_\alpha: M \to \mathbb{R}$, which is an approximation of the function $q \mapsto \rp(\|q-X_{i_\alpha}\|)$ obtained by averaging $\rp(\|q-X_i\|)$ for $i \in V_\alpha$. Intuitively, due to the concentration of measure, for any given $i \in {\bf V}$, we expect
$$
    \rp(\|X_i - X_{i_\alpha}\|) \simeq    \rp_\alpha(X_i) := \sum_{j\in V_\alpha} \frac{\rp(\|X_i-X_j\|)}{|V_\alpha|} \simeq  \frac{|N_i \cap V_{\alpha}|}{|V_\alpha|},
$$
which allows us to estimate $\|X_i-X_{i_\alpha}\|$. The goal of this section is to derive regularity property of $\rp_\alpha$, stated in Lemma~\ref{lem:rpalphaRegular}.

Let us first recall the definition of the cluster event from \eqref{defi clu}:
For a finite index set $A$ and a collection of pairs $\{(V_\alpha, i_\alpha)\}_{\alpha \in A}$ with $i_\alpha \in V_\alpha \subseteq {\bf V}$, the \defn{cluster event} of the collection $\{(V_\alpha, i_\alpha)\}_{\alpha \in A}$ is defined as
\begin{align*}
\Eclu(\{(V_\alpha, i_\alpha)\}_{\alpha \in A}) := \big\{ \text{for each } \alpha \in A, \, \text{the pair } (V_\alpha, i_\alpha) \text{ is a cluster} \big\}.
\end{align*}

The discussion in this section is within the event
\[
\Eclu(\{(V_\alpha, i_\alpha)\}_{\alpha \in A}).
\]
We also assume that the parameters are feasible throughout this section.

\begin{definition}\label{def:p-alpha}
For each $\alpha \in A$, define the function $\rp_\alpha: M \to \mathbb{R}$ by
\[
\rp_\alpha(q) := \sum_{j\in V_\alpha} \frac{\rp(\|q-X_j\|)}{|V_\alpha|}.
\]
\end{definition}

The lemma below describes a useful regularity property of the function $\rp_\alpha$, which we are going to use in Section~\ref
{sec: orthogonalClusters} and \ref{sec: next-cluster}.

\begin{lemma}[Regularity of $\rp_{\alpha}$] \label{lem:rpalphaRegular}
Fix an index $\alpha \in A$ and assume the event $\Eclu((V_\alpha, i_\alpha))$.
Suppose $q\in M$ satisfies $0.9r \le \|q-X_{i_\alpha}\| \le 2r$. Suppose that $\sigma \in \{-1, +1\}$, $16 \eta/r \le \tau \le 1$, and $u \in \mathbb{S}^{N-1} = \left\{ x \in \mathbb{R}^N \, : \, \|x\| = 1 \right\}$ satisfy
\[
\Big\langle u, \sigma \frac{q-X_{i_\alpha}}{\|q-X_{i_\alpha}\|} \Big\rangle  = \tau,
\]
where $\langle \cdot, \cdot \rangle$ denotes the usual Euclidean inner product in $\mathbb{R}^N$. Then we have the following items.
\begin{itemize}
\item[(a)] For every $i \in V_\alpha$,
\[
\frac{\tau}{2} \le \Big\langle u, \sigma \frac{q-X_i}{\|q-X_i\|}\Big\rangle \le \frac{3}{2}\tau.
\]
\item[(b)] If $\sigma = +1$, then for any $0 \le t < 0.1 \tau r$,
\[
\rp_\alpha(q) - 2L_\rp \tau t \le \rp_\alpha(q +tu) \le \rp_\alpha(q) - \frac{1}{4} \ell_\rp \tau t.
\]
\item[(c)] If $\sigma = -1$, then for any $0 \le t < 0.1 \tau r$,
\[
\rp_\alpha(q) + \frac{1}{4} \ell_\rp \tau t \le \rp_\alpha(q +tu) \le \rp_\alpha(q) + 2L_\rp \tau t.
\]
\end{itemize}
\end{lemma}

\begin{proof}
{\bf (a)} Take an arbitrary $i \in V_\alpha$. First, since
$\|X_i - X_{i_\alpha}\| < \eta$, we have
\begin{align}
    \Big|\langle u, \sigma(q-X_i)\rangle - \tau\|q - X_{i_\alpha}\|\Big|
& =
    \Big|
     \langle u,
        \sigma (X_{i_\alpha} - X_i)
     \rangle
    +
     \langle u,
         \sigma (q-X_{i_\alpha})
     \rangle
    - \tau \|q-X_{i_\alpha}\|
    \Big| \notag \\
& =
    \big| \langle u, \sigma( X_{i_\alpha} - X_i) \rangle \big|
<
    \eta. \label{eq: less-than-eta}
\end{align}

Together with $\Big|\|q-X_{i_\alpha}\| - \|q-X_i\| \Big| \le \| X_i - X_{i_\alpha} \| < \eta$ (the triangle inequality), we find
\begin{align*}
\bigg|\Big\langle u,  \sigma \frac{q- X_i}{\|q-X_i\|}\Big\rangle - \tau \bigg|
& =
     \bigg| \Big\langle u,
       \sigma \frac{q- X_i}{\|q-X_i\|} \Big\rangle
      - \tau \frac{\|q-X_{i_\alpha}\| - \|q-X_{i_\alpha}\|+\|q-X_i\|}{\|q-X_i\|} \bigg|\\
& \le
     \bigg| \Big\langle u,  \sigma \frac{q- X_i}{\|q-X_i\|} \Big\rangle - \tau \frac{\|q-X_{i_\alpha}\|}{\|q-X_i\|} \bigg|
+    \frac{\eta}{\|q-X_i\|}\\
& \overset{\eqref{eq: less-than-eta}}{\le}
    \frac{2\eta}{\|q-X_i\|}
\le
    \frac{2\eta}{\|q-X_{i_\alpha}\| - \eta}
\le
    \frac{4 \eta}{r},
\end{align*}
where the last inequality follows from a coarse estimate $\eta \le \frac{r}{\cdr} \le 0.1r$ and the assumption $\|q-X_{i_\alpha}\|\ge 0.9r$.
With the assumption that $\tau \ge 16 \eta/r$, part~(a) of the lemma follows.

\smallskip

{\bf (b)} Now we are ready to prove the second statement.
To bound
\[
\rp_\alpha(q+tu) = \sum_{i \in V_\alpha} \frac{\rp(\|q+tu-X_i\|)}{|V_\alpha|},
\]
we will need to estimate $\|q+tu-X_i\|$ for $i \in V_\alpha$. From now on, let us fix $i \in V_\alpha$.

For $t \ge 0$, using part~(a), we can show
\begin{align*}
\|q+tu -X_i\|
\ge
  \Big\langle q+tu -X_i, \frac{q-X_i}{\|q-X_i\|} \Big\rangle
=
  \|q-X_i\| + t \Big\langle u, \frac{q-X_i}{\|q-X_i\|} \Big\rangle
\ge
  \|q-X_i\| + \frac{\tau}{2}t.
 \end{align*}

Imposing the condition that $t < 0.1\tau r$, we can also derive an upper bound for $\|q+tu-X_i\|$ by a similar approach.
\begin{align}
\nonumber
 \|q+tu -X_i\| & \le \sqrt{ \Big\langle q+tu -X_i, \frac{q-X_i}{\|q-X_i\|}\Big\rangle^2 + t^2}\\
 \nonumber
& \le \sqrt{\Big(\|q-X_i\| + t\frac{3\tau}{2}\Big)^2+t^2}\\
 \nonumber
& = \Big(\|q-X_i\| + t\frac{3\tau}{2}\Big) \sqrt{1 + \frac{t^2}{(\|q-X_i\| + t\frac{3\tau}{2})^2}}\\
\nonumber
& \le \Big(\|q-X_i\| + t\frac{3\tau}{2}\Big) \Big(1 + \frac{t^2}{(\|q-X_i\| + t\frac{3\tau}{2})^2}\Big)\\
& \le
\|q-X_i\| + t\frac{3\tau}{2} + \frac{t^2}{\|q-X_i\|}.
\label{eq: rpalphaRegular00}
\end{align}

Moreover, by the assumption $\|q-X_{i_\alpha}\|\ge 0.9r$,
\begin{align*}
  \frac{t}{\|q-X_i\|}
\le
  \frac{0.1\tau \cdot r}{\|q-X_{i_\alpha}\|- \| X_{i_\alpha} -X_i\|}
\le
  \frac{0.1 \tau \cdot r}{ 0.9r - \eta }
\le
  \frac{0.1 \tau \cdot r}{ r/2 }
=
   0.2 \tau.
\end{align*}

Substituting the above estimate into \eqref{eq: rpalphaRegular00}, we get $\|q+tu -X_i\| \le \|q-X_i\| + 1.7 t \tau$. To summarize, we have obtained
\begin{equation}\label{eq: summary-0517}
\|q - X_i\| + 0.5 \tau t
\le \|q + tu -X_i\|
\le \|q - X_i\| + 1.7 \tau t.
\end{equation}
Since the above expression is valid for every $i \in V_\alpha$, we have
\begin{align*}
\rp_\alpha(q+ tu)
& = \sum_{i \in V_\alpha} \rp(\| q+tu-X_i\|)/|V_\alpha| \\
& \overset{\eqref{eq: summary-0517}}{\le} \sum_{i \in V_\alpha}\rp\big( \|q-X_i\| + 0.5 \tau t \big)/|V_\alpha| \\
& \le \sum_{i \in V_\alpha} \rp(\|q-X_i\|)/|V_\alpha| - 0.5 \ell_\rp \tau t \\
& = \rp_\alpha(q) - 0.5 \ell_\rp \tau t,
\end{align*}
where the last inequality is valid since by the upper bound $\|q-X_{i_\alpha}\|\le 2r$, the parameter assumptions, and the feasibility assumption, we have
\[
\|q-X_i\| + 0.5 \tau t
\le  \|q-X_{i_\alpha}\| + \eta + (0.5 \tau)(0.1 \tau r)
\le  2r + \eta + r
\le r_M.
\]
For the lower bound of $\rp_{\alpha}(q+tu)$, we use the Lipschitz constant of $\rp$:
\begin{align*}
\rp_{\alpha}(q+tu)
&= \sum_{i \in V_{\alpha}} \rp(\| q + tu - X_i\|)/|V_{\alpha}| \\
&\overset{\eqref{eq: summary-0517}}{\ge} \sum_{i \in V_{\alpha}} \rp\big(\| q - X_i\| + 1.7 \tau t \big)/|V_{\alpha}| \\
& \ge \sum_{i \in V_{\alpha}} \rp(\| q - X_i\|)/|V_{\alpha}| - 1.7 L_{\rp} \tau t \\
&= \rp_{\alpha}(q) - 1.7 L_{\rp} \tau t.
\end{align*}
As a consequence, we have obtained a slightly stronger estimate than the one stated in the lemma.

\smallskip

{\bf (c)} In the case $\sigma = -1$, we can estimate in the same way to get
\begin{align*}
   \|q-X_i\| - \frac{1}{4}\tau t
    \ge \|q+tu -X_i\| \ge \|q-X_i\| -\frac{3}{2}\tau t,
\end{align*}
for any $i \in V_\alpha$. The statement for this case follows by applying the same estimate for $\rp_\alpha(q+tu)$.
\end{proof}

\section{Forming almost orthogonal clusters}
\label{sec: orthogonalClusters}
Now that under appropriate assumptions we have managed to find one cluster, let us find more. Suppose we start with one cluster $(V_0, i_0)$. In this section, we would like to find $d$ more clusters, $(V_1, i_1), \ldots, (V_d, i_d)$ so that the list $X_{i_0}, X_{i_1}, \ldots, X_{i_d}$ is ``almost orthogonal'' in the sense that the $d$ vectors $\{X_{i_\alpha} - X_{i_0} \, : \, \alpha \in [d]\}$ are geometrically close to an orthogonal frame of vectors in $\mathbb{R}^d$ and the size of each vector in the collection is close to $r$ (see Definition~\ref{def:ao-event} below).

Suppose that for some $0 \le k \le d-1$, we have found clusters $(V_0, i_0), \ldots, (V_k, i_k)$ such that $X_{i_0}, \ldots, X_{i_k}$ are almost orthogonal. With a fresh batch $W \subseteq {\bf V}$ of $n$ vertices---fresh meaning we have not used any vertex in it before in previous analysis---we find a new cluster $(V_{k+1}, i_{k+1})$ such that $X_{i_0}, \ldots, X_{i_{k+1}}$ is almost orthogonal.

Our plan to find the new cluster is as follows. We define a subset $W' \subseteq W$ with the property that for any $i \in W'$, the point $X_i$ would make the list $X_{i_0}, X_{i_1}, \ldots, X_{i_k}, X_i$ an almost orthogonal list (see Proposition~\ref{prop:X_i-ao}). Then we show that $(W,W')$ is a good pair, and hence we can find a desired cluster inside $W'$, using the good pair lemma (Lemma~\ref{lem: good-pair-gives-cluster}).

\begin{definition}\label{def:ao-event}
For a non-negative inteer $k$ and $i_0,\dots, i_k \in {\bf V}$, we define the \defn{almost-orthogonal event}
\begin{align} \label{eq: ao-event}
    {\cal E}_{\rm ao}(i_0, \ldots, i_k)
= &
    \bigg\{
        \forall \alpha \in [k]\,, \Big| \|X_{i_\alpha} - X_{i_0}\| - r \Big| \le \delta \quad \mbox{and}  \quad
    \forall \{\alpha, \beta\} \in {[k] \choose 2},\,
        \Big| \|X_{i_\alpha}-X_{i_\beta}\| - \sqrt{2}r \Big| \le \delta
    \bigg\}\,.
\end{align}
We also extend the definition to ${\cal E}_{\rm ao}(i_0)$, which is the trial event.
\end{definition}

Also, let us recall the navigation event introduced in Definition \ref{def:navi}:
Suppose that $A$ is an index set. For a collection $\{V_\alpha\}_{\alpha \in A}$ of nonempty subsets of ${\bf V}$ and for a subset $W \subseteq {\bf V}$, the \defn{navigation event} of $(\{V_\alpha\}_{\alpha \in A}, W)$ is
\[
\Enavi(\{V_\alpha\}_{\alpha \in A}, W) := \bigg\{
\forall i \in W,
\forall \alpha \in A,
\bigg| \frac{|N(i) \cap V_\alpha|}{ |V_\alpha|} -
\sum_{j\in V_\alpha} \frac{\rp(\|X_i-X_j\|)}{|V_\alpha|}
\bigg| \le n^{-1/2 + \varsigma}
\bigg\}.
\]

\bigskip

Throughout this section, we fix $k \in [0,d-1]$. Consider a collection of pairs $(V_\alpha, i_\alpha)$ with $i_\alpha \in V_\alpha \in {\bf V}$ for $\alpha \in [0,k]$, and a new batch $W$ of vertices:
\[
W \subseteq {\bf V} \setminus \bigcup_{\alpha = 0}^k V_\alpha,
\]
such that $|W| = n$.

The discussion in this section is within the event
\begin{align}
\nonumber
\Eao &= \Eao(\{(V_\alpha, i_\alpha)\}_{\alpha \in [0,k]},W) \\
&:=
{\cal E}_{\rm ao}(i_0, \ldots, i_k) \cap \Eclu(\{(V_\alpha, i_\alpha)\}_{\alpha \in [0,k]}) \cap \Enavi(\{V_\alpha\}_{\alpha \in [0,k]}, W) \cap \Ecn(W) \cap \Enet(W).
\label{eq: Eao}
\end{align}

Consider the subset $W' \subseteq W$ defined as
\begin{align}
\label{eq: W'orthogonal}
 W' := \Big\{ i \in W\,:\,  \forall \alpha \in [k], \, \, & \rp(\sqrt{2}r+0.95\delta) \le |N(i) \cap V_\alpha|/|V_\alpha| \le \rp( \sqrt{2}r-0.95\delta)\\
\nonumber
& \text{and} \quad \rp(r+0.95\delta)  \le |N(i) \cap V_0|/ |V_0| \le \rp( r-0.95\delta)
\Big\}.
\end{align}

Recall that $N(i)$ denotes the set of neighbors of $i$ in ${\bf V}$. Thus, $N(i) \cap V_{\alpha}$ is the same set as $N_{V_{\alpha}}(i)$.

Consider that given a collection of vertex sets $\{V_\alpha\}_{\alpha \in [0,k]}$ and a vertex set $W$, we can determine the vertex set $W'$ by examining the structure of graph $G$. The primary objective of this section is to demonstrate that, conditioned on the event $\Eao$, it is possible to identify a specific cluster $(\gc{W}, \gc{i})$ within $W'$. Furthermore, the index $\gc{i}$, in conjunction with the sequence of indices $(i_0,\ldots, i_k)$, will be shown to satisfy the event ${\cal E}_{\rm ao}(i_0,\ldots, i_k, \gc{i})$. Here is the main objective in this section:

\begin{proposition}
\label{prop:nextnavicluster}
For $k \in [0,d-1]$, let $(V_0,i_0),\dots, (V_k,i_k)$ be $k+1$ pairs with $i_\alpha \in V_\alpha \subseteq {\bf V}$ and $W \subseteq {\bf V}$ be a subset of size $|W| = n$ which is disjoint from $\cup_{\alpha \in [0,k]} V_\alpha$. Let $W'$ be the set defined in \eqref{eq: W'orthogonal}.
Given the occurrence of the event $\Eao(\{(V_\alpha, i_\alpha)\},W)$ as described in \eqref{eq: Eao}.
  If we run the Algorithm~\gclu~ with input $(W,W')$, then the output $(\gc W, \gc i)$ is a cluster and $(i_0,\ldots,i_k, \gc i)$ satisfies the event ${\cal E}_{\rm ao}(i_0,\ldots,i_k, \gc i)$.
  In other words,
  \begin{align*}
        \Eao(\{(V_\alpha, i_\alpha)\},W)
    \subseteq
        \Eclu(\gc W, \gc i)
        \cap {\cal E}_{\rm ao}(i_0,\ldots,i_k, \gc i).
  \end{align*}
\end{proposition}

In this section, we will first prove Proposition~\ref{prop:nextnavicluster} and then prove the corollary.
Before we proceed, let us introduce additional notations.

For brevity, let us introduce some additional notations to be used in this section. Without loss of generality, let us assume
\[
    X_{i_0} = \vec{0} \in \mathbb{R}^N
\]
and recall that
\begin{align} \label{eq: AO-H0}
 \TM = T_{X_{i_0}}M.
\end{align}
Let
\[
P : \mathbb{R}^N \to \TM
\]
be the orthogonal projection to $\TM$. For each point $q\in M$, we write
\[
q = q^\top + q^\perp,
\]
where $q^\top = P q$ and $q^\perp =  q - q^\top$.
We apply Proposition \ref{prop: SecondFundamentalForm} with $p = \vec{0}$, $H=H_0$, and $\zeta = 1$.
Together with $r_M \le 0.01 \kappa$ from its definition, there exists  a local inverse of $P$
\begin{align*}
 \phi :  B(\vec{0}, r_M) \cap \TM \to M
\end{align*}
 such that
\begin{align}
    & P\circ \phi = {\rm Id}_{B(\vec{0}, r_M) \cap \TM}   \label{eq: phiID}, \\
    & B(\vec{0},r_M) \cap M \subseteq \phi(B(\vec{0},r_M) \cap \TM), \label{eq: phiSubset} \\
    & \forall x \in B(\vec{0},r_M) \cap \TM, \, \|\phi(x) -x \| \le \kappa \|x\|^2, \mbox{ and } \label{eq: phiNorm} \\
    & \forall x \in B(\vec{0},r_M) \cap \TM, \, \sd_{\rm gd}(\phi(x), q_0) \le \|x\|(1+ \kappa^2 \|x\|^2 /2), \label{eq: phiGeodesic}
\end{align}
where in \eqref{eq: phiGeodesic}, $\sd_{\rm gd}(x,y)$ is the geodesic distance between two points $x,y \in M$. We refer the reader to Appendix~\ref{sec: ERM} for details.

\subsection{$W'$ is nonempty}
Since our objective is to extract a cluster from $W'$ by applying Algorithm~\gclu~ to $(W,W')$,
in this subsection we will first show $W' \neq \varnothing$, which is one requirement of a good pair (see Definition~\ref{def:good-pair})

\begin{lemma}[Existence of a navigation point in $M$]\label{lem: ENP}
Assuming the event ${\cal E}_{\rm ao}(i_0,i_1,\dots,i_\alpha)$.
There exists $p\in M$ such that
\begin{align*}
   \forall \alpha \in [k],\,  \Big| \|p-X_{i_\alpha}\|-\sqrt{2}r \Big| \le 0.9\delta \qquad \mbox{and} \qquad \Big| \|p-X_{i_0}\|-r \Big| \le 0.9\delta.
\end{align*}
\end{lemma}

 \begin{proof}
Let $p' \in \TM$ be a point with $\|p'\|=\|p'-X_{i_0}\|=r$ which is orthogonal to $X_{i_\alpha}^\top = PX_{i_\alpha}$ for every $\alpha \in [k]$. Such a point $p'$ exists because $k < d= {\rm dim}(\TM)$. Since $p' \in B(\vec{0},r_M)\cap \TM$ is in the domain of $\phi$, we can set $p = \phi(p') \in M$. Notice that $p^\top = Pp = P\phi(p')=p'$.

Fix $\alpha \in [k]$. Our first step is to compare $\|X_{i_\alpha} - p\|$ and  $\|X_{i_\alpha}^\top - p^\top\|$. We have
\begin{align*}
 \|X_{i_\alpha}^\top - p^\top\|  \le
  \|X_{i_\alpha} - p\| \le
    \|X_{i_\alpha}^\top - p^\top\|
    +  \|p - p^\top\| + \|X_{i_\alpha} - X_{i_\alpha}^\top\|,
\end{align*}
where the first inequality relies on the fact that $P$ is a contraction, and the second inequality follows from the triangle inequality. Let us first simplify the right-hand side of the above expression.
For the second summand, using \eqref{eq: phiNorm}, we have
\begin{align}
\label{eq:ENP04}
    \|p - p^\top\|
\le
    \kappa \|p^\top\|^2
=
    \kappa r^2
\le
    \frac{0.01}{r_M}  r^2
\le
    0.01 \frac{1}{\cdr d^2} r
    2^{-4} \delta,
\end{align}
where we relied on $r_M \le 0.01/\kappa$, the parameter assumptions, and the feasibility assumption. The third summand can be bounded similarly. Within the event ${\cal E}_{\rm ao}(i_0,i_1,\dots, i_\alpha)$,
$$ X_{i_\alpha}  \in B(\vec{0},r_M) \cap M \subseteq \phi(B(\vec{0},r_M)\cap T_{q_0}M),$$
and together with \eqref{eq: phiID} we have $\phi \circ P(X_{i_\alpha}) = X_{i_\alpha}$.
Now repeat the same derivation as in \eqref{eq:ENP04},
\begin{align}
\label{eq:ENP00}
 \|X_{i_\alpha}^\perp\| = \|\phi(X_{i_\alpha}^\top) - X_{i_\alpha}^\top\|
\le \kappa \|X_{i_\alpha}^\top\|^2
\le \kappa \|X_{i_\alpha}\|^2 \le
\kappa (2r)^2
\le \frac{r}{2^4\cdr d^2} = 2^{-4}\delta.
\end{align}

Together we conclude that
\begin{align}
\label{eq:ENP01}
 \|X_{i_\alpha}^\top - p^\top\|  \le
  \|X_{i_\alpha} - p\| \le
    \|X_{i_\alpha}^\top - p^\top\|
      +  2^{-3}\delta.
\end{align}
The second step is to estimate $ \|X_{i_\alpha}^\top - p^\top\|$.
Let us begin with the upper estimate.
Within the event ${\cal E}_{\rm ao}(i_0,i_1,\dots, i_k)$,  we have
\begin{align*}
   \|X_{i_\alpha}^\top \|
\le
    \|X_{i_\alpha}\|
\le
    r+\delta,
\end{align*}
and by taking $\langle X_{i_\alpha}^\top, p^\top \rangle = 0$ into account, we obtain
\begin{align*}
    \|X_{i_\alpha}^\top -p^\top\|
=
    \sqrt{ \|X_{i_\alpha}^\top \|^2 + \|p^\top\|^2}
\le
    \sqrt{(r+\delta)^2 + r^2}
\le
    \sqrt{2}r + \frac{1}{\sqrt{2}}\delta  + \frac{1}{4}\frac{\delta^2}{ r }.
\end{align*}

Recalling $\delta = \frac{r}{\cdr d^2}$, we have
\begin{align}
\label{eq:ENP02}
     \|X_{i_\alpha}^\top -p^\top\|
\le
    \sqrt{2}r + \frac{1}{\sqrt{2}}(1+ 2^{-6}) \delta.
\end{align}
To derive a lower estimate of $\|X_{i_\alpha}^\top -p^\top\|$, we recycle the estimate of $\|X_{i_\alpha}^\perp\|$ from \eqref{eq:ENP00} to get
\begin{align*}
    \|X_{i_\alpha}^\top\|  \ge \|X_{i_\alpha} \| - \|X_{i_\alpha}^\perp\| \ge r-\delta - 2^{-4}\delta,
\end{align*}
and hence
\begin{align}
\|X_{i_\alpha}^\top -p^\top\|
\ge \sqrt{ (r - (1 + 2^{-4})\delta)^2 +r^2}
\ge \sqrt{2}r - (1+2^{-4})\delta /\sqrt{2}
\ge \sqrt{2}r - 0.9\delta.
\label{eq:ENP03}
\end{align}
By combining \eqref{eq:ENP01}, \eqref{eq:ENP02}, and \eqref{eq:ENP03}, we obtain
\begin{align*}
   \sqrt{2}r - 0.9\delta \le \|X_{i_\alpha} - p\| \le   \sqrt{2}r + 0.9\delta.
\end{align*}

It remains to derive the statement for $\alpha=0$, which follows immediately from our previous estimate of $\|p-p^\top\|$ in \eqref{eq:ENP04}:
\[
\Big|\| X_{i_0} -p\| -r \Big|
=
\Big| \|X_{i_0} - p\| - \|X_{i_0}-p^\top\| \Big|
\le
\|p -p^\top \| \le 2^{-4}\delta.
\]
The proof is completed.
\end{proof}

\begin{lemma} \label{lem: W'-size-lower-bound}
Assuming the events
${\cal E}_{\rm ao}(i_0, \ldots, i_k)$, $\Eclu(\{(V_\alpha, i_\alpha)\}_{\alpha \in [0,k]})$, $\Enavi(\{V_\alpha\}_{\alpha \in [0,k]}, W)$, and $\Enet(W)$.
We have $|W'| \ge n^{1-\varsigma}$.
\end{lemma}

\begin{proof}
With the occurrence of event ${\cal E}_{\rm ao}(i_0,\ldots, i_k)$, we can pick a point $p\in M$ described in Lemma~\ref{lem: ENP}.
Within the event $\Enet(W)$, the set
\[
W'' := \{i \in W\,:\, \|X_i-p\| < \varepsilon\}\]
satisfies $|W''|\ge n^{1-\varsigma}$. Thus it suffices to show that $W'' \subseteq W'$.

Take an arbitrary $i \in W''$. For any $\alpha \in [k]$, the event $\Enavi(\{(V_\alpha \}_{\alpha \in [0,k]},W)$ guarantees that
\begin{equation}\label{eq: using-E-navi}
\frac{|N(i) \cap V_{\alpha}|}{|V_{\alpha}|} \ge p_\alpha(X_i) - n^{-1/2 + \varsigma}.
\end{equation}
For each $j \in V_\alpha$, within the event $\Eclu(\{(V_\alpha,i_\alpha)\}_{\alpha \in [0,k]})$ we have
\begin{align*}
    \| X_i - X_j \|
&\le
    \| X_i - p \| + \| p - X_{i_\alpha} \| + \| X_{i_\alpha} - X_j \|  \\
& \phantom{AAA AAA AAA}\le
    \varepsilon + \sqrt{2}r + 0.9 \delta + \varepsilon
<
    \sqrt{2} r + 0.91 \delta.
\end{align*}

Therefore, $\rp(\|X_i - X_j\|) > \rp(\sqrt{2} r + 0.91 \delta)$. Using this in \eqref{eq: using-E-navi} yields
\begin{equation}\label{eq: N-i-V-alpha-0.91}
    \frac{|N(i) \cap V_{\alpha}|}{|V_{\alpha}|}
\ge
    \rp_\alpha(X_i) - n^{-1/2+\varsigma}
> \rp(\sqrt{2} r + 0.91 \delta) - n^{-1/2 + \varsigma}.
\end{equation}
By the feasibility assumption, we have
\[
0 < \sqrt{2} r + 0.91 \delta < \sqrt{2} r + 0.95 \delta < r_M,
\]
and therefore,
\[
\rp(\sqrt{2} r + 0.91 \delta) - \rp(\sqrt{2} r + 0.95 \delta) \ge \ell_{\rp} \cdot 0.04 \delta > n^{-1/2 + \varsigma}.
\]
Using the above inequality in \eqref{eq: N-i-V-alpha-0.91} yields
\[
\frac{|N(i) \cap V_{\alpha}|}{|V_{\alpha}|} > \rp(\sqrt{2} r + 0.95 \delta).
\]
In a similar fashion, we can show
\begin{align*}
& \frac{|N(i) \cap V_{\alpha}|}{|V_{\alpha}|} < \rp(\sqrt{2} r - 0.95 \delta), \\
& \frac{|N(i) \cap V_0|}{|V_0|} > \rp( r + 0.95 \delta), \qquad \text{and} \\
& \frac{|N(i) \cap V_0|}{|V_0|} < \rp( r - 0.95 \delta).
\end{align*}
This shows that $i \in W'$. Since $i$ was arbitrarily chosen, we conclude that $W'' \subseteq W'$, and hence $|W'| \ge n^{1-\varsigma}$.
\end{proof}

\subsection{Existence of a point $p'$}

In this subsection, we show that for any $i \in W'$, there exists $p' \in M$ such that $\|p'-X_i\|$ is small and $\{j \in W : X_j \in B(p',\varepsilon)\} \subseteq W'$. This is the key condition in the definition of a good pair (See Definition \ref{def:good-pair}).

To demonstrate the proof idea, let us make a false assumption for now that the following terms are the equal with no gaps between them:
$$
 \frac{|N(i)\cap V_\alpha|}{|V_\alpha|} = \rp_\alpha(i) = \rp(\|X_i - X_{i_\alpha}\|)
$$
 for every $i\in W$ and $\alpha \in [0,k]$.
 Consider $i \in W'$. In this case, $\|X_i - X_{i_\alpha}\|$ falls within the interval
$(\sqrt{2}r - 0.95\delta, \sqrt{2}r + 0.95\delta)$ for $\alpha \in [k]$, and similarly for $\alpha =0$  but with $\sqrt{2}r$ replaced by $r$.

Solving a linear equation allows us to identify a small vector $u$ with its magnitude proportional to $\varepsilon$ (also depending on $d$) within  span$(\{X_i - X_{i_\alpha}\}_{\alpha \in [0,k]})$ such that $ \|X_i + u -X_{i_\alpha}\|$ is contained in a narrower interval
$$
    ( \sqrt{2}r - 0.95\delta + C\varepsilon , \sqrt{2}r + 0.95\delta - C\varepsilon)$$
for each $\alpha \in [k]$ (and similarly for $\alpha = 0$ with $\sqrt{2}r$ replaced by $r$), where $C\ge 1$ is a constant greater than $1$ to absorb some error terms.

Now, for every $j \in W$ with $X_j \in B(X_i +u, \varepsilon)$, the triangle inequality implies
$\|X_j - X_{i_\alpha}\|$
lies within
$$
    ( \sqrt{2}r - 0.95\delta + C\varepsilon - \varepsilon, \sqrt{2}r + 0.95\delta - C\varepsilon + \varepsilon)
$$
for $\alpha \in [k]$ (and the same for $\alpha=0$ with $\sqrt{2}r$ replaced by $r$). This in turn implies that $ j \in W'$, which is our desired property for $p'$. However, there is a complication: the point $X_i + u$ does not necessarily lie in $M$. To address that, we will to find $u^*$, perpendicular to $X_i-X_{i_\alpha}$ for $\alpha \in [0,k]$, ensuring $X_i + u + u^* \in M$. If the span of $\{ X_i - X_{i_\alpha} \}_{\alpha \in [0,k]}$ aligns closely with $T_{X_i}M$, then $u^*$ can be chosen with a magnitude proportional to that of $u$. Furthermore, because $u^*$ is orthogonal to  $\{ X_i - X_{i_\alpha} \}_{\alpha \in [0,k]}$ and because $X_i+u-X_{i_\alpha}$ has a significantly larger magnitude than $u^*$, one can show
$$
    \|X_i + u + u^* - X_{i_\alpha}\| - \|X_i + u - X_{i_\alpha}\|
$$
is of order proportional to $\|u^*\|^2 = O(\varepsilon^2)$, which has negligible impact on the argument above. Therefore, one can take $p' = X_i+u+u^*$.

The actual argument, while based on the simple idea above, must account for the gaps difference between $\frac{|N(i)\cap V_\alpha|}{|V_\alpha|}$, $\rp_\alpha(X_i)$, and $\rp(X_i - X_{i_\alpha})$. Notably, the gap of $\rp_{\alpha}(X_i)$ and $\rp(\|X_i-X_{i_\alpha}\|)$ can be proportionally related to $\eta$, which is significantly larger than $\|u\|$.
If we were to na\"ively extend the above argument and attempt to treat the gap between $|N(i)\cap V_\alpha|/ |V_\alpha|$ and $\rp(\|X_i - X_{i_\alpha}\|)$ merely as error terms, then we could only construct such $u$ with $\|u\| = O(\eta)$, a size proportional the radius of the current clusters. This approach leads to a geometrically increasing radius for the clusters subsequently generated, which is not desirable. Hence, to attain the required level of preciseness, we will work directly on $\rp_{\alpha}$, which relies on its regularity property established in Lemma \ref{lem:rpalphaRegular}.

Let us begin with a distance estimate $\|X_i-X_{i_\alpha}\|$ for $i \in W'$.

\begin{proposition}\label{prop:X_i-ao}
Assume that the event $\Eclu(\{(V_\alpha,i_\alpha)\}_{\alpha \in A})$ occurs.
For any $i \in W'$, the point $X_i \in M$ satisfies
\begin{align} \label{eq: palphaNorm}
  \forall \alpha \in [k], \,\,\,\,
  \sqrt{2}r - \delta \le \| X_i - X_{i_\alpha}\| \le \sqrt{2}r + \delta
  \qquad \mbox{and} \qquad
   r - \delta \le \| X_i - X_{i_0} \| \le r + \delta.
\end{align}
\end{proposition}

\begin{proof}
Given the occurrence of event $\Eclu(\{(V_\alpha,i_\alpha)\}_{\alpha \in A})$ and based on the definition $W$, it follows that for $\alpha \in [k]$,
\begin{equation}\label{eq:R_1-1}
\rp(\sqrt{2}r+0.95\delta ) - n^{-1/2+\varsigma}  \le \rp_\alpha(X_i) \le   \rp(\sqrt{2}r-0.95\delta ) +  n^{-1/2+\varsigma}
\end{equation}
and
\begin{equation}\label{eq:R_1-2}
\rp(r+0.95\delta) -  n^{-1/2+\varsigma}
\le \rp_0(X_i)
\le   \rp(r-0.95\delta) +  n^{-1/2+\varsigma}.
\end{equation}

Notice that we can bound $\|X_i - X_{i_\alpha}\|$ from the assumption on $X_i$.
Within the event $\Eclu(\{V_\alpha,i_\alpha\}_{\alpha \in A})$, for $\alpha \in [k]$ we have
\begin{align*}
    \rp(\|X_i - X_{i_\alpha}\| + \eta )
\le
    \rp_\alpha(X_i)
\overset{\eqref{eq:R_1-1}}{\le}
    \rp(\sqrt{2}r-0.95\delta ) +  n^{-1/2+\varsigma}.
\end{align*}

Next, from the assumptions of the parameters,
we have
$$
    \delta
\ge
    \cdr \sqrt{d} \cdot \eta
\ge
    \frac{\cdr^2  L_{\rp}^2}{ \ell_{\rp}^2 }\varepsilon
\ge
    \frac{\cdr^2 L_{\rp} }{\sqrt{\|\rK\|_\infty} \ell_\rp^2}  n^{-1/2+\varsigma}
\ge
    \frac{100}{\ell_\rp} n^{-1/2+\varsigma},
$$
where the last inequality relies on the facts that $\frac{L_\rp}{\ell_\rp} \ge 1$ by definition, $\|\rK\|_\infty \le 1$, and $\cdr \ge 100$.
Now relying on the definition of $\ell_p$,
\begin{align*}
\rp(\sqrt{2}r -0.95\delta ) + n^{-1/2+\varsigma}
& \le
    \rp(\sqrt{2}r - 0.96\delta) - \ell_\rp \cdot 0.01 \delta + n^{-1/2+\varsigma}
 \le
    \rp(\sqrt{2}r-0.96\delta),
\end{align*}
and hence $\|X_i - q_\alpha\| \ge \sqrt{2}r -0.96\delta - \eta \ge \sqrt{2}r - \delta$.

With the same approach, from \eqref{eq:R_1-1} and \eqref{eq:R_1-2}, we can show, for each $\alpha \in [k]$,
\[
\sqrt{2}r - \delta \le \|X_i - q_\alpha\| \le \sqrt{2}r + \delta,
\]
and $r - \delta \le \|X_i - q_0\| \le r + \delta$.
\end{proof}

\subsubsection{Small angles between planes} \label{sec: SmallAnglesBtwnPlanes}
Here we will derive that ${\rm span}(\{ X_i - X_{i_\alpha} \}_{\alpha \in [0,k]})$ aligns closely with the tangent plane $T_{X_i}M$.
\begin{lemma}
\label{lem: Hangle}
Consider $\iW \in W'$.  We define a $N$ by $k+1$ matrix
$$
    Y = (Y_0,\ldots, Y_k)
$$
with column vectors $\{Y_\alpha\}_{\alpha \in [0,k]}$ defined as
\[
Y_\alpha := X_{\iW} - X_{i_\alpha},
\]
for each $\alpha \in [0,k]$.
Furthermore, we define two linear subspaces
\[
H_Y := {\rm span}\big( \{Y_\alpha\}_{\alpha \in [0,k]} \big) \subseteq \mathbb{R}^N
\quad
\mbox{and}
\quad
H_{\iW} := T_{X_{\iW}}M.
\]

Then, within the event $\Eao$, we have
$$\sd( \widetilde H_Y, H_{\iW}) \le 1/\cdr,$$
where $\widetilde H_Y$ is any $d$-dimensional subspace in $\R^N$ satisfying
$$
    H_Y \subseteq \widetilde H_Y \subseteq {\rm span}(H_Y, \TM \cap H_Y^\perp).
$$
\end{lemma}

The will be obtained by showing that both spaces form only a small angle with $\TM$.

\begin{lemma}
\label{lem:Hflat}
Consider $\iW \in W'$ and adapt the notations described in Lemma \ref{lem: Hangle}. Recall that $P$ is the orthogonal projection to $\TM$ and $u^\perp = u - Pu$ for every $u \in \R^N$.
Within the event $\Eao$, we have
$$
    \max_{u \in H_Y\,:\, \|u\|=1} \|u^\perp\| \le  \frac{1}{4\cdr}.
$$
\end{lemma}

Notice that
\begin{align}
\label{eq: Hflat00}
    \max_{u \in H_Y\,:\, \|u\|=1} \|u^\perp\|
=
    \max_{v \in \R^{[0,k]}: \|v\|=1}  \Big\| P_{\TM^\perp} \frac{Y v }{\|Yv\|}\Big\|
\le
        \frac{ \|P_{H_0^\perp}Y\|_{\rm op} }{{\rm s}_{\rm min}(Y)},
\end{align}
where $P_{\TM^\perp}$ is the orthogonal projection to $\TM^\perp$,
$Y$ is the $N$ by $k+1$ matrix with columns $(Y_0,Y_1,\dots,Y_k)$,
and ${\rm s}_{\rm min}(Y)$ is the least singular value of $Y$.
With $k+1\le d \le N$, we have
$$
    {\rm s}_{\rm min}(Y)
=
    \sqrt{ {\rm s}_{\rm min}(Y^\tp Y)},
$$
and $(Y^\tp Y)_{\alpha,\beta } =  \langle Y_\alpha , Y_\beta \rangle$ for $\alpha,\beta \in [0,k]$.

\begin{lemma}
\label{lem: YalphaYbeta}
   Consider $\iW \in W'$ and adapt the notations described in Lemma \ref{lem: Hangle}.
   Within the event $\Eao$, for $\alpha, \beta \in [0,k]$,
   \begin{align*}
        \big| \langle Y_\alpha, Y_\beta \rangle  -r^2 - r^2{\bf 1}\big( \alpha = \beta \neq 0) \big|
    \le
        15r\delta,
    \end{align*}
    where ${\bf 1}(\alpha = \beta \neq 0)$ is the indicator function which equals $1$ when $\alpha = \beta = s$ for some $s\in[k]$.
\end{lemma}

\begin{proof}
 By definition,
 \begin{align}
    \label{eq: sminB00}
      \langle Y_\alpha, Y_\beta \rangle
  =
      \big\langle X_{\iW} , X_{\iW} \big\rangle
      -   \big \langle X_{i_\alpha} , X_{\iW} \big\rangle
      -   \big \langle X_{i_\beta} , X_{\iW} \big\rangle
      +   \big \langle X_{i_\alpha}, X_{i_\beta} \big\rangle.
 \end{align}

 Now we will estimate the four terms on the right-hand side of the above expression. We start with the expression $\langle X_{i_\alpha}, X_{i_\beta} \rangle$. For distinct $\alpha, \beta \in [k]$, using our assumption $\big| \|X_{i_\alpha} \| - r \big\| \le \delta$
 and $\big| \|X_{i_\alpha} - X_{i_\beta} \| - \sqrt{2}r \big| \le \delta$ from event ${\cal E}_{\rm ao}(i_0, \ldots, i_k)$, we find
  \begin{align}
  \big| \big\langle X_{i_\alpha}, X_{i_\beta} \big\rangle \big| \nonumber
  & =
  \frac{1}{2} \Big|\|X_{i_\alpha}\|^2 + \|X_{i_\beta}\|^2 - \|X_{i_\alpha} - X_{i_\beta}\|^2\Big| \\
  \nonumber
  & \le
  \frac{1}{2} \Big(\Big|\|X_{i_\alpha}\|^2 -r^2\Big|
  + \Big|\|X_{i_\beta}\|^2 - r^2\Big|
  + \Big|\|X_{i_\alpha} - X_{i_\beta}\|^2 - 2r^2 \Big| \Big) \\
  \nonumber
  & \le
  \frac{1}{2} \big( 2r\delta +\delta^2  + 2r\delta + \delta^2 + 2\sqrt{2}r\delta + \delta^2) \\
  & < 4 r \delta.
  \label{eq: sminB01}
  \end{align}
  where the last inequality relies on the coarse bound $\delta^2 \le 0.1 r \delta$ which follows from the parameter assumptions. Similarly, when $\alpha=\beta \in [k]$, we have
  \begin{align*}
      \big | \big \langle X_{i_\alpha}, X_{i_\beta} \big \rangle - r^2 \big |
  =
      \Big| \|X_{i_\alpha}\|^2 - r^2 \Big| \le 2 r\delta +\delta^2 < 3r\delta.
  \end{align*}

And if either $\alpha=0$ or $\beta=0$, we simply have $\langle X_{i_\alpha},\,X_{i_\beta} \rangle = 0$ since $X_0 = \vec{0}$.

Now consider the second and the third terms on the right-hand side of \eqref{eq: sminB00}.
Within the event $\Eclu(\{(V_\alpha,i_\alpha)\}_{\alpha \in A})$, we can apply \eqref{eq: palphaNorm} from Proposition \ref{prop:X_i-ao} with $i = \iW$ to show that for $\alpha \in [k]$,
\begin{align*}
   \big | \big \langle X_{i_\alpha} , X_{\iW} \big \rangle \big |
& =
  \frac{1}{2} \Big|  \|X_{i_\alpha}\|^2 + \|X_{\iW}\|^2-\|X_{\iW}-X_{i_\alpha}\|^2 \Big| \\
& \le
    \frac{1}{2}
    \Big( \Big| \|X_{i_\alpha}\|^2 - r^2 \Big|
        + \Big| \|X_{\iW}-X_{i_0}\|^2 - r^2 \Big|
        + \Big| \|X_{\iW}-X_{i_\alpha}\|^2 - 2r^2 \Big|\Big)
< 4 r \delta,
\end{align*}
where the last inequality is carried out in the same way as that for \eqref{eq: sminB01}.
Since $X_{i_0} = \vec{0}$, the above expression also holds when $\alpha=0$.

For the last term on the right-hand side of \eqref{eq: sminB00}, it also follows from \eqref{eq: palphaNorm} that
$\big|\langle X_{\iW}, X_{\iW} \rangle -r^2 \big| < 3r\delta$. Now substituting all these estimates in \eqref{eq: sminB00}, the lemma follows.
\end{proof}

Relying on the estimates of $\langle Y_\alpha, Y_\beta\rangle$,
one can give a bound on ${\rm s}_{\rm min}(Y^\tp Y)$.

\begin{lemma} \label{lem: sminB}
    Suppose $B = (b_{\alpha,\beta})_{\alpha,\beta \in [0,k]}$
    is a $k+1$ by $k+1$ matrix  such that for $\alpha, \beta \in [0,k]$,
    $$ \big| b_{\alpha,\beta} - r^2 -  r^2 {\bf 1}(\alpha = \beta \neq 0) \big|  \le 15r\delta.$$
    We have
    $$
        {\rm s}_{\rm min}(B) \ge \frac{r^2}{4(k+1)}.
    $$

\end{lemma}
Given that this is simply a standard result in linear algebra. We postpone the proof to Appendix \ref{appx: basic result proof}.

\begin{remark}
    \label{rem: sminY}
   As a immediate consequence of Lemmas~\ref{lem: YalphaYbeta} and \ref{lem: sminB}, within the event $\Eao$ we have $${\rm s}_{\rm min}(Y) = \sqrt{ {\rm s}_{\rm min}(Y^\tp Y)} \ge \frac{r}{2\sqrt{k+1}}.$$
\end{remark}

Now we are ready to prove Lemma \ref{lem:Hflat}.
\begin{proof}[Proof of Lemma \ref{lem:Hflat}]
    By \eqref{eq: Hflat00}, we have
\begin{align}
\label{eq: Hflat01}
    \max_{u \in H_Y\,:\, \|u\|=1} \|u^\perp\|
\le
    \frac{\| P_{\TM^\top} Y \|_{\rm op} }{{\rm s}_{\rm min}(Y)}.
\end{align}
Let us first derive an upper bound for $\| P_{\TM^\top} Y \|_{\rm op}$.

Using $\|X_{\iW}\| = \|X_{\iW}-X_{i_0}\| \le r + \delta \le 2r <r_M$ from \eqref{eq: palphaNorm} and the feasibility assumption, we find that $X_{\iW}$ is contained in the image of $\phi$ by \eqref{eq: phiSubset}. With the properties of $\phi$ (see \eqref{eq: phiID} and \eqref{eq: phiNorm}) we obtain
\[
    \| P_{\TM} X_{\iW}\|
=
    \| X_{\iW}^\perp\|
\le
    \kappa \|X_{\iW}^\top\|^2
\le
    \kappa \|X_{\iW}\|^2
\le
    2^{-4} \delta,
\]
where we once again use $r_M \le 0.01/\kappa$. Furthermore, the same estimate also holds for $X_{i_\alpha}$, since $\|X_{i_\alpha} \|=\|X_{i_\alpha} -X_{i_0}\|\le 2r$. Hence, we conclude that
\begin{equation}\label{eq:p-alpha-perp}
\| Y_\alpha^\perp \|
= \| ( X_{\iW} - X_{i_\alpha})^\perp\|
\le \| X_{\iW}^\perp\| + \|X_{i_\alpha}^\perp\|
\le 2^{-3}\delta,
\end{equation}
for any $\alpha \in [0,k]$.

Therefore, we conclude that $P_{\TM^\perp}Y = (P_{\TM^\perp}Y_0,\ldots P_{\TM^\perp}Y_k)$ is a matrix with columns whose Euclidean norms are bounded by $2^{-3}\delta$, and hence
$$
    \| P_{\TM^\top} Y \|_{\rm op}
\le
    \sqrt{\sum_{\alpha \in [0,k]} \|P_{\TM^\perp}Y_\alpha\|^2}
\le
    2^{-3}\delta \sqrt{k+1}.
$$

Substitute the above this bound and ${\rm s}_{\rm min}(Y) \ge \frac{r}{2\sqrt{k+1}}$ from Remark \ref{rem: sminY} into \eqref{eq: Hflat01}, we conclude that
$$
    \max_{u \in H_Y\,:\, \|u\|=1} \|u^\perp\|
\le
    \frac{\delta(k+1)}{4r}
\le
    \frac{\delta d}{4r}
\le
    \frac{1}{4\cdr},
$$
where the last inequality follows from the parameter assumptions.
\end{proof}

\begin{proof}[Proof of Lemma \ref{lem: Hangle}]
Recall from \eqref{eq: palphaNorm} that $\|X_{\iW}-X_{i_0}\|\le r + \delta \le r_M \le 0.01/\kappa$. Hence, Corollary~\ref{cor:p-p'} gives
\begin{equation}\label{eq:sd2}
\sd(\TM, H_{\iW} ) \le 2\kappa \cdot \|q_0 - p\| \le 2\kappa \cdot 2r \le  \frac{0.04}{\cdr d^2} \le \frac{0.04}{\cdr}.
\end{equation}

Let $\widetilde H_Y$ be a $d$-dimensional subspace containing $H_Y$ and that is contained in the span of $H_Y$ and $\TM \cap H_Y^\perp$. By doing this, we have
\begin{equation}\label{eq:sd1}
    \sd(\widetilde H_Y, \TM)
\le \frac{1}{4\cdr}.
\end{equation}

Using the triangle inequality, \eqref{eq:sd1}, and \eqref{eq:sd2}, we obtain
\[
\sd( \widetilde H_Y, H_{\iW}) \le \sd(\widetilde{H}_Y, \TM) + \sd(\TM, H_{\iW}) \le \frac{0.25}{\cdr} + \frac{0.04}{\cdr} < \frac{1}{\cdr}.
\]

\end{proof}

\subsubsection{Existence of $p$}
\label{subsubs: Existencep}

The goal in Section \ref{subsubs: Existencep} is to establish the following proposition.
\begin{proposition}
\label{prop:orthogonal}
    With the occurrence of the event $\Eao$,
    for any $\iW \in W'$,
    there exists a point $p \in M$ satisfying $\|X_{\iW} - p\| \le 2^8 d \frac{L_\rp}{\ell_\rp} \varepsilon$
    so that the following holds:
    $$
        \{ j \in W \,:\, X_j \in B(p,\varepsilon) \} \subseteq W'.
    $$
\end{proposition}

Before we proceed, let us set up additional notations.
Here we fix $\iW \in W'$.
For $\alpha \in [k]$, let
\begin{align} \label{eq: sigmaAlpha}
\sigma_\alpha := \sign \big( \rp_\alpha(p) - \rp(\sqrt{2}r) \big)
\qquad \mbox{and} \qquad
\sigma_0 := \sign \big( \rp_0(p) - \rp(r) \big).
\end{align}

Here, the function $\sign: \mathbb{R} \to \{-1, +1\}$ is defined as
\begin{equation}\label{eq:defn-sign}
\sign(x) := \begin{cases}
    +1 & \text{if } x \ge 0, \\
    -1 & \text{if } x < 0,
\end{cases}
\end{equation}
for each $x \in \mathbb{R}$. In particular, by our convention, $\sign$ outputs either $-1$ or $+1$, but never $0$.

The key ingredient is the following lemma.

\begin{lemma} \label{lem:orthogonal}
With the occurrence of $\Eao$, for any $\iW \in W'$, there
exists a point $p \in M$ with $\|p-X_{\iW}\| \le 2^8d \frac{L_\rp}{\ell_\rp} \varepsilon$ such that  \[
\forall \alpha \in [k], \quad \rp(\sqrt{2}r+0.95 \delta ) + 2 L_{\rp} \varepsilon  \le \rp_\alpha(p) \le   \rp(\sqrt{2}r-0.95 \delta) - 2 L_{\rp} \varepsilon,
\]
and
\[
\rp(r+0.95 \delta  )  + 2 L_{\rp} \varepsilon
\le \rp_0(p)
\le   \rp(r-0.95 \delta ) - 2L_{\rp} \varepsilon.
\]
\end{lemma}

Given the above lemma, let us derive the proof of Proposition \ref{prop:orthogonal}.
\begin{proof}[Proof of Proposition \ref{prop:orthogonal}]

Suppose that $j \in W$ satisfies $\|X_j - p\| < \varepsilon$. Then by using the Lipschitz constant $L_{\rp}$ and the triangle inequality, we obtain, for each $\alpha \in [0,k]$,
\begin{align}
| \rp_{\alpha}(X_j) - \rp_{\alpha}(p) |
&\le \sum_{i \in V_{\alpha}} \frac{\left| \rp(\|X_j - X_i\|) - \rp(\|p-X_i\|) \right|}{|V_{\alpha}|} \notag \\
&\le \sum_{i \in V_{\alpha}} \frac{L_{\rp} \cdot \Big| \|X_j - X_i\| - \|p - X_i\| \Big|}{|V_{\alpha}|} \notag \\
&\le \sum_{i \in V_{\alpha}} \frac{L_{\rp} \cdot \|X_j - p\|}{|V_{\alpha}|} < L_{\rp} \varepsilon. \label{ineq:L-rp}
\end{align}

Next, $\Enavi$ and the parameter assumptions give for each $\alpha \in [0,k]$,
\begin{equation}\label{ineq:from-E-navi}
\bigg| \frac{|N(j) \cap V_\alpha|}{ |V_\alpha|} - \rp_\alpha(X_j) \bigg|
\le n^{-1/2 + \varsigma}
< L_{\rp} \varepsilon.
\end{equation}
Combining \eqref{ineq:L-rp}, \eqref{ineq:from-E-navi}, and the properties of $p$ from Lemma~\ref{lem:orthogonal}, we find that for each $\alpha \in [k]$,
\[
\rp(\sqrt{2} r + 0.95 \delta) \le \frac{|N(j) \cap V_{\alpha}|}{|V_{\alpha}|} \le \rp(\sqrt{2} r - 0.95 \delta),
\]
and
\[
\rp(r + 0.95 \delta) \le \frac{|N(j) \cap V_0|}{|V_0|} \le \rp(r - 0.95 \delta),
\]
whence $j \in W'$.
\end{proof}

\begin{proof}

Let $\widetilde H_Y$ and $H_{\iW}$ be the subspaces introduced in Lemma \ref{lem: Hangle}. With the occurrence the event $\Eao$, we can apply the Lemma to get
$$
   \sd(\widetilde H_Y, H_{\iW} ) \le \frac{1}{\cdr}.
$$
Observe that $X_{\iW} = X_{\iW} - X_{i_0} = Y_0 \in H_Y \subseteq \widetilde H_Y$.
By applying Proposition \ref{prop: SecondFundamentalForm} (with $X_{\iW}$, $\widetilde H_Y$, $P_{\widetilde H_Y}$, and $1-1/\cdr$ playing the roles of $p$, $H$, $P_H$, and $\zeta$ in the proposition, respectively), we find that there exists a local inverse map
\[
\phi_Y:  B\!\left(X_{\iW}, 0.09 (1 - 1/\cdr)^2 /\kappa \right) \cap \widetilde H_Y  \to M
\]
such that $P_{\widetilde H_Y}(\phi_Y(x)) =x$, for every point $x \in B\!\left(X_{\iW}, 0.09 (1 - 1/\cdr)^2 /\kappa \right) \cap \widetilde H_Y$. Since
\[
0.09 (1 - 1/\cdr)^2 /\kappa \ge  0.01/\kappa
\ge
r_M,
\]
we may restrict the domain of $\phi_Y$ to $B(X_{\iW}, r_M)  \cap \widetilde H_Y$. Furthermore, Proposition~\ref{prop: SecondFundamentalForm}(c) gives
\begin{align} \label{eq: orthogonal00}
    \| \phi_Y(x) - X_{\iW}\| \le 2\|x-X_{\iW}\|,
\end{align}
for any $x \in B(X_{\iW}, r_M) \cap \widetilde H_Y$.

Let $\varu \in H_Y$ be the vector satisfying
\begin{equation}\label{eq:def-u}
\forall \alpha \in [0,k], \quad
\Big\langle \sigma_\alpha \frac{Y_\alpha}{\|Y_\alpha\|}, \varu \Big\rangle = 12 \frac{L_{\rp}}{\ell_{\rp}} \varepsilon.
\end{equation}
Equivalently, $\varu \in H_Y$ is the vector so that $(Y^\tp \varu)_{\alpha} = \sigma_\alpha 12 \frac{L_{\rp}}{\ell_{\rp}} \varepsilon \|Y_\alpha\|$, for every $\alpha \in [0,k]$.

The existence (and uniqueness) of $\varu \in H_Y = {\rm Im}(Y)$ follows from Remark \ref{rem: sminY} that the matrix $Y$ has full rank. By Lemma~\ref{lem: sminB} and a coarse bound that $\|Y_\alpha\|\le 2r$ for $\alpha \in [0,k]$ (see \eqref{eq: palphaNorm}),
\begin{align}  \label{eq: orthogonal01}
\|\varu\| \le
&  \Big( \frac{r}{2\sqrt{k+1}}\Big)^{-1} \|Y^\tp \varu\|
\le
  \Big( \frac{r}{2\sqrt{k+1}}\Big)^{-1}
    \sqrt{k+1} \cdot 12\frac{L_\rp}{\ell_\rp}\varepsilon \cdot 2r
\le
  48d \frac{L_{\rp}}{\ell_{\rp}} \varepsilon.
\end{align}

We take $p \in M$ to be the following point:
$$p:= \phi_Y(X_{\iW}+\varu),$$
and claim that $p'$ satisfies the desired property.

To begin with, by \eqref{eq: orthogonal00} and \eqref{eq: orthogonal01}, $p$ is not too far from $X_{\iW}$:
\[
\|p - X_{\iW}\| \le 2\|\varu\| \le 96 d \frac{L_\rp}{\ell_\rp}\varepsilon < 2^8 d \frac{L_\rp}{\ell_\rp}\varepsilon.
\]

Next, note that from \eqref{eq:def-u}, since $12 \frac{L_{\rp}}{\ell_{\rp}} \varepsilon > 0$, we know that $\varu \neq \vec{0}$. Because $\phi_Y$ is a diffeomorphism, we then have that $p \neq X_{\iW}$. Hence, it makes sense to consider $u := \frac{p-X_{\iW}}{\|p-X_{\iW}\|}$. Fix $\alpha \in [0,k]$. Define
\[
\tau := \Big\langle u, \sigma_\alpha \frac{Y_\alpha}{\|Y_\alpha\|} \Big\rangle.
\]
Observe that
\begin{align}
\label{eq: orthogonal02}
\tau
= \Big\langle \frac{p - (X_{\iW} + \varu) + \varu}{\|p-X_{\iW}\|}, \sigma_\alpha \frac{ Y_\alpha}{\| Y_\alpha\|} \Big\rangle
= \Big\langle \frac{\varu}{\|p-X_{\iW}\|}, \sigma_\alpha \frac{ Y_\alpha}{\| Y_\alpha\|} \Big\rangle
\overset{\eqref{eq:def-u}}{=} \frac{1}{\|p-X_{\iW}\|} \cdot 12 \frac{L_{\rp}}{\ell_{\rp}} \varepsilon,
\end{align}
where the middle equality follows from the fact that $Y_\alpha \in \widetilde H_Y$
and $p - (X_{\iW}+\varu) \perp \widetilde H_Y $ since
\[
P_{\widetilde H_Y}(p) = P_{\widetilde H_Y}(\phi_Y(X_{\iW}+\varu)) = X_{\iW}+\varu.
\]

Now, we would like to estimate $\rp_\alpha(p)$ for $\alpha \in [0,k]$ by applying Lemma~\ref{lem:rpalphaRegular} with $X_{\iW}$, $\sigma_{\alpha}$, $\tau$, $u$, and $\|p - X_{\iW}\|$ playing the roles of $q$, $\sigma$, $\tau$, $u$, and $t$ in the lemma, respectively. We shall show that $\tau$ and $t$ satisfy the assumptions stated in the lemma, namely $16 \eta/r \le \tau \le 1$ and $0 \le \|p-X_{\iW}\| < 0.1\tau r$.

We start with the lower bound for $\tau$. With $\|p-X_{\iW}\|\le 2\|\varu\| \le 96d \frac{L_{\rp}}{\ell_{\rp}} \varepsilon$ and $\eta = \frac{1}{C_{\rm gap}^2 d^{5/2}} r \le \frac{r}{128d}$, we have
\[
\tau = \frac{1}{\|p-X_{\iW}\|} \cdot 12 \frac{L_{\rp}}{\ell_\rp} \varepsilon \ge \frac{1}{8d} \ge \frac{16\eta}{r}.
\]
The upper bound for $\tau$ follows from:
\[
\tau = \left\langle u, \sigma_{\alpha} \frac{p_{\alpha}}{\|p_{\alpha}\|} \right\rangle \le \|u\| \cdot |\sigma_{\alpha}| = 1.
\]

The lower bound for $\|p-X_{\iW}\|$ is trivial, and the upper bound follows from:
\[
\frac{\|p'-X_{\iW}\| }{ 0.1 \tau r } = \frac{\|p'-X_{\iW}\|^2}{ 0.1  \cdot 12 (L_\rp/\ell_\rp) \varepsilon r}
\le
\frac{\left( 96d (L_\rp/\ell_\rp) \varepsilon\right)^2}{0.1 \cdot 12 (L_\rp/\ell_\rp) \varepsilon r}
= 7680 d^2 \frac{L_{\rp}}{\ell_{\rp}} \cdot \frac{\varepsilon}{r} < 1.
\]

We can now apply Lemma~\ref{lem:rpalphaRegular}. With $\tau t = \tau \cdot \|p' - p\| = 12 \frac{L_\rp}{\ell_\rp}\varepsilon$, we get

\begin{align*}
\mbox{if }\sigma_\alpha =&+1, &
\rp_\alpha(p) - 3L_{\rp}\varepsilon
\ge  \rp_\alpha(p') &\ge \rp_\alpha(p) - 24 \frac{L_{\rp}^2}{\ell_\rp} \varepsilon, \mbox{ and} \\
\mbox{if }\sigma_\alpha =&-1, &
\rp_\alpha(p) + 3L_{\rp}\varepsilon
\le  \rp_\alpha(p') &\le \rp_\alpha(p) + 24 \frac{L_{\rp}^2}{\ell_\rp} \varepsilon.
\end{align*}

Suppose $\alpha \in [k]$ and $\sigma_\alpha = +1$.
From the assumption of $p$ and the definition of $\sigma_\alpha$ (see \eqref{eq: sigmaAlpha}), we have
$$\rp(\sqrt{2}r) \le \rp_\alpha(p) \le \rp(\sqrt{2}r-0.95\delta) + n^{-1/2+\varsigma}.$$

With the assumption that $L_{\rp} \varepsilon \ge n^{-1/2+\varsigma}$,
\[
\rp_\alpha(p')
\le \rp_\alpha(p) - 3L_\rp \varepsilon
\le \rp(\sqrt{2}r-0.95\delta)-2L_{\rp}\varepsilon.
\]

For the lower bound of $\rp_{\alpha}(p')$, notice that
\[
\rp(\sqrt{2}r) - \rp(\sqrt{2}r+ 0.95 \delta)
\ge
  \ell_\rp \cdot 0.95 \delta
\ge
  100 \frac{ L_{\rp}^2}{\ell_\rp} \varepsilon .
\]
We have
$$\rp_{\alpha}(p')
\ge \rp_\alpha(p) - 24 \frac{L_\rp^2}{\ell_\rp} \varepsilon
\ge \rp(\sqrt{2}r) - 24 \frac{L_\rp^2}{\ell_\rp} \varepsilon
\ge \rp(\sqrt{2}r + 0.95\delta) +2 L_\rp \varepsilon.$$
Thus, the statement of the lemma follows when $\alpha \in [k]$ and $\sigma_\alpha = +1$. For the rest of the cases (when $\alpha = 0$ or $\sigma_{\alpha} = -1$), the derivations are the same. We omit the proof here.
\end{proof}

\subsection{Finding a cluster in $W'$.}
It remains to prove main statement for this section.

\begin{proof}[Proof of Proposition \ref{prop:nextnavicluster}]
Let us check that $(W,W')$ is a good pair. Lemma~\ref{lem: W'-size-lower-bound} shows that $W' \neq \varnothing$. It is clear by construction that $W' \subseteq W$ and $|W| = n$. Now for any $i \in W'$, by Lemma~\ref{lem:orthogonal} and Proposition~\ref{prop:orthogonal}, there exists $p \in M$ with
\[
\|X_i - p\| \le 2^8 d \frac{L_\rp}{\ell_\rp} \varepsilon < \cdr d \frac{L_\rp}{\ell_\rp} \varepsilon,
\]
such that
\[
\left\{ j \in W \, : \, \|X_j - p\| < \varepsilon \right\} \subseteq W'.
\]
This implies that $(W,W')$ is a good pair. Therefore,  by Lemma~\ref{lem: good-pair-gives-cluster}, $(\gc W, \gc i)$ is a cluster.

Further, since $\gc i \in  \gc W \subset W'$, for $\alpha \in [k]$
we have
\begin{align*}
  \rp(\sqrt{2}r+0.95\delta) \le |N(\gc i) \cap V_\alpha|/|V_\alpha| \le \rp( \sqrt{2}r-0.95\delta).
\end{align*}

Within the event $\Enavi(\{V_\alpha\}_{\alpha \in [0,k]}, W) \supseteq \Eao$,
\begin{align*}
    \big| |N(\gc i) \cap V_\alpha|/|V_\alpha| - \rp_\alpha(X_{\gc i})\big|
\le
    n^{-1/2+\varsigma}.
\end{align*}
Within the event $\Eclu(\{(V_\alpha,i_\alpha)\}_{\alpha \in [0,k]})$,
\begin{align*}
    \big| \rp(\|X_{i_\alpha} - X_{\gc i}\| ) - \rp_\alpha(X_{\gc i})\big|
\le
    L_{\rp}\eta.
\end{align*}

Combining these two bound, we obtain
\begin{align*}
  \sqrt{2}r - \delta \le  \|X_{i_\alpha} - X_{\gc i} \| \le \sqrt{2}r + \delta.
\end{align*}
We can derive the same bound for $\alpha=0$ with $\sqrt{2}r$ replaced by $r$. Finally, together with the occurrence of ${\cal E}_{\rm ao}(i_0,\ldots, i_k)$, we conclude that the list $(i_0,\ldots, i_k, \gc i)$ satisfies ${\cal E}_{\rm ao}(i_0,\ldots, i_k, \gc i)$.
\end{proof}

\section{Using navigation clusters to capture the next cluster}\label{sec: next-cluster}

In this section, we consider the following scenario. We have $d+1$ clusters $(V_\alpha, i_\alpha)$ which forms an ``almost orthogonal'' clusters described in the previous section. Further, there is a vertex $\inx \in {\bf V}$ which is known that the underlying point $X_{\inx}$ is approximately $0.5\delta$ away from $X_{i_0}$. Now a new batch of vertices $W \in {\bf V}$ of size $n$ is given. The objective in this section is to identify a cluster in close proximity to $X_{\inx}$, which relies on the established 'almost orthogonal' clusters represented by $(V_\alpha, i_\alpha)_{\alpha \in [0,d]}$.
    Let us first formulate the assumptions described here as events.

\begin{definition}
For points $i, j \in {\bf V}$, we define the \defn{distance event} of $i$ and $j$ as
\[
{\cal E}_{\rm dist}(i,j) := \big\{ 0.4 \delta \le \|X_i-X_j\| \le 0.6 \delta \big\}.
\]
\end{definition}

The discussion in this section is always within the event
\begin{align}
\label{eq: Enx}
    \Enx
:=&
    \Enx( \{(V_\alpha, i_\alpha)\}_{\alpha \in [0,d]},\inx,W) \\
\nonumber
=&
    \Eclu(\{(V_\alpha, i_\alpha)\}_{\alpha \in [0,d]})
    \cap {\cal E}_{\rm ao}(i_0, i_1, \ldots, i_d) \\
\nonumber
&
    \cap {\cal E}_{\rm dist}(\inx, i_0)
    \cap \Enavi(\{V_\alpha\}_{\alpha \in [0,d]}, W)
    \cap \Enavi(\{V_\alpha\}_{\alpha \in [d]}, \{\inx\})
    \cap \Enet(W),
\end{align}
and the goal of this section is to show the following:
\begin{proposition}\label{prop: NEXTcore}
Consider \( d+1 \) pairs \( \{(V_\alpha, i_\alpha)\}_{\alpha \in [0,d]} \), where each \( i_\alpha \) is an element of \( V_\alpha \), and each \( V_\alpha \) belongs to the set \( \mathbf{V} \). Let \( \inx \in \mathbf{V} \), and let \( W  \subset \mathbf{V} \) be a subset such that \( |W| = n \). Further,  \( \bigcup_{\alpha \in [0,d]}V_\alpha \), $\{\inx\}$, and \( W \) are pairwise disjoint.
Let
\begin{align*}
W^\sharp := \bigg\{  i \in W \, : \,
& \forall \alpha \in [d],
\bigg|\frac{|N(i) \cap V_\alpha|}{|V_\alpha|} - \frac{|N(i_0) \cap V_\alpha|}{|V_\alpha|} \bigg|
\le  c_3 \delta \\
& \mbox{ and }
\frac{|N(i)\cap V_0|}{|V_0|} \ge \rp(2 \delta )
\bigg\}.
\end{align*}

Considering the occurrence of the event \( \Enx( \{(V_\alpha, i_\alpha)\}_{\alpha \in [0,d]},\inx,W) \), when Algorithm~\gclu~ is applied with inputs \( V_1 = W \) and \( V_2 = W^\sharp \), its output \( (\gc {W}, \gc {i}) \) is a cluster and
 $\|X_{\gc i} - X_{\inx}\| \le 0.1\delta$.
\end{proposition}

Through this section, we will fix a triple $\{(V_\alpha,i_\alpha)\}_{\alpha \in [0,d]}$, $\inx$, and $W$ as described in Proposition \ref{prop: NEXTcore}.

The proof of Proposition \ref{prop: NEXTcore} essentially follows the same idea as shown in the previous section.
We need to show
\begin{enumerate}
    \item For every $i \in W^\sharp$, $\|X_i- X_{\inx}\| \le 0.1\delta$.
    \item $(W,W^\sharp)$ is a good pair.
\end{enumerate}

To achieve these two objectives, we need to begin with the derivation of the closeness of certain subspaces, similar to the step presented in Section \ref{sec: SmallAnglesBtwnPlanes}:

\begin{lemma}\label{lem:dHTX0M}
Suppose $i \in {\bf V}$ satisfies $\|X_{i} - X_{\inx}\| \le \delta$.

For each $\alpha \in [d]$, let
$$
    Z = (Z_1,\dots,Z_d)
$$
be the $N \times d$ matrix with
$Z_\alpha := X_i - X_{\inx}$ and $H_Z:= {\rm span}(Z_1, \ldots, Z_d)$. Given the occurrence of the event $\Enx$, we have
\begin{itemize}
    \item[(a)] ${\rm s}_{\rm min}(Z) \ge \frac{r}{2}$.
    \item[(b)] $\sd(H_Z,\TM) \le 2^{-3}/\cdr$, and
    \item[(c)] $\sd(H_Z, H_i) \le 2^{-2}/\cdr$,
\end{itemize}
where $H_{i_0}$ and $H_i$ are the notions for $T_{X_{i_0}}M$ and $T_{X_i}M$, respectively.
\end{lemma}
The proof is essentially analogous to the derivation of Lemma \ref{lem: Hangle}, relying on Lemma \ref{lem:Hflat}, Lemma \ref{lem: YalphaYbeta}, and Lemma \ref{lem: sminB}. Given its relative simplicity, we will provide a sketch of the proof in the Appendix \ref{appx: basic result proof}.

\subsection{$W^\sharp$ is the desired set of vertices}

\begin{lemma} \label{lem:closetoXi0}
Given the occurrence of $\Enx$. For each $i\in W^\sharp$, we have
\begin{itemize}
    \item[(a)] $\|X_i -X_{\inx} \|\le 0.1 \delta$.
    \item[(b)] $\|X_i - X_{i_0}\| \le 0.7 \delta$.
\end{itemize}
\end{lemma}

To prove this lemma, we need to derive a weaker statement:

\begin{lemma} \label{lem: NEXTW'q0}
Within the event $\Enx$, for every $i \in W^\sharp$, we have $\|X_{i_0}-X_i\| < 3\delta$.
\end{lemma}
\begin{proof}
Suppose that $i \in W^\sharp$. If $\|X_{i_0} - X_i\| \le \eta$, then the result follows by the parameter assumptions. From now on, suppose $\|X_{i_0} - X_i\| > \eta$.

We have
\begin{align*}
\rp( \|X_{i_0} -X_i\| - \eta)
&\ge \rp_0(X_i)
& & \mbox{(since $(V_0,X_{i_0})$ is a cluster)} \\
&\ge \frac{|N(i)\cap V_0|}{|V_0|} - n^{-1/2+\varsigma}
& & \mbox{(by the event $\Enavi(\{V_\alpha\}_{\alpha \in [0,d]}, W)$)} \\
&\ge \rp(2\delta) - n^{-1/2+\varsigma} \\
&\ge \rp(2.5\delta) + \frac{1}{2}\ell_\rp \delta - n^{-1/2+\varsigma}
& & \mbox{(by the definition of $\ell_\rp$ (see \eqref{eq: lp}))} \\
&\ge \rp(2.5\delta).
\end{align*}
Since $\rp$ is strictly decreasing, we conclude $\|X_{i_0} - X_i\| \le 2.5 \delta + \eta < 3 \delta$.
\end{proof}

\begin{proof}
Without loss of generality, by a shift of $\R^N$ we assume
$$
    X_{\inx} = \vec{0}.
$$

Let $Y = (Y_1,\ldots, Y_d)$ be the matrix with $Y_\alpha = X_{\inx}-X_{i_\alpha}$
and let $H_Y = {\rm span}(Y_1,\ldots,Y_d)$. With the assumption that $X_{\inx}=0$,
let $P_{H_Y}$ be the projection from $\R^N$ to $H_Y$.

{\bf (a)}
From Lemma~\ref{lem:dHTX0M}, we apply Proposition~\ref{prop: SecondFundamentalForm} with
\[
\zeta = 1 - \sd(H_Y,\TM) \ge 1 - \frac{1}{4\cdr} > 0.99,
\]
and find that there exists a map
\[
\phi_{H_Y}: B(X_{\inx},r_M) \cap H_Y \to M
\]
such that $P_{H_Y}(\phi_{H_Y}(x)) = x$ and
\begin{align} \label{eq: NEXTclosetop01}
\| \phi_{H_Y} (x) - X_{\inx}\| \le \frac{3}{2} \| x - X_{\inx} \|,
\end{align}
for every $x \in B(p,r_M) \cap H$.

Take an arbitrary $i \in W^\sharp$.

By Lemma~\ref{lem: NEXTW'q0}, we can show the rough estimate
\begin{equation} \label{eq: NEXTclosetop02}
    \|P_{H_Y} X_i - X_{\inx}\|
\le
    \|X_i-X_{\inx}\|
\le
    \|X_{i}-X_{i_0}\|+\|X_{i_0}-X_{\inx}\|
<
    4\delta < r_M.
\end{equation}
This shows that $P_{H_Y}X_i$ is in the domain of $\phi_{H_Y}$, and hence we can apply \eqref{eq: NEXTclosetop01} to compare $\|X_i-X_{\inx}\|$ and $\|P_{H_Y}X_i-X_{\inx}\|$:
\begin{equation}\label{ineq:3-2-PH-Xi}
    \|X_i - X_{\inx}\|
=
    \|\phi_{H_Y}(P_{H_Y}(X_i)) - X_{\inx}\|
\le
    \frac{3}{2} \|P_{H_Y}X_i - X_{\inx}\|.
\end{equation}

If $X_i = X_{\inx}$, then the inequality we want to prove holds trivially. From now on, suppose $X_i - X_{\inx} \neq \vec{0}$, and so we can consider
\[
u := \frac{X_i-X_{\inx}}{\|X_i-X_{\inx}\|}.
\]

We have
\[
    P_{H_Y}(u)
=
    \frac{P_{H_Y}(X_i - X_{\inx})}{\|X_i - X_{\inx}\|}
=
    \frac{P_{H_Y}X_i - P_{H_Y}X_{\inx}}{\|X_i - X_{\inx}\|}
=
    \frac{P_{H_Y}X_i - X_{\inx}}{\|X_i - X_{\inx}\|},
\]
and thus from \eqref{ineq:3-2-PH-Xi}, we find that $\|P_{H_Y}(u)\| \ge \frac{2}{3}$.

From Lemma \ref{lem:dHTX0M} we have
$$
    \|Y^\tp u\|
=
    \|Y^\tp P_{H_Y}u\|
=
    \big( \min_{v\in {\rm Im}Y\,:\, \|v\|=1} \|Y^\tp v\| \big) \cdot \|P_{H_Y}u\|
=
    {\rm s}_{\rm min}(Y) \|P_{H_Y}u\|
\ge
    \frac{r}{2} \cdot \frac{2}{3}
\ge
    \frac{r}{3},
$$
where the third equality is a standard consequence of the singular value decomposition.

This implies that there exists $\alpha \in [d]$ such that
\begin{equation}\label{ineq:p-alpha-u}
|\langle Y_\alpha , u\rangle| \ge \frac{r}{3\sqrt{d}}.
\end{equation}

From now on, we fix such $\alpha \in [d]$. Our plan is to bound $\|X_i -X_{\inx}\|$ from above by $\|\rp_\alpha(X_i) - \rp_\alpha(X_{\inx})\|$ using Lemma~\ref{lem:rpalphaRegular}, and bound the latter term by $\left|\frac{|N(i) \cap V_\alpha|}{|V_\alpha|} - \frac{|N(\inx) \cap V_\alpha|}{|V_\alpha|} \right|$ from the conditions of the events $\Enavi(\{V_\alpha\}_{\alpha \in [0,d]}, W)$ and $\Enavi(\{V_\alpha\}_{\alpha \in [d]}, \{\inx\})$. Once this is completed, we will see this provides the desired upper bound for $\|X_i-p\|$ from the definition of $W^\sharp$.

Let us set up the proper parameters and show that they satisfy the constraints from Lemma~\ref{lem:rpalphaRegular}.

\begin{itemize}
\item First, we have
\[
\|Y_\alpha\| \le \| X_{\inx} - X_{i_0}\| + \|X_{i_0}-X_{i_\alpha}\| \le 0.6\delta + (r+\delta) \le 2r,
\]
and similarly,
$\|Y_\alpha\| \ge
    \|X_{i_0}-X_{i_\alpha}\|- \| X_{\inx} - X_{i_0}\|
 \ge r-\delta - 0.6\delta \ge 0.9r$. Hence, $X_{\inx}$ satisfies the condition of $q$ in Lemma~\ref{lem:rpalphaRegular}.

\item Second, let $\sigma \in \{-1,+1\}$ be the sign of $\langle Y_\alpha, u\rangle$ and define
\[
\tau :=
\left\langle u, \sigma \frac{ Y_\alpha }{ \|Y_\alpha\| }\right\rangle.
\]
We find that
\begin{equation}\label{ineq:tau-r-sqrt-d}
\tau =
\frac{1}{\|Y_{\alpha}\|} \cdot |\langle Y_{\alpha}, u \rangle|
\overset{\eqref{ineq:p-alpha-u}}{\ge}
\frac{1}{\|Y_\alpha\|} \cdot \frac{r}{3 \sqrt{d}} \ge \frac{1}{6\sqrt{d}}
\ge \frac{16 \eta}{r},
\end{equation}
from the parameter assumptions. Hence, $\tau$ also satisfies the assumption in Lemma~\ref{lem:rpalphaRegular}.

\item Lastly, let $t:=\|X_i-X_{\inx}\|$ so that $X_{\inx}+tu = X_i$. By \eqref{eq: NEXTclosetop02}, we have
\[
t < 4\delta
= \frac{4r}{\cdr d^2}
\le \frac{r}{100\sqrt{d}}
\le 0.1\tau r,
\]
where the last step uses $\tau \ge \frac{1}{10\sqrt{d}}$, which we know from \eqref{ineq:tau-r-sqrt-d}. This shows that $t$ satisfies the condition in Lemma~\ref{lem:rpalphaRegular}.
\end{itemize}

Now we can apply Lemma~\ref{lem:rpalphaRegular} to get
\begin{equation}
\label{eq: NEXTclosetop03}
  \big|\rp_\alpha(X_i) - \rp_\alpha(X_{\inx})\big|
=
  \big|\rp_\alpha(X_{\inx} +tu) - \rp_\alpha(X_{\inx})\big|
\ge
  \ell_\rp \frac{\tau}{4} \|X_i-X_{\inx}\|
\ge
\frac{\ell_\rp}{24\sqrt{d}}\|X_i-X_{\inx}\|.
\end{equation}
Within the events $\Enavi(\{V_\alpha\}_{\alpha \in [0,d]}, W)$ and $\Enavi(\{V_\alpha\}_{\alpha \in [d]}, \{\inx\})$, we have
\begin{align}\label{eq: NEXTclosetop04}
\left|\rp_\alpha(X_i) - \rp_\alpha(X_{\inx})\right|
&\le
\left|\frac{|N(i) \cap V_\alpha|}{|V_\alpha|} - \frac{|N(\inx) \cap V_\alpha|}{|V_\alpha|} \right| + 2 n^{-1/2+\varsigma} \\
&\le
c_3 \delta
+ 2 n^{-1/2+\varsigma} \le
\frac{25}{24} c_3 \delta \le
\frac{25}{24} \frac{\ell_\rp}{\cdr \sqrt{d}} \delta. \nonumber
\end{align}
Combining \eqref{eq: NEXTclosetop03} and \eqref{eq: NEXTclosetop04}, we conclude that $\|X_i - X_{\inx} \| \le \frac{25}{\cdr} \delta < 0.1 \delta$.

\medskip

{\bf (b)} This follows immediately from part~(a) and the triangle inequality:
\[
\|X_i - X_{i_0}\| \le \|X_i-X_{\inx}\| + \|X_{\inx}-X_{i_0}\| \le 0.1 \delta + 0.6 \delta = 0.7 \delta.
\]
We have finished the proof.
\end{proof}

\subsection{ $(W,W^\sharp)$ is a good pair}
We begin this subsection by using the second condition in the definition of $W^\sharp$ to prove the following lemma, which gives a rough upper estimate for $\|X_{i_0} - X_i\|$ for each $i \in W^\sharp$. Later, we are going to obtain a better estimate for the same quantity in Lemma~\ref{lem:closetoXi0}(b).

\begin{lemma}\label{lem: NEXTW'>0}
Within the event $\Enx$, we have $|W^\sharp| \ge n^{1-\varsigma}$.
\end{lemma}
\begin{proof}
Consider the set $S := \left\{ i \in W \, : \, \| X_i - X_{\inx} \| < \varepsilon \right\}$. Within the event $\Enet(W)$ (see \eqref{eq:Enet}), we know $|S| \ge n^{1-\varsigma}$. We claim that $S \subseteq W^\sharp$.

Take any $i \in S$. Let us check the two conditions for $i$ to be in $W^\sharp$. First, within the events $\Enavi(\{V_\alpha\}_{\alpha \in [0,d]}, W)$ and $\Enavi(\{V_\alpha\}_{\alpha \in [d]}, \{\inx\})$, for any $\alpha \in [d]$, we have
\begin{align*}
  \bigg|\frac{|N(i) \cap V_\alpha|}{|V_\alpha|}
    - \frac{|N(\inx) \cap V_\alpha|}{|V_\alpha|} \bigg|
& \le
  \Big| \rp_\alpha(X_i) - \rp_\alpha(X_{\inx}) \Big| + 2 n^{-1/2+\varsigma} \\
& \le
 \sum_{j \in V_\alpha} \frac{1}{|V_\alpha|} \big|\rp(\|X_i-X_j\|) - \rp(\|X_j-X_{\inx}\|) \big|
   + 2 n^{-1/2+\varsigma} \\
& \le L_\rp\varepsilon + 2n^{-1/2+\varsigma}
\le 3 L_\rp \varepsilon
\le c_3 \delta.
\end{align*}

For the second condition,
\begin{align*}
\frac{|N(i)\cap V_0|}{|V_0|}
& \ge
\sum_{j\in V_0} \frac{1}{|V_0|} \rp( \|X_j- X_i \|)
- n^{-1/2+\varsigma} \\
& \ge \sum_{j\in V_0} \frac{1}{|V_0|} \rp( \|X_j-X_{i_0}\| + \|X_{i_0}-X_{\inx}\| + \|X_{\inx}-X_i\|) - n^{-1/2+\varsigma}\\
& \ge \sum_{j\in V_0} \frac{1}{|V_0|} \rp( \eta + 0.6\delta + \varepsilon)
- n^{-1/2+\varsigma} \\
&
\ge
\rp(0.8\delta)- n^{-1/2+\varsigma}
>
  \rp(2\delta).
\end{align*}
Hence, $i \in W^\sharp$. The proof is complete.
\end{proof}

\begin{lemma}\label{lem:sharp-before-move}
Given the occurrence of the event $\Enx$. For any $i \in W^\sharp$ and for any $\alpha \in [d]$, we have
\[
|\rp_\alpha(X_i) - \rp_\alpha(X_{\inx})| \le c_3 \delta + 2 L_\rp \varepsilon.
\]
\end{lemma}
\begin{proof}
By the triangle inequality and the parameter assumptions,
\begin{align*}
&|\rp_\alpha(X_i) - \rp_\alpha(X_{\inx})| \\
&\le
    \underbrace{ \left| \rp_\alpha(X_i) - \frac{|N(i) \cap V_\alpha|}{|V_\alpha|} \right| }_{ \Enavi(\{(V_\alpha,i_\alpha)\}_{\alpha \in [0,d]},W)}
+ \left| \frac{|N(i) \cap V_\alpha|}{|V_\alpha|}
- \frac{|N(\inx) \cap V_\alpha|}{|V_\alpha|} \right|
+ \underbrace{\left| \frac{|N(\inx) \cap V_\alpha|}{|V_\alpha|} - \rp_\alpha(X_{\inx}) \right| }_{ \Enavi(\{(V_\alpha,i_\alpha)\}_{\alpha \in [0,d]},\{\inx\})}
\\
&\le
    n^{-\frac{1}{2} + \varsigma}
    + c_3 \delta
    + n^{-\frac{1}{2}
    + \varsigma}
\le
    c_3 \delta + 2 L_\rp \varepsilon,
\end{align*}
as desired.
\end{proof}

The main technical result is the following.
\begin{lemma} \label{lem: NEXTcore}
For any $\iWs \in W^\sharp$, there exists a point $p\in M$
so that the following holds:
\begin{itemize}
    \item[(a)]$\|X_{\iWs} - p\| \le 2^8 \sqrt{d} \, \frac{L_{\rp}}{\ell_{\rp}} \varepsilon$;
    \item[(b)] For every $\alpha \in [d]$, \(
|\rp_\alpha(p) - \rp_\alpha(X_{\inx})|
\le c_3 \delta - 3L_{\rp}\varepsilon;
\)
\item [(c)] $\|p-X_{\inx}\| \le \delta$.
\end{itemize}
\end{lemma}
\begin{proof}

Let us fix $\iWs \in W^\sharp$.
Without loss of generality, we make a shift of $\R^N$ and assume
$$
    X_{\iWs} = \vec{0}.
$$

\underline{Step 1. Find $p$ through construction.}

By Lemma~\ref{lem:closetoXi0}(a), we have $\|X_{\iWs}- X_{\inx}\| \le 0.1 \delta$. Hence, we can apply Lemma~\ref{lem:dHTX0M} with $i = \iWs$.
Let $Z  = (Z_1,\dots, Z_d) $ with $Z_\alpha = X_{\iWs}- X_{i_\alpha}$ for $\alpha \in [d]$ and $H_Z = {\rm span}(\{Z_\alpha\}_{\alpha \in [d]})$.
By Lemma~\ref{lem:dHTX0M}, we have
\begin{equation}\label{ineq:wt-H-Tw}
    \sd(H_Z, H_{\iWs})< 2^{-2}/\cdr.
\end{equation}

Let $P_{H_Z}: \mathbb{R}^N \to H_Z$ be the orthogonal projection to $H_Z$. From \eqref{ineq:wt-H-Tw}, we can apply Proposition~\ref{prop: SecondFundamentalForm} with $\zeta = 1 - \sd(H_Z, H_{\iWs}) > 1 - \frac{1}{\cdr} > 0.99$ to obtain a map
\begin{align}
\label{eq: NEXTcore04}
\phi_{H_Z}: B(X_{\iWs},r_M) \cap H_Z \to M
\end{align}
such that $P_{H_Z}(\phi_{H_Z}(x)) = x$ and
\begin{equation}\label{eq: NEXTcore01}
\| \phi_{H_Z} (x) - X_{\iWs} \| \le 2\|x-X_{\iWs}\|,
\end{equation}
for any $x \in B(X_{\iWs},r_M) \cap H_Z$.

\vspace{0.3cm}

For each $\alpha \in [d]$, let $\sigma_{\alpha} := \sign(\rp_{\alpha}(X_{\iWs}) - \rp_{\alpha}(X_{\inx}))$. (Recall the definition of $\sign$ from \eqref{eq:defn-sign}.)

Now using the same argument as the proof of Lemma~\ref{lem:orthogonal}, since the matrix $Z$ has full rank (from Lemma \ref{lem:dHTX0M} (a)), there exists a unique $\varpi \in H_{Z}$ such that
\begin{align}
\label{eq: NEXTCORE02}
    \Big\langle \sigma_\alpha \frac{Z_\alpha}{\|Z_\alpha\|},
    \varpi \Big\rangle
=
    2^5 \frac{L_{\rp}}{\ell_{\rp}} \varepsilon,
\end{align}
for every $\alpha \in [d]$. We attempt to set the point $p$ described in the statement of the lemma to be
\begin{align}
\label{eq: NEXTcore03}
    p = \phi_{H_Z}(X_{\iWs} + \varpi ).
\end{align}
To achieve that, we need to show $X_{\iWs}+\varpi$ is in the domain of $\phi_{H_Z}$, which is equivalent to show $\|\varpi\| <r_M$ (See \eqref{eq: NEXTcore04}).

First, relying on ${\rm s}_{\rm min}(Z) \ge \frac{r}{2}$ from Lemma \ref{lem:dHTX0M} (a),
\begin{align*}
    \|\varpi\|
\le
    \frac{\|Z^\tp \varpi\|}{ {\rm s}_{\rm min}(Z)}
=
    \frac{2}{r}\sqrt{\sum_{\alpha \in [d]} \langle Z_\alpha, \varpi \rangle^2}
\overset{\eqref{eq: NEXTCORE02}}{=}
    \frac{2^6}{r} \cdot \frac{L_\rp}{\ell_\rp} \varepsilon \sqrt{ \sum_{\alpha \in [d]} \|Z_\alpha\|^2}.
\end{align*}

Using the estimate
\begin{align*}
    \|Z_\alpha\|
\le
    \underbrace{\|X_{\iWs} - X_{i_0}\|}_{ \rm{Lemma}~\ref{lem:closetoXi0}(b)}
    + \underbrace{\|X_{i_0} - X_{i_\alpha}\|}_{{\cal E}_{\rm ao}(i_0,\ldots,i_d)}
\le
    0.7 \delta + r + \delta
    <
    2r,
\end{align*}

we find that

\begin{equation}\label{ineq:varpi-upper-bound}
\|\varpi\| \le 2^7 \sqrt{d} \frac{L_\rp}{\ell_\rp} \varepsilon < 0.1 \delta < r_M,
\end{equation}
which shows that $X_{\iWs}+\varpi$ is in the domain of $\phi_{H_Z}$. Hence, the definition of $p$ in \eqref{eq: NEXTcore04} is valid.

\vspace{0.3cm}

\underline{Step 2: Demonstrating that $p$ possesses the required properties.}

First, property $(a)$ follows from \eqref{eq: NEXTcore01}:
\begin{equation}\label{ineq:w'-w-varpi}
    \|p -X_{\iWs} \|
=
    \| \phi_{H_Z}(X_{\iWs}+\varpi)
    - X_{\iWs}\| \le 2\|\varpi\|
\overset{\eqref{ineq:varpi-upper-bound}}{\le}
    2^8 \sqrt{d} \frac{L_\rp}{\ell_\rp} \varepsilon.
\end{equation}

Next, let us estimate $\rp_\alpha(p)- \rp_\alpha(X_{\inx})$ for $\alpha \in [d]$.
Once again, our tool is Lemma~\ref{lem:rpalphaRegular}, and so we have to set up some parameters. From the definition of $\varpi$, we know that $\varpi \neq \vec{0}$, and therefore $p = \phi_{H_Z}(X_{\iWs} +\varpi) \neq X_{\iWs}$. Hence, it makes sense to define
\[
u := \frac{p-X_{\iWs}}{\|p-X_{\iWs}\|} \in \mathbb{S}^{N-1}
\quad \mbox{and set}
\quad
\tau := \left\langle u, \sigma_{\alpha} \frac{Z_\alpha}{\|Z_\alpha\|} \right\rangle.
\]

We would like to apply Lemma~\ref{lem:rpalphaRegular} with $p$, $\sigma_{\alpha}$, $\tau$, $u$, and $\|p-X_{\iWs}\|$ playing the roles of $q$, $\sigma$, $\tau$, $u$, and $t$ in the lemma, respectively. Let us check the following items.
\begin{itemize}
    \item[(i)] $0.9 r \le \|Z_\alpha\| \le 2r$,
    \item[(ii)] $16 \eta/r \le \tau$, and
    \item[(iii)] $\|p-X_{\iWs}\| < 0.1 \tau r$.
\end{itemize}

Item~(i) follows from the triangle inequality and Lemma~\ref{lem:closetoXi0}(b):
\begin{align*}
    \Big| \|Z_\alpha\| - r \Big|
&\le
    \Big| \|Z_\alpha\| - \|X_{i_0}-X_{i_\alpha}\| \Big|
    + \underbrace{\Big| \|X_{i_0}-X_{i_\alpha}\| - r \Big|}_{{\cal E}_{\rm ao}(i_0,\ldots,i_d)} \\
&\le
    \|X_{\iWs}-X_{\inx}\| + \delta \le 1.7 \delta < 0.1 r.
\end{align*}
For Item~(ii), we have
\begin{align*}
    \tau
=
  \Big\langle \frac{p-X_{\iWs}}{\|p-X_{\iWs}\|}, \sigma_\alpha \frac{ Z_\alpha}{\|Z_\alpha\|} \Big\rangle
=
  \Big \langle \frac{\varpi}{\|p-X_{\iWs}\|}, \sigma_\alpha \frac{ Z_\alpha}{\|Z_\alpha\|} \Big\rangle
\overset{\eqref{eq: NEXTCORE02}}{=}
  \frac{1}{\|p- w \|} \cdot 2^5 \frac{L_{\rp}}{\ell_{\rp}} \varepsilon
\overset{\eqref{ineq:w'-w-varpi}}{\ge} \frac{1}{8\sqrt{d}} \ge  16 \frac{\eta}{r}.
\end{align*}
Finally, for Item~(iii), we have
\[
    \frac{\|p-X_{\iWs}\|}{\tau}
=
    \frac{\|p-X_{\iWs}\|^2}{2^5 (L_\rp/\ell_\rp) \varepsilon}
\le
    2^{11} d \frac{L_\rp}{\ell_\rp}\varepsilon
    \le
    2^{11}d \frac{1}{\cdr}\eta
    \le
    \frac{2^{11}}{d^{1.5} \cdr^3 } r
<
    0.1 r.
\]

Hence, Lemma~\ref{lem:rpalphaRegular} is applicable.
For the moment let us assume $\sigma_\alpha = +1$.
 In this case, $\rp_\alpha(X_{\iWs}) - \rp_\alpha(X_{\inx}) \ge 0$. Lemma~\ref{lem:rpalphaRegular}(b) gives
\begin{equation}\label{ineq:p-a-w-w'-b}
    \rp_\alpha(X_{\iWs}) - 2L_\rp \cdot 2^5 \frac{L_\rp}{\ell_\rp} \varepsilon
\le
    \rp_\alpha(p)
\le
    \rp_\alpha(X_{\iWs}) - \frac{1}{4} \ell_\rp
    \cdot 2^5 \frac{L_\rp}{\ell_\rp} \varepsilon,
\end{equation}
where we applied
$\tau \|p-X_{\iWs}\| = 2^5 \frac{L_\rp}{\ell_\rp} \varepsilon$.
 we have
 \begin{equation}\label{ineq:compare-w'-p-ge}
     \rp_\alpha(p)
 \overset{\eqref{ineq:p-a-w-w'-b}}{\ge} \rp_\alpha(X_{\iWs}) - 2^6 \frac{L_\rp^2}{\ell_\rp} \varepsilon
 \ge
    \rp_\alpha(X_{\inx}) - 2^6 \frac{L_\rp^2}{\ell_\rp} \varepsilon.
 \end{equation}
Note that Lemma~\ref{lem:sharp-before-move} implies that
\begin{equation}\label{lem:w-le-p-2}
    \rp_\alpha(X_{\iWs})
\le
    \rp_\alpha(X_{\inx}) + c_3 \delta + 2 L_\rp \varepsilon.
\end{equation}
Thus,
\begin{equation}\label{ineq:compare-w'-p-le}
    \rp_\alpha(p)
\overset{\eqref{ineq:p-a-w-w'-b}}{\le}
    \rp_\alpha(X_{\iWs}) - 8 L_\rp \varepsilon
\overset{\eqref{lem:w-le-p-2}}{\le}
    \rp_\alpha(X_{\inx}) + c_3 \delta - 6 L_\rp \varepsilon.
\end{equation}
Considering \eqref{ineq:compare-w'-p-ge} and \eqref{ineq:compare-w'-p-le}, we conclude that
\[
|\rp_\alpha(p) - \rp_\alpha(X_{\inx})|
\le
    \max\Big\{ c_3 \delta - 6 L_\rp \varepsilon,\,
    2^6 \frac{L_\rp^2}{\ell_\rp} \varepsilon\Big\}
< c_3 \delta - 3 L_\rp \varepsilon.
\]
The proof for the case when $\sigma_\alpha=-1$ is analogous to the argument above, where we applied Lemma~\ref{lem:rpalphaRegular}(c) instead. We omit the proof here.

Finally, it remains to show Property $(c)$: $\|p-X_{\inx}\|\le \delta$. This last piece follows from the triangle inequality:
\[
\|p-X_{\inx}\| \le \|p-w\| + \|w-X_{\inx}\| \overset{\eqref{ineq:w'-w-varpi}}{\le} 2^8\sqrt{d} \frac{L_\rp}{\ell_\rp}\varepsilon + 0.7\delta < \delta.
\]
We have finished the proof.
\end{proof}

\begin{proof}[Proof of Proposition \ref{prop: NEXTcore}]
Let $(\gc W, \gc i)$ be the output of Algorithm~\gclu~ with input $(W,W^\sharp)$.

\vspace{0.1cm}
\underline{Step 1. $(\gc W, \gc i)$ is a cluster. }
\vspace{0.1cm}

From Lemma \ref{lem: good-pair-gives-cluster}, considering $\Enx \subseteq \Ecn(W) \cap \Enet(W)$,
it suffices to show that $(W,W^\sharp)$ is a good pair. We have $W^\sharp \subseteq W$ and $|W| = n$ by construction, and we know $W^\sharp \neq \varnothing$ from Lemma~\ref{lem: NEXTW'>0}.

Take an arbitrary point $\iWs \in W^\sharp$. Let $p \in M$ be a point obtained from Lemma~\ref{lem: NEXTcore}. We claim that this point $p$ may play the role of $p$ in Definition~\ref{def:good-pair} (when $\iWs$ plays the role of $i$).

Note that
\[
\|X_{\iWs} - p\| \le 2^8 \sqrt{d} \frac{L_\rp}{\ell_\rp}\varepsilon < \cdr d \frac{L_\rp}{\ell_\rp} \varepsilon,
\]
so we only need to establish the inclusion $\left\{ j \in W \, : \, \|p - X_j\| < \varepsilon\right\} \subseteq W^\sharp$. To that end, take any $j \in W$ with $\|p - X_j\| < \varepsilon$. By the triangle inequality, for each $\alpha \in [d]$, we have
\begin{align*}
\left| \frac{|N(j) \cap V_\alpha|}{|V_\alpha|} - \frac{|N(i_0) \cap V_\alpha|}{|V_\alpha|} \right| &\le \left| \frac{|N(j) \cap V_\alpha|}{|V_\alpha|} - \rp_\alpha(X_j) \right| + \left| \rp_\alpha(X_j) - \rp_\alpha(p) \right| \\
&\hphantom{\le}+ \left| \rp_\alpha(p) - \rp_\alpha(p) \right| + \left| \rp_\alpha(p) - \frac{|N(i_0) \cap V_\alpha|}{|V_\alpha|} \right| \\
&\le n^{-1/2+\varsigma} + L_\rp \varepsilon + c_3 \delta - 3 L_\rp \varepsilon + n^{-1/2+\varsigma} \\
&\le c_3 \delta.
\end{align*}
Next, note that for each $j' \in V_0$, we have
\[
\|X_j - X_{j'}\| \le \|X_j - p\| + \|p-X_{i_0}\| + \|X_{i_0} - X_{j'}\| < \varepsilon + \delta + \eta < 1.1 \delta.
\]
Therefore,
\[
\frac{|N(j) \cap V_0|}{|V_0|} \ge \sum_{j' \in V_0} \frac{\rp(\|X_j - X_{j'}\|)}{|V_0|} - n^{-1/2+\varsigma} \ge \rp(1.1 \delta) - n^{-1/2+\varsigma} > \rp(2\delta).
\]
This shows that $j \in W^\sharp$.

\underline{Step 2. $\|X_{\gc i} - X_{\inx}\| \le 0.1\delta$}

Since $\gc i \in W^\sharp$ and each $i \in W^\sharp$ satisfies
$\|X_{i}-X_{\inx}\|\le 0.1\delta$ by Lemma~\ref{lem:closetoXi0}, the statement follows immediately.

\end{proof}

\medskip

\bigskip

\section{Next pivot for the net}
\label{sec: netCheck}
Our algorithm will construct a sequence of clusters. If successful, in the $\ell^\text{th}$ round we will have
\[
(U_1,u_1),(U_2,u_2),\ldots,(U_\ell,u_\ell),
\]
where each pair $(U_\alpha, u_\alpha)$ is a cluster.  Furthermore, for any $\{\alpha, \beta \} \in {[\ell] \choose 2}$, $\|X_{u_\alpha}-X_{u_\beta}\| \ge 0.3\delta$.

In this section, we consider the scenario where a desired $\ell$ pairs $\{(U_\alpha,u_\alpha)\}_{\alpha \in [\ell]}$ has been found, and we would like to find the next pair $(U_{\ell+1},u_{\ell+1})$. As an intermediate step, we will use a new batch of vertices $W$ and find a vertex $\inx \in W$ such that  $\|X_{u_\alpha}-X_{\inx}\| \ge 0.4 \delta$ for each $\alpha \in [\ell]$.

\medskip

\medskip

Let us introduce some necessary notations and build the algorithm. In this section, we fix $\ell \in {\mathbb N}$, suppose $\{(U_\alpha, u_\alpha)\}_{\alpha \in [\ell]}$ are $\ell$ pairs where $u_\alpha \in U_{\alpha} \subseteq {\bf V}$. Further, let $W \subseteq {\bf V}$ be a subset of size $|W|=n$.
We define the ``repulsion'' event
\begin{align} \label{eq: eventRps}
    {\cal E}_{\rm rps}(\{u_\alpha\}_{\alpha \in [\ell]}) := \bigg\{
    \forall \{\alpha, \beta\} \in \binom{[\ell]}{2}, \, \|X_{u_\alpha} - X_{u_\beta}\| \ge 0.3 \delta
    \bigg\}.
\end{align}

For each $q \in M$ and for each $\alpha \in [\ell]$, write
\[
\rp_\alpha(q) := \sum_{j \in U_\alpha} \frac{\rp(\|q-X_j\|)}{|U_\alpha|}.
\]
Next, we define two sets
\begin{align*}
W^\flat := \left\{i \in W \, : \, \forall \alpha \in [\ell], \, \frac{|N(i) \cap U_{\alpha}|}{|U_\alpha|} \le \rp(0.45 \delta) \right\}.
\end{align*}
and for each $\alpha \in [\ell]$, we define
\begin{align*}
W^\natural_\alpha := \left\{i \in W\,:\, \frac{|N(i)\cap U_{\alpha}|}{|U_\alpha|} \ge  \rp(0.55\delta)\right\}.
\end{align*}
We emphasize that $W^\flat$ and $W^\natural_\alpha$ can be identified given $\{(U_\alpha, u_\alpha)\}_{\alpha \in [\ell]}$, $W$, and the graph ${\bf G}$.

The discussion in this section is within the event
\begin{align} \label{eq: Enc}
\Enc :=& \Enc\Big(\{(U_\alpha, u_\alpha)\}_{\alpha \in [\ell]}, W \Big) \\
\nonumber
=&
\Eclu\Big(\{(U_\alpha, u_\alpha)\}_{\alpha \in [\ell]}\Big)
\cap
{\cal E}_{\rm rps}(\{u_\alpha\}_{\alpha \in [\ell]})
\cap \Enavi\big( \{U_\alpha\}_{\alpha \in [\ell]}, W \big)
\cap \Enet(W).
\end{align}

The following Algorithm~\nc~ is to find a vertex $\inx \in W$ such that
$\|X_{\inx}-X_{u_\alpha}\| \ge 0.4\delta$ for every $\alpha \in [\ell]$
at the same time there exists a specific $\alpha \in [\ell]$ such that
$\|X_{\inx} - X_{u_\alpha}\| \le 0.6\delta$:

\begin{algorithm}
\caption{\nc}
\label{alg:net-check}

\SetKwInOut{Input}{Input}
\SetKwInOut{Output}{Output}

\Input{
$\{(U_\alpha, u_\alpha)\}_{\alpha \in [\ell]}$ with $|U_\alpha|\le n$\,,
\\
$W \subseteq {\bf V}$ with $|W|=n$.
}
\Output{
    $\inx \in W$ and \\
    $(V_0, i_0) \in \{(U_\alpha, u_\alpha)\}_{\alpha \in [\ell]}$.
}

$W^\flat \gets \left\{i \in W \, : \, \forall \alpha \in [\ell], \, \frac{|N(i) \cap U_{\alpha}|}{|U_\alpha|} \le \rp(0.45 \delta) \right\}$\;
\For{$\alpha \in [\ell]$}{
    $W^\natural_\alpha \gets \left\{i \in W\,:\, \frac{|N(i)\cap U_{\alpha}|}{|U_\alpha|} \ge  \rp(0.55\delta)\right\}$\;
    \For{$w \in W^\natural_\alpha \cap W^\flat$}{
        $\inx \gets w$\;
        $(V_0, i_0) \gets (U_\alpha, u_\alpha)$\;
        \Return{$(\inx, (V_0, i_0))$}\;
    }
}
\Return {\rm null}
\end{algorithm}

\paragraph{Complexity}
The algorithm runs in time
\begin{align}
    \label{eq: cnet-runningtime}
 O(\ell \cdot n^2),
\end{align}
 dominated by computing $|N(i) \cap U_\alpha|$ for each $i \in W$ and each $\alpha \in [\ell]$. This involves evaluating neighborhood intersections between each candidate node and each cluster, resulting in $\ell \cdot n^2$ such computations.

The main statement we will prove in this section is the following.
\begin{proposition}
\label{prop: net-check}
Condition on $\Enc$.
Running Algorithm~\nc~ will result in one of the following two cases:
\begin{itemize}
    \item If it returns a pair $(\inx, (V_0, i_0))$, then
        \begin{align*}
            \Enc
        \subseteq
            \big\{ \forall \alpha \in [\ell],\, \|X_{\inx} - X_{u_\alpha} \| \ge 0.4 \delta \big\}
         \cap
            \big\{ \|X_{\inx} - X_{i_0} \| \le 0.6 \delta  \big\}
        \cap
            \Eclu((V_0,i_0)).
        \end{align*}
    \item If it returns ${\rm null}$, then
    $\{X_{u_\alpha}\}_{\alpha \in [\ell]}$ is a $\delta$-net of $M$.
\end{itemize}
\end{proposition}

The proof of the Proposition \ref{prop: net-check} will be break into the following three lemmas.

\begin{lemma}\label{lem: 0.4-delta}
Condition on $\Enc$.
If $i \in W^\flat$, then for every $\alpha \in [\ell]$, we have $\|X_i - X_{u_\alpha}\| > 0.4 \delta$.
\end{lemma}
\begin{proof}
We argue by contraposition. Suppose $i \in W$ is an index for which there exists $\alpha \in [\ell]$ with $\|X_i - X_{u_\alpha}\| \le 0.4 \delta$. For any $j \in U_\alpha$, we have
\[
0 \le \|X_i - X_j\| \le \|X_i - X_{u_\alpha}\| + \|X_{u_\alpha} - X_j\| < 0.4 \delta + \eta < 0.41 \delta,
\]
and thus $\rp(\|X_i - X_j\|) > \rp(0.41 \delta)$. Since $\rp(0.41 \delta) - \rp(0.45 \delta) \ge \ell_\rp \cdot 0.04 \delta > n^{-1/2+\varsigma}$, we have
\[
\rp(\|X_i - X_j\|) > \rp(0.45 \delta) + n^{-1/2+\varsigma}.
\]
Therefore, from $\Enavi\big( \{U_\alpha\}_{\alpha \in [\ell]}, W \big)$, we have
\[
\frac{|N(i) \cap U_\alpha|}{|U_\alpha|} \ge \sum_{j \in U_\alpha} \frac{\rp(\|X_i - X_j\|)}{|U_\alpha|} - n^{-1/2+\varsigma} > \rp(0.45 \delta),
\]
which implies $i \notin W^\flat$.
\end{proof}

\begin{lemma}\label{lem: 0.6-delta}
Condition on $\Enc$.
For any $\alpha \in [\ell]$, if $i \in W^\natural_\alpha$, then $\|X_i - X_{u_\alpha}\| < 0.6 \delta$.
\end{lemma}
\begin{proof}
This proof is similar to the proof of Lemma~\ref{lem: 0.4-delta}, so we omit the details.
\end{proof}

The last piece to prove Proposition~\ref{prop: net-check} is the following lemma.
\begin{lemma} \label{lem: deltaNet}
Condition on $\Enc$. If
\[
\bigcup_{\alpha \in [\ell]} W^\natural_{\alpha} \cap W^\flat = \varnothing,
\]
then there is no point $q\in M$ such that for every $\alpha \in [\ell]$, $\|q-X_{u_{\alpha}}\| \ge \delta$. In other words, $\{X_{u_{\alpha}}\}_{\alpha\in[\ell]}$ is a $\delta$-net.
\end{lemma}

\begin{proof}
Suppose, for the sake of contradiction, that there exists $q \in M$ such that for every $\alpha \in [\ell]$, $\|q-X_{u_\alpha}\| \ge \delta$. Since $M$ is connected, there exists a path $\gamma:[0,1]\to M$ from $q$ to $X_{i_1}$.

The function $f:[0,1] \to \mathbb{R}$ given by
\[
f(t) := \min_{\alpha \in [\ell]} \|\gamma(t) - X_{u_\alpha}\|
\]
is continuous with $f(0)\ge \delta$ and $f(1)=0$.
Let $t_0 \in [0,1]$ be the smallest number for which $f(t_0) = 0.5\delta$ and let $q' := \gamma(t_0)$.

Now we have $q'$ such that $\|q' - X_{u_\alpha}\| \ge 0.5\delta$ for every $\alpha \in [\ell]$. Within the event $\Enet(W) \supseteq \Enc$, there exists $i \in W$ such that $\|X_i -q'\| < \varepsilon$. By the triangle inequality, for any $j \in U_\alpha$, we have
\[
\|X_i - X_j\| \ge \|q' - X_{i_\alpha}\| - \|q'-X_i\| - \|X_j - X_{i_\alpha}\| \ge 0.5 \delta - \varepsilon - \eta > 0.48 \delta.
\]
Therefore, within $\Enavi\big( \{U_\alpha\}_{\alpha \in [\ell]}, W \big)\supseteq \Enc$, we have
\[
\frac{|N(i) \cap U_\alpha|}{|U_\alpha|} \le \sum_{j \in U_\alpha} \frac{\rp(\|X_i - X_j\|)}{|U_\alpha|} + n^{-1/2 + \varsigma} < \rp(0.48 \delta) + n^{-1/2+\varsigma} < \rp(0.45 \delta).
\]
Hence, $i \in W^\flat$.

Also, since $f(t_0) = 0.5 \delta$, there exists $a \in [\ell]$ such that $\|q' - X_{i_a}\| = 0.5 \delta$. By a similar triangle inequality argument as above, we find that for any $j \in U_a$, we have $\|X_i - X_j\| < 0.52 \delta$. Therefore, within $\Enavi\big( \{U_\alpha\}_{\alpha \in [\ell]}, W \big)$, we have
\[
\frac{|N(i) \cap U_a|}{|U_a|} \ge \sum_{j \in U_a} \frac{\rp(\|X_i - X_j\|)}{|U_a|} - n^{-1/2+\varsigma} \ge \rp(0.52\delta) - n^{-1/2+\varsigma} > \rp(0.55\delta),
\]
which shows that $i \in W^\natural_a$. We have reached a contradiction.
\end{proof}

\medskip

\begin{proof}[Proof of Proposition \ref{prop: net-check}]
   If $$\bigcup_{\alpha \in [\ell]} W^\natural_{\alpha} \cap W^\flat \neq \varnothing,$$
   then Algorithm~\nc~ will return a pair $(\inx, (V_0, i_0))$.

   In particular, $\inx \in W^\natural_{\alpha} \cap W^\flat$ for some specific $\alpha \in [\ell]$. Let us fix such $\alpha$. In particular, we also have $(V_0,i_0)= (U_\alpha,u_\alpha)$.
   By Lemma \ref{lem: 0.4-delta} and Lemma \ref{lem: 0.6-delta}, we have
    $$\forall \alpha \in [\ell],\, \|X_{\inx} - X_{u_\alpha} \| \ge 0.4 \delta
    \quad \text{and} \quad
    \|X_{\inx} - X_{i_0} \| \le 0.6 \delta .$$
    Furthermore, $(V_0,i_0)$ is a cluster simply follows from $(U_\alpha,u_\alpha)$ is a cluster, since we condition on $\Enc$.

    If $$\bigcup_{\alpha \in [\ell]} W^\natural_{\alpha} \cap W^\flat = \varnothing,$$
    then by Lemma \ref{lem: deltaNet}, we know that $\{X_{u_\alpha}\}_{\alpha \in [\ell]}$ is a $\delta$-net. Therefore, the proposition follows.
\end{proof}

\section{The algorithm \bnet\,and Proof of Theorem~\ref{thm: NetInformal}}\label{sec:proof-main}

In this section, we construct the algorithm \bnet~and prove Theorem~\ref{thm: NetInformal}. We begin by partitioning ${\bf V}$
into $(d+2) \times \lceil n^\varsigma \rceil$ subsets, each of size
$n$:
$$\{W^\ell_\alpha\}_{\ell \in [\lceil n^\varsigma \rceil], \alpha \in [0,d+1]}.$$

For ease of notation, we use the lexicographic order $\{(\ell, k)\}_{\ell \ge 1, k \in [d+1]}$, where $$(\ell',k') \preceq (\ell,k)$$ if and only if one of the following holds:
\begin{itemize}
    \item $\ell' < \ell$, or
    \item $\ell '= \ell$ and $k' \le k$.
\end{itemize}

Given this notation, let us define
\begin{align*}
     W^\ell_{\preceq k} = \bigcup_{(\ell',k') \preceq (\ell,k)} W^{\ell'}_{k'}
     \qquad \mbox{and} \qquad
    Y^\ell_k = \Big(  X_{W^\ell_{\preceq k}}, \eU_{W^\ell_{\preceq k}} \Big).
\end{align*}
(For a sanity check, the induced subgraph on the vertex set $W_{\preceq k}^\ell$ is a function of $Y^\ell_k$.)

\medskip

Our goal is to run Algorithm Algorithm \bnet~ (defined below) to generate a sequence of clusters $$(U_1,u_1),\, (U_2,u_2),\, (U_3,u_3) \dots $$ using the $d+2$ batches of vertices $W^\ell_0, W^\ell_1,\dots, W^\ell_{d+1}$ to generate the $\ell^{\text{th}}$  cluster $(U_\ell,u_\ell)$.
After $\ell$ steps, if the algorithm is not terminated, we expect that
\begin{enumerate}
    \item $\forall s \in [\ell]$, $(U_s,u_s)$ is a cluster.
    \item $\forall \{s,s'\} \in \binom{[\ell]}{2}$, $\|X_{u_s} -X_{u_{s'}}\| \ge 0.3 \delta$.
\end{enumerate}
By a volumetric argument, the second property implies the process must terminate after finitely many steps.
Upon termination, the algorithm produces the desired $\delta$-cluster net with high probability.

\medskip

\noindent
We now give a brief outline of the algorithm, and the complete algorithm is presented in Algorithm~\bnet.

\noindent
\textbf{Initialization.} First, we apply Algorithm \gclu~ with input $W_{d+1}^1$ to get the initial cluster $(U_1,u_1)$. (Note: The subsets $W_k^1$ for $k \in [0,d]$ are unused at this stage, maintained solely for notational consistency.)

\noindent
\textbf{Iteration.}
At the beginning of the $\ell$-th iterations, we have already found $\ell-1$ clusters $\{(U_s,u_s)\}_{s \in [\ell-1]}$. The goal of the $\ell$th iteration is to find the next cluster $(U_{\ell},u_{\ell})$ which is $0.3\delta$-far from the previous clusters, if possible. The $\ell$-th iteration consists of $d+2$ sub-steps:

\noindent \textbf{Sub-step $0$. Pivot vertex for next cluster}
First, we apply Algorithm \nc~ with input $\{(U_s,u_s)\}_{s \in [\ell-1]}$ and the vertex set $W^\ell_0$ to find a vertex $\inx$ satisfying:
\begin{itemize}
    \item $\|X_{u_s} - X_{\inx}\| \ge 0.4 \delta$ for every $s \in [\ell-1]$ and
    \item $\|X_{u_{s'}} - X_{\inx}\| \le 0.6 \delta$ for some $s' \in [\ell-1]$.
\end{itemize}
  If such a vertex $\inx$ exists, then the Algorithm \nc~ returns a pair $(\inx, (V_0, i_0))$ with $(V_0,i_0) = (U_{s'},u_{s'})$.

\noindent \textbf{Sub-step $1$-$d$. Forming an orthogonal cluster frame}
If Algorithm \nc~ returns a pair $(\inx, (V_0, i_0))$, we then perform the next $d$ sub-steps in the $\ell$th iterations to form an othogonal" set of clusters centered at $(V_0,i_0)$.  Within this subroutine, at the $k$th sub-step, Algorithm \gclu~ was applied with a subset of vertices $W'^\ell_k \subset W^\ell_k$ was chosen subject to constraints coming from previously generated clusters $(V_0,i_0), (V_1,i_1), \dots, (V_{k-1},i_{k-1})$, and the output is a new cluster $(V_k,i_k)$.

\noindent \textbf{Sub-step $d+1$. Extract the next cluster}
Finally, in the $d+2$th sub-step, we use Algorithm \gclu~ with input $W^\ell_{d+1}$ with constraints coming from the clusters $(V_0,i_0), (V_1,i_1), \dots, (V_{d},i_{d})$. This produces the next desired cluster $(U_{\ell},u_{\ell})$.

\noindent \textbf{Termination.}
If at any iteration Algorithm \nc~ returns ${\rm null}$, then we conclude that $\{(U_s,u_s)\}_{s \in [\ell]}$ is a $\delta$-net of clusters.

We will demonstrate that each sub-step of this process succeeds with probability at least $1 - n^{-\omega(1)}$. Additionally, as the algorithm performs at most polynomially many iterations (in $n$), it terminates successfully with overall probability $1 - n^{-\omega(1)}$.

\begin{algorithm}
\caption{\bnet}
\label{alg:buildNet}

\SetKwInOut{Input}{Input}
\SetKwInOut{Output}{Output}

\Input{
    $\{W^s_\alpha\}_{s \in [\lceil n^\varsigma \rceil], \alpha \in [0,d+1]}$.
}
\Output{
    $\{(U_s,u_s)\}_{s\in [\ell]}$.
}

$(U_1, u_1) \gets  \gclu(W^1_{d+1},W^1_{d+1})$; \\
$\ell \gets 2$; \\
$(\inx^\ell, (V^\ell_0,i^\ell_0)) \gets \nc((U_1,u_1),W^{\ell}_0)$; \\

\While{$(\inx^\ell, (V^\ell_0,i^\ell_0)) \neq {\rm null}$}
{
    $k \gets 1$\;

\While{$k \le d$}{

    $W'^\ell_k \gets$ the set of all vertices $i \in W^\ell_k$ such that
    \[
    \rp(\sqrt{2}r + 0.95 \delta) \le \frac{|N(i) \cap V^\ell_\alpha|}{|V^\ell_\alpha|} \le \rp(\sqrt{2}r-0.95\delta),
    \]
    for every $\alpha \in [k-1]$, and such that
    \[
    \rp(r+0.95\delta) \le \frac{|N(i) \cap V^\ell_0|}{|V^\ell_0|} \le \rp(r-0.95\delta);
    \]
    $(V^\ell_k, i^\ell_k) \gets \gclu(W^\ell_k,W'^\ell_k)$\;
    $k \gets k+1$\;
}

$W^{\sharp\,\ell} \gets$ the set of all vertices $i \in W^\ell_{d+1}$ such that
\[
\bigg|\frac{|N(\inx) \cap V^\ell_\alpha|}{|V^\ell_\alpha|} - \frac{|N(i_0) \cap V^\ell_\alpha|}{|V^\ell_\alpha|} \bigg| \le  c_3 \delta,
\]
for every $\alpha \in [d]$, and such that
\[
\frac{|N(i)\cap V^\ell_0|}{|V^\ell_0|} \ge \rp(2 \delta );
\]
\noindent
$(U_\ell, u_\ell) \gets \gclu(W^\ell_{d+1},W^{\sharp\,\ell})$\;
   $\ell \gets \ell+1$; \\
    $(\inx^\ell, (V^\ell_0,i^\ell_0)) \gets \nc(\{(U_s,u_s)\}_{s \in [\ell -1 ]},W^\ell_0)$; \\
}

\Return{$\{(U_s,u_s)\}_{s\in [\ell]}$}
\end{algorithm}
\paragraph{Complexity}
 At the $\ell$th iteration, computing $|N(i) \cap V^\ell_\alpha|$ for all $i \in W^\ell_k$ and $\alpha \in [0, k-1]$ takes $O((d+1) \cdot n^2)$ time. Each call to Algorithm \gclu\, runs in $O(n^3)$ time \eqref{eq: gclu-runningtime}. Additionally, invoking Algorithm \nc~ costs $O(\ell n^2)$ (see \eqref{eq: cnet-runningtime}) per iteration due to comparisons against all existing clusters. Summing across $\ell$ iterations, the total cost of running $\ell$ iterations becomes
\begin{align}
    \label{eq: bnet runningtime}
    O(\ell n^3 + \ell^2 n^2)\,.
\end{align}

Associated with this algorithm, we define the following events.
For each integer $\ell \ge 2$,
\begin{align*}
{\cal E}^\ell_k := \begin{cases}
   \Enc\big( \{(U_s,u_s)\}_{s \in [\ell- 1]}, W^\ell_0\big)
   & k=0, \\
   \Eao\big( \{(V^\ell_\alpha, i^\ell_\alpha)\}_{\alpha \in [0,k-1]}, W^\ell_k \big)
   & k \in [d], \\
   \Enx\big( \{(V^\ell_\alpha,i^\ell_\alpha)\}_{\alpha \in [0,d]}, i^\ell_{\rm nx}, W^\ell_{d+1}\big)
   & k=d+1.
\end{cases}
\end{align*}
Recall that $\Enc$, $\Eao$, and $\Enx$ are defined in \eqref{eq: Enc}, \eqref{eq: Eao}, and \eqref{eq: Enx}, respectively. From the derivation in the previous sections, each event ${\cal E}_k^\ell$ ensures that the corresponding sub-algorithms in Algorithm \bnet~ produce the expected outcome. Here we list them for the reader's convenience:
\begin{align}
\label{eq: Enc2}
    &\Enc\big( \{(U_s,u_s)\}_{s \in [\ell- 1]}, W^\ell_0\big)  \\
    :=\,&
    \nonumber
    \Eclu\Big(\{(U_\alpha, u_\alpha)\}_{\alpha \in [\ell-1]}\Big)
    \cap \Enavi\big( \{U_\alpha\}_{\alpha \in [\ell-1]}, W^\ell_0 \big)
    \cap \Enet(W^\ell_0)\cap
    {\cal E}_{\rm rps}(\{u_\alpha\}_{\alpha \in [\ell]}) \,; \\
\label{eq: Eao2}
    &\Eao\big( \{(V^\ell_\alpha, i^\ell_\alpha)\}_{\alpha \in [0,k-1]}, W^\ell_k \big)  \\
    \nonumber
    :=\, &
        {\cal E}_{\rm ao}(i^\ell_0, \ldots, i^\ell_{k-1}) \cap
         \Eclu(\{(V^\ell_\alpha, i^\ell_\alpha)\}_{\alpha \in [0,k-1]})
         \cap \Enavi(\{V^\ell_\alpha\}_{\alpha \in [0,k-1]}, W^\ell_k) \cap \Ecn(W^\ell_k) \cap \Enet(W^\ell_k)
        \,;\\
\label{eq: Enx2}
        &\Enx\big( \{(V^\ell_\alpha,i^\ell_\alpha)\}_{\alpha \in [0,d]}, i^\ell_{\rm nx}, W^\ell_{d+1}\big) \\
        \nonumber
        :=\,&
            \Eclu(\{(V^\ell_\alpha, i^\ell_\alpha)\}_{\alpha \in [0,d]})
            \cap {\cal E}_{\rm ao}(i^\ell_0, i^\ell_1, \ldots, i^\ell_d)
            \cap \big\{ 0.4 \delta \le \|X_{\inx^\ell}-X_{i^\ell_0}\| \le 0.6 \delta \big\}
            \cap \Enavi(\{V^\ell_\alpha\}_{\alpha \in [0,d]}, W^\ell_{d+1})  \\
        \nonumber
            &\cap \Enavi(\{V^\ell_\alpha\}_{\alpha \in [d]}, \{i^\ell_{\rm nx}\})
            \cap \Enet(W^\ell_{d+1})\,.
        \end{align}

Note that these events can be ambiguous since
 $\{(U_s,u_s)\}_{s\in [\ell-1]}$ or $\{(V^\ell_\alpha, i^\ell_\alpha)\}_{\alpha \in [k-1]}$ might be null. In that case, we simply treat the event as empty empty.
 We now restate key propositions clearly summarizing the implications of these events.
 \begin{proposition}[Proposition~\ref{prop: net-check}]
    \label{prop: net-check2}
    Suppose ${\cal E}^\ell_0 = \Enc\big( \{(U_s,u_s)\}_{s \in [\ell- 1]}, W^\ell_0\big)$ is not empty for some configuration $\{(U_s,u_s)\}_{s \in [\ell- 1]}$.
    Condition on ${\cal E}^\ell_0$.
    Running Algorithm~\nc~ with input $\{(U_s,u_s)\}_{s \in [\ell- 1]}, W^\ell_0$
    will result in one of the following two cases:
    \begin{itemize}
        \item If it returns a pair $(\inx, (V_0, i_0))$, then
            \begin{align*}
                {\cal E}^\ell_0
            \subseteq
                \big\{ \forall \alpha \in [\ell],\, \|X_{\inx} - X_{u_\alpha} \| \ge 0.4 \delta \big\}
             \cap
                \big\{ \|X_{\inx} - X_{i_0} \| \le 0.6 \delta  \big\}
            \cap
                \Eclu((V_0,i_0)).
            \end{align*}
        \item If it returns ${\rm null}$, then
        $\{X_{u_\alpha}\}_{\alpha \in [\ell]}$ is a $\delta$-net of $M$.
    \end{itemize}
\end{proposition}

\begin{proposition} [Proposition~\ref{prop:nextnavicluster}]
    \label{prop:nextnavicluster2}
    Suppose ${\cal E}_k^\ell = \Eao\big( \{(V^\ell_\alpha, i^\ell_\alpha)\}_{\alpha \in [0,k-1]}, W^\ell_k \big)$ is not empty for some choices of $\{(V^\ell_\alpha, i^\ell_\alpha)\}_{\alpha \in [k-1]}$ and $W^\ell_k$.
    Let $W'^\ell_k$ defined according to \eqref{eq: W'orthogonal}:
    \begin{align}
         W'^\ell_k := \Big\{ i \in W^\ell_k\,:\,  \forall \alpha \in [k-1], \, \, & \rp(\sqrt{2}r+0.95\delta) \le |N(i) \cap V^\ell_\alpha|/|V^\ell_\alpha| \le \rp( \sqrt{2}r-0.95\delta)\\
        \nonumber
        & \text{and} \quad \rp(r+0.95\delta)  \le |N(i) \cap V^\ell_0|/ |V^\ell_0| \le \rp( r-0.95\delta)
        \Big\}.
    \end{align}
    Condition on ${\cal E}_k^\ell$. If we run the Algorithm~\gclu~ with input $(W^\ell_k,W'^\ell_k)$, then its output, denoted as $(V^\ell_k,i^\ell_k)$, is a cluster and $(v_0,\ldots,v_k)$ satisfies the event ${\cal E}_{\rm ao}(i_0,\ldots,i_k)$.
      In other words,
      \begin{align*}
            {\cal E}_k^\ell
        \subseteq
            \Eclu( V^\ell_k,  i^\ell_k)
            \cap {\cal E}_{\rm ao}(i_0,\ldots,i_k).
      \end{align*}
    \end{proposition}

\begin{proposition}[Proposition \ref{prop: NEXTcore}]
    \label{prop: NEXTcore2}
Suppose ${\cal E}_{d+1}^\ell = \Enx\big( \{(V^\ell_\alpha,i^\ell_\alpha)\}_{\alpha \in [0,d]}, i^\ell_{\rm nx}, W^\ell_{d+1}\big)$ is not empty for some choices of $\{(V^\ell_\alpha,i^\ell_\alpha)\}_{\alpha \in [0,d]}$ and $W^\ell_{d+1}$.
Let
\begin{align*}
W^{\sharp\,\ell}_{d+1} := \bigg\{  i \in W^\ell_{d+1} \, : \,
& \forall \alpha \in [d],
\bigg|\frac{|N(i) \cap V^\ell_\alpha|}{|V^\ell_\alpha|} - \frac{|N(i^\ell_{\rm nx}) \cap V_\alpha|}{|V_\alpha|} \bigg|
\le  c_3 \delta \\
& \mbox{ and }
\frac{|N(i)\cap V_0|}{|V_0|} \ge \rp(2 \delta )
\bigg\}.
\end{align*}
Condition on ${\cal E}_{d+1}^\ell$. If we run the Algorithm~\gclu~ with input  $(W^\ell_{d+1},W^{\sharp\,\ell}_{d+1})$, then its output, denoted as \( (U_{\ell}, u_\ell) \), is a cluster and  $\|X_{u_\ell} - X_{i^\ell_{\rm nx}}\| \le 0.1\delta$.
\end{proposition}

Furthermore, for ease of notation, we also define ${\cal E}_k^1$ to be the trivial event for $k \in [d]$, and
$$
    {\cal E}_{d+1}^1
:=
    \Ecn(W_{d+1}^1)
        \cap \Enet(W_{d+1}^1),
$$
which guarantees that the $\gclu(W^1_{d+1},W^1_{d+1})$
returns a cluster $(U_1,u_1)$, by Proposition \ref{prop: 1cluster}. For $\ell \ge 2$, let
    $$
        {\cal E}^{\ell}_{\rm halt} :=
 \Big\{\nc(\{(U_s,u_s)\}_{s \in [\ell -1 ]},W^\ell_{0})  = {\rm null}\Big\}.
    $$

We define the intersection of such events according to the $\preceq$ order. For $\ell \ge 2$ and $k \in [d+1]$, we define
\begin{align*}
    \Omega^\ell_k = \bigcap_{(\ell',k') \preceq (\ell,k)} {\cal E}^{\ell'}_{k'}.
\end{align*}
Notice that we also have the relation that, for $\ell \ge 2$,
$$
    \Omega_1^{\ell} \subseteq \Omega_0^\ell \cap ({\cal E}^\ell_{\rm halt})^c,
$$
since $\Omega_1^{\ell}$ requires $(V_0^\ell,i_0^\ell)$ is non-null.

Let us rephrase the Theorem~\ref{thm: NetInformal} here for convenience and in terms of the parameters introduced in Section~\ref{sec: RGM} and Algorithm~\bnet.

\begin{theor}[Theorem~\ref{thm: NetInformal}, restated] \label{thm: net}
Let $(M, \mu, \rp)$ satisfy Assumptions~\ref{assump: M-and-pm} and \ref{assump: pdf} from Section~\ref{sec: RGM}.
    Then with probability at least $1-n^{-\omega(1)}$, applying Algorithm \bnet~with input $G({\bf V}, M, \mu, \rp)$ returns a $(\delta,\eta)$-cluster-net $\{(U_s,u_s)\}_{s \in [\ell]}$ with $u_s \in U_s \subseteq {\bf V}$ and $\ell \le n^{3\varsigma/4}$
    in the sense that
    \begin{enumerate}
        \item $\forall s \in [\ell]$, $(U_s,u_s)$ is a cluster, and
        \item $\{X_{u_s}\}_{s \in [\ell]}$ is a $\delta$-net of $M$.
    \end{enumerate}
    Furthermore, the running time of the algorithm is bounded by
    $$
        O(n^{3+ 3\varsigma/4}) = O(|{\bf V}|^{3}).
    $$
\end{theor}
\begin{remark}
    The running time $O(n^{3+ 3\varsigma/4})$ is better than what is stated in Theorem~\ref{thm: NetInformal} since we renomralized the size of the graph to have $(d+2)n^{1+ \varsigma}$ vertices. Further, running time bounded simply follows from $\ell \le n^{3\varsigma/4}$ and the fact that Algorithm \bnet~ has a running time of $O(\ell n^3 + \ell^2 n^2)$ as shown in \eqref{eq: bnet runningtime} for $\ell$ iterations.
\end{remark}

Let
\begin{align}\label{eq: netFound}
    \Omega_{\rm netFound} \end{align}
denote the event that the output of Algorithm \bnet\, is a $(\delta,\eta)$-cluster-net with $\ell \le n^{3\varsigma/4}$.
Observe that, by Proposition \ref{prop: net-check2}, we have
$$
    \Omega_{\rm netFound} \supseteq \bigcup_{2 \le \ell \le n^{3\varsigma/4}} \Omega_0^{\ell} \cap {\cal E}^\ell_{\rm halt}.
$$
Our goal is to show
$$
\mathbb{P} \Big(\bigcup_{2 \le \ell \le n^{3\varsigma/4}} \Omega_0^{\ell} \cap {\cal E}^\ell_{\rm halt} \Big) = 1 - n^{-\omega(1)}\,.
$$
At this point, the proof of Theorem~\ref{thm: net} becomes a routine exercise using standard tools. The key events $\Ecn,\Enet,\Eclu,$ and \(\Enavi\) all occur with overwhelming probability, thanks to standard $\varepsilon$-net arguments and Hoeffding-type concentration inequalities. Moreover, each proposition (Propositions \(\ref{prop: net-check2}\), \(\ref{prop:nextnavicluster2}\), \(\ref{prop: NEXTcore2}\)) enforces a conditional structure: {\it if the event ${\cal E}_k^\ell$ in the previous step holds, then the next event also holds with high probability.} Consequently, the proof of Theorem~\ref{thm: net} follows by stringing these propositions together, with a few additional remarks on (conditional) independence to justify when applying the relevant bounds successively.

\subsection{Probability Estimates for the Events ${\cal E}^\ell_k$}
Let us begin with stating the probability estimates for the basic events. We postpone the proof of the three lemmas below to Appendix \ref{appx: basic result proof}.

\begin{lemma} \label{lem: cn}
    For any $W \subseteq {\bf V}$ with $|W|=n$, we have $\mathbb{P} \{ (\Ecn(W))^c \} = n^{-\omega(1)}$.
\end{lemma}

\begin{lemma} \label{lem: epsilonNetEvent}
For any $W \subseteq {\bf V}$ with $|W| = n$, we have
\begin{align}
   \Prob \big\{
       (\Enet(W))^c
   \big\} \le \exp( - n^{1/2}).
\end{align}
\end{lemma}

\begin{lemma} \label{lem: Enavi}
Let $V,W$ be a pair of disjoint subset of ${\bf V}$.
For a positive integer $k$, suppose $\{(V_\alpha, i_\alpha)\}_{[0,k]}$ are $k+1$ random pairs with $i_\alpha \in V_\alpha \subseteq V$, which are functions of $X_V \cup \eU_V$.
For any realization $(X_V, \eU_V) = (x_V, {\rm u}_V)$ with $(x_V,{\rm u}_V) \in \Eclu(\{(V_\alpha, i_\alpha)\}_{\alpha \in [0,k]})$ and any realization of $X_W=x_w$,
\begin{align*}
     \mathbb{P} \Big\{
        (\Enavi(\{V_\alpha\}_{\alpha \in [0,k]}, W))^c
        \,\Big|\,
        X_V = x_V, \eU_V = {\rm u}_V, X_W = x_w
    \Big\}
=
    n^{-\omega(1)}.
\end{align*}
In particular, the above inequality holds without conditioning on $X_W=x_w$.

\end{lemma}

\noindent
Next, we proceed to the probability estimates of the main events ${\cal E}^\ell_k$ defined in this section.

\begin{lemma} \label{lem: eventlk}
   For $\ell \ge 2$ and $k \in [d]$,  if $\mathbb{P}\{\Omega_k^\ell\}>0$, then
    \begin{align*}
        \Prob\{ ({\cal E}^\ell_{k+1})^c \, \vert \, \Omega^\ell_k \} =  n^{-\omega(1)}.
    \end{align*}
    In particular,
    $\mathbb{P}\{ \Omega_{k+1}^{\ell}\} = \mathbb{P}\{\Omega_k^\ell\} (1 - n^{-\omega(1)}) >0$.
\end{lemma}
\begin{proof}
    Let us consider the case $k \in [d-1]$, since in the case $k=[d]$, the definition of ${\cal E}^\ell_{d+1}$ is slightly different.

    \noindent
    \underline{Case 1: $k \in [d-1]$.}
     Recall that by definition from \eqref{eq: Eao2},
    \begin{align*}
        {\cal E}_{k+1}^\ell
    =&
        \Eao\big( \{(V^\ell_\alpha, i^\ell_\alpha)\}_{\alpha \in [0,k]}, W^\ell_{k+1} \big)\\
    =&
        {\cal E}_{\rm ao}(i_0^\ell, \ldots, i_k^\ell)
        \cap \Eclu(\{(V_\alpha^\ell, i_\alpha^\ell)\}_{\alpha \in [0,k]})
        \cap \Enavi(\{V_\alpha^\ell\}_{\alpha \in [0,k]}, W_{k+1}^\ell)
        \cap \Ecn(W_{k+1}^\ell)
        \cap \Enet(W_{k+1}^\ell).
    \end{align*}

    By Proposition~\ref{prop:nextnavicluster2}, we have
    \begin{align*}
         \Omega^\ell_k
    \subseteq
        {\cal E}^\ell_k
    \subseteq
        {\cal E}_{\rm ao}(i_0^\ell,\dots, i_{k}^\ell)
    \cap
        \Eclu(V_{k}^\ell, i_{k}^\ell),
    \end{align*}
    and in particular,
    \begin{align}
        \label{eq: omega1-d00}
         \Omega^\ell_k
    \subseteq
        {\cal E}_{\rm ao}(i_0^\ell, \ldots, i_k^\ell )
        \cap \Eclu(\{(V_\alpha^\ell, i_\alpha^\ell)\}_{\alpha \in [0,k]}),
    \end{align}
    since ${\cal E}^\ell_k \subseteq
    \Eclu(\{(V_\alpha^\ell, i_\alpha^\ell)\}_{\alpha \in [0,k-1]})$.
  Consequently, on the event \(\Omega_k^\ell\), the occurrences of
  \({\cal E}_{\mathrm{ao}}\bigl(i_0^\ell,\ldots,i_k^\ell\bigr)\) and
  \(\Eclu\bigl(\{(V_\alpha^\ell, i_\alpha^\ell)\}_{\alpha \in [0,k]}\bigr)\) are already guaranteed. Hence,
    \begin{align*}
        \Prob\{ ({\cal E}^\ell_{k+1})^c \, \vert \, \Omega^\ell_k \}
    =
        \Prob \Big\{
            (\Enavi(\{V_\alpha^\ell\}_{\alpha \in [0,k]}, W_{k+1}^\ell))^c
            \cup (\Ecn(W_{k+1}^\ell))^c
            \cup (\Enet(W_{k+1}^\ell))^c
        \, \Big\vert \, \Omega^\ell_k \Big\}.
    \end{align*}

    Next, let us fix any realization $y^\ell_k \in \Omega^\ell_k$.
    Since \(y^\ell_k \in \Eclu\bigl(\{(V_\alpha^\ell, i_\alpha^\ell)\}_{\alpha \in [0,k]}\bigr)\),
    we may apply Lemma~\ref{lem: Enavi} to conclude that
    \begin{align*}
        \Prob \big\{ (\Enavi(\{V_\alpha^\ell\}_{\alpha \in [0,k]}, W_{k+1}^\ell))^c
        \,\big\vert\, Y^\ell_k = y^\ell_k  \big\}
    \le
        n^{-\omega(1)}.
    \end{align*}
    Notice that $\Enavi(\{V_\alpha^\ell\}_{\alpha \in [0,k]}, W_{k+1}^\ell)$ depends only on the tuple
    $$\Big(Y^\ell_k, X_{W^\ell_{k+1}}, \eU_{W^\ell_{\preceq k}, W_{k+1}^\ell} :=(\eU_{i,j})_{i \in W^\ell_{\preceq k}, j \in W_{k+1}^\ell} \Big)\,.$$
    Then, by applying Fubini's theorem, we have
    \begin{align*}
     &  \Prob \big\{ (\Enavi(\{V_\alpha^\ell\}_{\alpha \in [0,k]}, W_{k+1}^\ell))^c
        \,\big\vert\,  Y_k \in \Omega_k^{\ell}  \big\}\\
    =
        & \mathbb{E} \Big[ {\bf 1}_{(\Enavi(\{V_\alpha^\ell\}_{\alpha \in [0,k]}, W_{k+1}^\ell))^c }
        \big( Y^\ell_k, X_{W^\ell_{k+1}}, \eU_{W^\ell_{\preceq k}, W_{k+1}^\ell} \big)
            \,\Big\vert\,  Y^\ell_k \in \Omega^\ell_k \Big] \\
    =
        & \mathbb{E} \Big[ \mathbb{E}_{X_{W^\ell_{k+1}}, \eU_{W^\ell_{\preceq k}, W_{k+1}^\ell}}
            \big[{\bf 1}_{(\Enavi(\{V_\alpha^\ell\}_{\alpha \in [0,k]}, W_{k+1}^\ell))^c }
            \big( Y^\ell_k, X_{W^\ell_{k+1}}, \eU_{W^\ell_{\preceq k}, W_{k+1}^\ell} \big) \big]
            \,\Big\vert\,  Y^\ell_k \in \Omega^\ell_k \Big] \\
    =
        &\mathbb{E} \Big[ n^{-\omega(1)} \,\Big\vert\,  Y^\ell_k \in \Omega^\ell_k \Big]
    =
        n^{-\omega(1)}.
    \end{align*}

    On the other hand, the events
    \(\Ecn(W_{k+1}^\ell)\) and \(\Enet(W_{k+1}^\ell)\)
    do not depend on \(Y^\ell_k\). Hence, by Lemma~\ref{lem: cn} and Lemma~\ref{lem: epsilonNetEvent}, we have
    \begin{align*}
         \Prob \big\{ (\Ecn(W_{k+1}^\ell))^c \cup (\Enet(W_{k+1}^\ell))^c
        \,\big\vert\,  Y_k \in \Omega_k^{\ell}  \big\}
    = &
        \Prob \big\{ (\Ecn(W_{k+1}^\ell))^c \cup (\Enet(W_{k+1}^\ell))^c \}
    =
        n^{-\omega(1)}.
    \end{align*}
    Combining these estimates and recalling \eqref{eq: omega1-d00}, it follows that
    \begin{align*}
        \Prob\{ ({\cal E}^\ell_{k+1})^c \, \vert \, \Omega^\ell_k \}
    \overset{\eqref{eq: omega1-d00}}{=}
        \Prob\big\{
            (\Enavi(\{V_\alpha^\ell\}_{\alpha \in [0,k]}, W_{k+1}^\ell))^c
            \cup (\Ecn(W_{k+1}^\ell))^c
            \cup (\Enet(W_{k+1}^\ell))^c
        \, \big\vert \, \Omega^\ell_k \big\}
    =
        n^{-\omega(1)}.
    \end{align*}

    \smallskip

    \noindent
    \underline{Case 2: $k=d$.}
    Recall from \eqref{eq: Enx2} that
    \begin{align*}
    {\cal E}_{d+1}^\ell\, = \,& \Enx\big( \{(V^\ell_\alpha,i^\ell_\alpha)\}_{\alpha \in [0,d]}, i^\ell_{\rm nx}, W^\ell_{d+1}\big) \\
    \nonumber
    :=\,&
        \Eclu(\{(V^\ell_\alpha, i^\ell_\alpha)\}_{\alpha \in [0,d]})
        \cap {\cal E}_{\rm ao}(i^\ell_0, i^\ell_1, \ldots, i^\ell_d)
        \cap \big\{ 0.4 \delta \le \|X_{\inx^\ell}-X_{i^\ell_0}\| \le 0.6 \delta \big\}
        \cap \Enavi(\{V^\ell_\alpha\}_{\alpha \in [0,d]}, W^\ell_{d+1})  \\
    \nonumber
        &\cap \Enavi(\{V^\ell_\alpha\}_{\alpha \in [d]}, \{i^\ell_{\rm nx}\})
        \cap \Enet(W^\ell_{d+1})\,.
    \end{align*}
    Following a similar argument as in the case \(k \in [d-1]\), we obtain
    \begin{align*}
        \Prob\{ ({\cal E}_{d+1}^\ell)^c \, \vert\, \Omega_d^\ell \}
    \le  &
        \Prob\big\{
            (\Enavi(\{V_\alpha^\ell\}_{\alpha \in [d]}, \{\inx^\ell\}))^c
        \big\}
    +
    \Prob\big\{
        (\Enavi(\{V_\alpha^\ell\}_{\alpha \in [d]}, W_{d+1}^\ell))^c
    \big\}  +
        \Prob\big\{
            \|X_{\inx^\ell}-X_{i^\ell_0}\| \notin [0.4\delta,\,0.6\delta]
        \big\} \\
    \le &
    \Prob\big\{
        (\Enavi(\{V_\alpha^\ell\}_{\alpha \in [d]}, \{\inx^\ell\}))^c
    \big\}
    + \Prob\big\{
        \|X_{\inx^\ell}-X_{i^\ell_0}\| \notin [0.4\delta,\,0.6\delta]
    \big\} + n^{-\omega(1)}\,.
    \end{align*}
    It remains to show the first two ummands terms on the right are also $n^{-\omega(1)}$.
    Note that $\mathbb{P}\{\Omega^\ell_d\}>0$. Within the event $\Omega_{d}^\ell$, the triple
 $$(\inx^\ell, V_0^\ell, i_0^\ell)$$
 is a valid output of \nc$(\{(U_s,u_s)\}_{s \in [\ell- 1]}, W^\ell_0)$. By Proposition \ref{prop: net-check2}, this implies
 $$
    \Omega_d^\ell \subseteq {\cal E}_0^\ell
\subseteq
\big\{ 0.4 \delta \le \|X_{\inx^\ell}-X_{i^\ell_0}\| \le 0.6 \delta \big\}\,.
 $$
 Further, since $\inx^\ell \in W_0^\ell$ by the construction of Algorithm~\nc, we have
 $$
    \Omega_d^\ell
\subseteq
    {\cal E}_{0}^{\ell}
\subseteq
    \Enavi(\{U_s\}_{s \in [\ell-1]}, W_0^\ell)
\subseteq
    \Enavi(\{U_s\}_{s \in [\ell-1]}, \inx^\ell).
 $$
 Consequently,
 \[
  \Prob\Bigl\{\|X_{i_{\mathrm{nx}}^\ell}-X_{i_0^\ell}\|\notin[0.4\delta,\,0.6\delta]\Bigr\}
  \;=\;
  \Prob\Bigl\{
    \bigl(\Enavi(\{V_\alpha^\ell\}_{\alpha\in[d]}, \{i_{\mathrm{nx}}^\ell\})\bigr)^c
    \,\Big\vert\,
    \Omega_d^\ell
  \Bigr\}
  \;=\;
  0.
\]
This shows that both of those probabilities are zero when conditioned on \(\Omega_d^\ell\).  Hence,
\[
  \Prob\Big\{(\mathcal{E}_{d+1}^\ell)^c \,\Big\vert\, \Omega_d^\ell\Big\}
  \;=\;
  n^{-\omega(1)},
\]
which completes the proof for the case $k=d$.
\end{proof}

\begin{lemma} \label{lem: eventld}
    For $\ell \ge 1$, if $\mathbb{P}\{\Omega^\ell_{d+1}\}>0$, then
\begin{align*}
    \Prob \Big\{ ({\cal E}^{\ell+1}_{0})^c \,
        \Big\vert \, \Omega^\ell_{d+1}
    \Big\}
=
    n^{-\omega(1)}.
\end{align*}
In particular, $\mathbb{P}\{\Omega^{\ell+1}_{0}\}>0$.
\end{lemma}

\begin{proof}

We will consider the case $\ell \ge 2$ only.
(The case $\ell=1$ follows by a similar but simpler argument.)

Recall from \eqref{eq: Enc2} that
\begin{align*}
{\cal E}_{0}^{\ell+1} \,=\,
    &\Enc\big( \{(U_s,u_s)\}_{s \in [\ell]}, W^{\ell+1}_0\big)  \\
    :=\,&
    \nonumber
    \Eclu\Big(\{(U_\alpha, u_\alpha)\}_{\alpha \in [\ell]}\Big)
    \cap \Enavi\big( \{U_\alpha\}_{\alpha \in [\ell]}, W^{\ell+1}_0 \big)
    \cap \Enet(W^{\ell+1}_0)\cap
    {\cal E}_{\rm rps}(\{u_\alpha\}_{\alpha \in [\ell]}) \,;
\end{align*}
By the union bound,
\begin{align}
\label{eq: eventld00}
    \Prob\{ ({\cal E}^{\ell+1}_{0})^c \, \vert \, \Omega^\ell_{d+1} \}
\le &
    \Prob \Big\{
        \Big(\Eclu\big(
            \{(U_s, u_s)\}_{s \in [\ell]}\big)\Big)^c
        \, \Big \vert \, \Omega^\ell_{d+1}
    \Big\}
+
    \Prob \big\{
        \big({\cal E}_{\rm rps}(\{i_s\}_{s \in [\ell]})\big)^c
        \, \big \vert \, \Omega^\ell_{d+1}
    \big\}\\
\nonumber
&+
    \Prob\Big\{
    \big(\Enavi\big( \{U_s \}_{s \in [\ell]}, W_0^{\ell+1} \big) \big)^c
        \, \big \vert \, \Omega^\ell_{d+1}
    \Big\}
+
    \Prob\big\{
        \big(\Enet(W_0^{\ell+1})\big)^c
        \, \big \vert \, \Omega^\ell_{d+1}
    \big\}.
\end{align}

\noindent
Next, observe that
$$ \varnothing \neq \Omega^\ell_{d+1}  \subseteq {\cal E}^\ell_0 \subseteq
\Eclu\Big(\{(U_s, u_s)\}_{s \in [\ell-1]}\Big)
\cap
{\cal E}_{\rm rps}(\{u_s\}_{s \in [\ell-1]}).
$$
From \eqref{eq: eventRps}, we recall that
\begin{align*}
    {\cal E}_{\rm rps}(\{u_\alpha\}_{\alpha \in [\ell-1]}) := \bigg\{
    \forall \{\alpha, \beta\} \in \binom{[\ell]}{2}, \, \|X_{u_\alpha} - X_{_\beta}\| \ge 0.3 \delta
    \bigg\}.
\end{align*}
Therefore, we may replace the first two terms on the right-hand side of \eqref{eq: eventld00}, yielding
\begin{align}
\label{eq: eventld01}
    \Prob\{ ({\cal E}^{\ell+1}_{0})^c \, \vert \, \Omega^\ell_{d+1} \}
\le &
    \Prob \big\{
        \big(\Eclu\big(U_{\ell }, u_{\ell }\big)\big)^c
        \, \big \vert \, \Omega^\ell_{d+1}
    \big\}
    +
    \Prob \big\{
        \exists s \in [\ell],\, \|X_{i_s} - X_{i_{\ell }}\| < 0.3 \delta
        \, \big \vert \, \Omega^\ell_{d+1}
    \big\}\\
\nonumber
&+
    \Prob\Big\{
    \big(\Enavi\big( \{U_s \}_{s \in [\ell]}, W_0^{\ell+1} \big) \big)^c
        \, \big \vert \, \Omega^\ell_{d+1}
    \Big\}
+
    \Prob\big\{
        \big(\Enet(W_0^{\ell+1})\big)^c
        \, \big \vert \, \Omega^\ell_{d+1}
    \big\}.
\end{align}
It remains to show each of the summand above is  $n^{-\omega(1)}$.
First, observe that Proposition \ref{prop: NEXTcore2} implies
\begin{align}
\label{eq: eventld02}
    \Omega_{d+1}^\ell \subseteq {\cal E}_{d+1}^\ell
    \subseteq \Eclu(U_{\ell }, u_{\ell }) \cap \{ \|X_{u_{\ell}} - X_{\inx^\ell}\| \le 0.1\delta     \}.
\end{align}
An immediate consequence is that the first summand in \eqref{eq: eventld01} vanishes:
$$    \Prob \big\{
        \big(\Eclu\big(U_{\ell }, u_{\ell }\big)\big)^c
        \, \big \vert \, \Omega^\ell_{d+1}
    \big\} = 0.
$$

\noindent
Next, note that $\Omega_{d+1}^\ell \neq \varnothing$. Within the event $\Omega_{d+1}^\ell$, the triple
 $$(\inx^\ell, V_0^\ell, i_0^\ell)$$
 is a non-null output of \nc$(\{(U_s,u_s)\}_{s \in [\ell- 1]}, W^\ell_0)$.
 By Proposition~\ref{prop: net-check2}, this ensures
 $$
     \forall s \in [\ell-1],\,
     \|X_{\inx^\ell} - X_{u_s}\| \ge 0.4\delta.
 $$

Using the triangle inequality and the above estimate, we get for each $s \in [\ell-1]$,
\begin{align*}
    \| X_{u_\ell} - X_{u_s} \|
\ge
    \| X_{\inx^\ell} - X_{u_s} \|
    - \| X_{u_\ell} - X_{\inx^\ell} \|
\overset{\eqref{eq: eventld02}}{\ge}
    0.4\delta - 0.1 \delta
\ge
    0.3 \delta,
\end{align*}
which implies the second summand in \eqref{eq: eventld01} also vanishes:
$$
    \Prob \big\{
        \exists s \in [\ell],\, \|X_{i_s} - X_{i_{\ell }}\| < 0.3 \delta
        \, \big \vert \, \Omega^\ell_{d+1}
    \big\} = 0.
$$

\noindent
Finally, the remaining estimate
$$
\Prob\Big\{
    \big(\Enavi\big( \{U_s \}_{s \in [\ell]}, W_0^{\ell+1} \big) \big)^c
        \, \big \vert \, \Omega^\ell_{d+1}
    \Big\}
+
    \Prob\big\{
        \big(\Enet(W_0^{\ell+1})\big)^c
        \, \big \vert \, \Omega^\ell_{d+1}
    \big\} = n^{-\omega(1)}
$$
follows by the same argument used
in the proof of Lemma \ref{lem: eventlk}
and is therefore omitted.
Combining these observations completes the proof of the lemma.
\end{proof}

\begin{lemma} \label{lem: eventl0}
    For $\ell \ge 2$, if $\mathbb{P}\{\Omega^\ell_{0}\mbox{ and
        \nc}(\{(U_s,u_s)\}_{s \in [\ell -1 ]},W^\ell_{0}) \neq {\rm null} \}>0$, then
\begin{align*}
    \Prob \Big\{ ({\cal E}^{\ell+1}_{1})^c \,
        \Big\vert \, \Omega^\ell_{0} \mbox{ and
        \nc}(\{(U_s,u_s)\}_{s \in [\ell -1 ]},W^\ell_{0}) \neq {\rm null}
    \Big\}
=
    n^{-\omega(1)}.
\end{align*}
\end{lemma}
\begin{proof}
    Within the events $\Omega_0^\ell $ and
    $\nc(\{(U_s,u_s)\}_{s \in [\ell -1 ]},W^\ell_{0}) \neq {\rm null}$,
    $$
        (\inx^\ell, V_0^\ell, i_0^\ell)
    $$
    is a valid output of \nc$(\{(U_s,u_s)\}_{s \in [\ell- 1]}, W^\ell_0)$.

    Recall that
    $$
        {\cal E}^\ell_1 = \Eao\big( (V^\ell_\alpha, i^\ell_\alpha), W^\ell_1 \big)
    =
        \underbrace{{\cal E}_{\rm ao}(i_0^\ell)}_{\mbox{trivial event}} \cap \Eclu((V_0, i_0)) \cap \Enavi(V_0, W^\ell_1) \cap \Ecn(W^\ell_1) \cap \Enet(W^\ell_1).
    $$
    Since $(V_0^\ell, i_0^\ell) \in \{ (U_s, u_s) \}_{s \in [\ell-1]}$, we automatically have
    $$
        \Omega^\ell_{0} \cap \Big\{
        \nc(\{(U_s,u_s)\}_{s \in [\ell -1 ]},W^\ell_{0}) \neq \varnothing\Big\}
    \subseteq
        \Eclu(V_0^\ell, i_0^\ell) .
    $$
    Hence,
\begin{align*}
 &      \Prob \Big\{ ({\cal E}^{\ell+1}_{1})^c \,
        \Big\vert \, \Omega^\ell_{0} \mbox{ and
        \nc}(\{(U_s,u_s)\}_{s \in [\ell -1 ]},W^\ell_{0}) \neq \varnothing
    \Big\} \\
=&
    \Prob \Big\{  \, \big(\Enavi(V_0, W^\ell_1)\big)^c \cup
    \big(\Ecn(W^\ell_1)\big)^c \cap \big(\Enet(W^\ell_1)\big)^c
        \Big\vert \,
         \underbrace{   \Omega^\ell_{0} \mbox{ and
        \nc}(\{(U_s,u_s)\}_{s \in [\ell -1 ]},W^\ell_{0}) \neq \varnothing}_{\mbox{ event of } Y_0^\ell}
    \Big\} \\
\le &
  \Prob \Big\{  \, \big(\Enavi(V_0, W^\ell_1)\big)^c         \Big\vert \,
     \Omega^\ell_{0} \mbox{ and
        \nc}(\{(U_s,u_s)\}_{s \in [\ell -1 ]},W^\ell_{0}) \neq \varnothing
    \Big\}
+
    \Prob\{ (\Ecn(W^\ell_1))^c \}  + \Prob\{ (\Enet(W^\ell_1))^c \}\\
\le &
  \Prob \Big\{  \, \big(\Enavi(V_0, W^\ell_1)\big)^c         \Big\vert \,
     \Omega^\ell_{0} \mbox{ and
        \nc}(\{(U_s,u_s)\}_{s \in [\ell -1 ]},W^\ell_{0}) \neq \varnothing
    \Big\}
    + n^{-\omega(1)},
\end{align*}
where in the last inequality we applied Lemma \ref{lem: cn} and Lemma \ref{lem: epsilonNetEvent}.

Finally, the argument to show
$$\Prob \Big\{  \, \big(\Enavi(V_0, W^\ell_1)\big)^c         \Big\vert \,
     \Omega^\ell_{0} \mbox{ and
        \nc}(\{(U_s,u_s)\}_{s \in [\ell -1 ]},W^\ell_{0}) \neq \varnothing
    \Big\} = n^{-\omega(1)}$$
is precisely the same as shown in the proof of Lemma \ref{lem: eventlk}. We will omit the details here.
The lemma follows.
\end{proof}

\subsection{Proof of Theorem~\ref{thm: NetInformal}}

\begin{proof}
  Recall that for  $\ell \ge 2$,
    $$
        {\cal E}^{\ell}_{\rm halt} :=
 \Big\{\nc(\{(U_s,u_s)\}_{s \in [\ell -1 ]},W^\ell_{0})  = {\rm null}\Big\}.
    $$
   Sicne $ W^\ell_{0}$ does not exsits for $\ell > \lceil n^{\varsigma} \rceil$,  we have
   $$
        {\cal E}^{\ell}_{\rm halt} = \varnothing \quad \forall \ell > \lceil n^{\varsigma} \rceil.
   $$

   \smallskip
   \noindent
   \textbf{Claim.}
   $$
        \Omega^{\ell+1}_0 = \varnothing
    \quad
        \forall \ell  \ge n^{\frac{3}{4}\varsigma}.
   $$
  (The index shift by \(+1\) is just for convenience later.)

   Let us prove this claim: Suppose, for the sake of contradiction, that \(\ell\ge n^{\frac{3}{4}\varsigma}\) and \(\Omega_0^{\ell+1}\neq\varnothing\).  Then \(\Omega_0^\ell\) is also nonempty and
   \[
    \Omega_0^\ell
    \;\subseteq\;
    \mathcal{E}_0^\ell
    \;\subseteq\;
    \mathcal{E}_{\mathrm{rps}}(u_0, u_1, \ldots, u_\ell).
  \]
  Applying a standard volumetric argument, we obtain
    $$
        1
    =
        \mu(M)
    \ge
        \mu\Big(\bigcup_{s \in [\ell]} B(X_{u_s}, 0.15 \delta)\Big)
    \ge
        \sum_{s \in [\ell]} \mu(B(X_{u_s}, 0.15 \delta))
    \ge
        \mu_{\min}(0.15 \delta) \cdot \ell
    \ge
        c_{\mu} (0.15\delta)^d \ell.
    $$
    where \(\mu_{\min}(r)\) is a lower bound on the measure of any ball of radius \(r\) in \(M\).  Since \(\delta \gtrsim n^{-\varsigma/(2d)}\), there is a constant \(C\) (depending polynomially on the relevant parameters) such that
    \[
      \ell \;\le\; C\,n^{\varsigma/2}.
    \]
    This contradicts \(\ell \ge n^{\frac{3}{4}\varsigma}\) for sufficiently large \(n\). Hence \(\Omega_0^{\ell+1}=\varnothing\), proving the claim.

    \smallskip
    \noindent
    \underline{Conclusion of main argument.}
   By Proposition \ref{prop: net-check2} and the above claim,
   \begin{align*}
        \Omega_{\rm netFound}
    =
     \bigcup_{\ell \ge 2}
     (\Omega_0^\ell \cap {\cal E}_{\rm halt}^\ell)
    =
        \bigcup_{\ell \in [2,n^{3\varsigma/4}]}
        (\Omega_0^\ell \cap {\cal E}_{\rm halt}^\ell).
    \end{align*}

    Observe the union above is a disjoint union. Indeed, for $\ell_1 \neq \ell_2$,
    $$
        \Omega_0^{\ell+1} \subseteq \Omega_0^\ell \cap \big( {\cal E}_{\rm halt}^\ell\big)^c.
    $$
    Hence,
    \begin{align*}
       \Prob\{ \Omega_{\rm netFound} \}
    = &
        \sum_{\ell \in [2,n^{3\varsigma/4}]} \Prob\big\{
            \Omega^\ell_0
            \cap {\cal E}_{\rm halt}^\ell
        \big\}.
    \end{align*}

    \smallskip
    \noindent
    Next, by Lemmas~\ref{lem: eventlk}, \ref{lem: eventl0}, and \ref{lem: eventld}, for each \(\ell\ge 2\),
    \begin{align*}
         \Prob \Big\{ (\Omega^{\ell+1}_{0})^c
            \,\Big\vert\,  \Omega^\ell_0 \cap ( {\cal E}_{\rm halt}^\ell\big)^c
        \Big\}
    &=
        n^{-\omega(1)}
    \end{align*}
    which implies
    \begin{align*}
        \Prob\big\{
             \Omega^\ell_0
        \cap \big({\cal E}_{\rm halt}^\ell\big)^c
               \big\}
&\le
        \Prob \big\{ \Omega^{\ell+1}_{0} \big\} + n^{-\omega(1)},
        \end{align*}
        since
     $\Omega^{\ell+1}_{0} \subseteq  \Omega^\ell_0 \cap ({\cal E}_{\rm halt}^\ell)^c$.

     \smallskip
     \noindent
    We now apply this recurrence inductively. For instance,
     \begin{align}
    \label{eq: main00}
        \Prob\big\{
                 \Omega^\ell_0
            \cap \big({\cal E}_{\rm halt}^\ell\big)^c
                   \big\}
    \le &
        \Prob \big\{ \Omega^{\ell+1}_{0} \big\} + n^{-\omega(1)} \\
    \nonumber
    \le &
        \Prob \big\{ \Omega^{\ell+1}_{0} \cap  ({\cal E}_{\rm halt}^{\ell +1})^c \big\}
        +   \Prob \big\{ \Omega^{\ell+1}_{0} \cap  {\cal E}_{\rm halt}^{\ell +1} \big\}
        +  n^{-\omega(1)}.
   \end{align}
   Now we apply the above inequality recursively, we have
   \begin{align}
    \nonumber
         1
    =&
        \Prob\{ \Omega_{0}^2 \} + \Prob\{ (\Omega_{0}^1)^c\} \\
    \nonumber
    = &
        \Prob\{ \Omega_{0}^2 \cap {\cal E}_{\rm halt}^2\}
        +
        \Prob\{ \Omega_{0}^2 \cap {\cal E}_{\rm halt}^c\} + n^{-\omega(1)}  \\
      \le &
        \Prob\{ \Omega_{0}^2 \cap {\cal E}_{\rm halt}^2\}
    +
       \Prob\{ \Omega_0^3 \} + 2n^{-\omega(1)}  \\
    \le &
    \nonumber
        \sum_{ \ell \in [2,s]}
        \Prob\{ \Omega_{0}^s \cap {\cal E}_{\rm halt}^s\}
        + \Prob\{ \Omega_0^{\ell+1}\} + n^{3\varsigma/4} n^{-\omega(1)}
    \le
        \Prob\{\Omega_{\rm netFound}\} + 0 + n^{-\omega(1)}.
   \end{align}
   Rearranging the terms, the proof is complete.
\end{proof}

\section{From cluster-net to distance approximation}\label{sec: net-to-GH}

The goal of this section is to prove Theorems~\ref{thm: geodInformal} and \ref{thm: geodGHInformal}, which are based on the cluster-net $\{(U_s,u_s)\}_{s \in [\ell]}$ obtained from Algorithm \bnet. This section is organized as follows:
\begin{enumerate}
    \item First, we show that, with high probability, the distance between $u_s$ and any $v \in {\bf V}$ can be estimated from the number of points in the cluster $U_s$ that share an edge with $v$.
    \item Second, we construct a weighted graph $\Gamma = \Gamma(G)$, as stated in Theorem~\ref{thm: geodInformal}, along with the associated metric $\de$ also appearing in Theorem~\ref{thm: geodInformal}.
    \item Finally, we introduce the measure $\nu$ described in Theorem~\ref{thm: geodGHInformal} and complete the proof of the theorem.
\end{enumerate}

\subsection{Distance estimate via edge counting}
From Theorem~\ref{thm: net}, we know that with high probability, \bnet~ returns a $(\delta,\eta)$-cluster-net $\{(U_s,u_s)\}_{s \in [\ell]}$ of $G_{\bf V}$. Intuitively, for each $v \in {\bf V}$, the number of edges connecting $v$ to $U_s$ should reveal information about the distance $\|X_v - X_{u_s}\|$.
A complication arises because these same edges have influenced on the construction of $U_s$, introducing dependency issues. To circumvent this, we enlarge the vertex set by a factor of two:
\begin{align*}
    |{\bf V}| = 2n \cdot (d+2) \cdot \lceil n^{\varsigma} \rceil,
\end{align*}
and partition it into two disjoint subsets of equal size:
\begin{align*}
    {\bf V} = {\bf V}_1 \sqcup {\bf V}_2.
\end{align*}
This modification does not affect the statements of the theorems. We then apply Algorithm~\bnet~ to $G_{{\bf V}_1}$ instead of $G_{\bf V}$. Define
\begin{align}
    \label{eq: netFound}
    \Omega_{\rm netFound}
\end{align}
to be the event that the output of \bnet, $\{(U_s,u_s)\}_{s \in [\ell]}$, forms a $(\delta,\eta)$-cluster-net of $G_{{\bf V}_1}$ with $\ell \le n^{3\varsigma/4}$. Equivalently, $\Omega_{\rm netFound}$ is the same event as in Theorem~\ref{thm: net}, and occurs with probability at least $1 - n^{-\omega(1)}$.

Next, we partition ${\bf V}_2$ into $(d+2)\cdot \lceil n^{\varsigma}\rceil$ subsets of size $n$, indexed from $0$ to $(d+2)\cdot \lceil n^{\varsigma}\rceil - 1$:
\[
    {\bf V}_2 = \bigcup_{s \in [0,\, (d+2)\cdot \lceil n^{\varsigma}\rceil - 1]} V_s.
\]

For each $s \in [\ell]$, our goal is to extract a subset $U_s' \subseteq V_s$ such that $(U_s', u_s)$ forms a $3\eta$-cluster (see Definition~\ref{def:cluster}). This is possible if the event $\Enavi(U_s, V_s)$ holds. Let \begin{align}
    \label{eq: Us'}
    U'_{s} = \left\{ v \in V_s\, \middle\vert \, \frac{\left|N(v)\cap U_s\right|}{\left|U_s\right|} \ge \rp(1.5\eta) \right\}\,.
\end{align}
and define a good event
\begin{align}
    \label{eq: netFound'}
    \Omega_1
    =
    \Omega_{\rm netFound} \cap
    \left( \bigcap_{s \in [\ell]}
    \Enavi(U_s, V_s) \cap \Enet(V_s) \right).
\end{align}
Under the event $\Omega_1$, we have

\begin{lemma}
    \label{lem: clusterCopy}
    Given the occurrence of $\Omega_1$, $(U'_s,u_s)$ is a $3\eta$-cluster
    for each $s \in [\ell]$.
\end{lemma}

Let us postpone the proof of this lemma to the end of this subsection and focus on the discussion. In short, the key idea is to replace each \((U_s, u_s)\) with \((U_s', u_s)\), where \(U_s'\) is defined in~\eqref{eq: Us'}, and the cluster radius is increased from \(\eta\) to \(3\eta\). Crucially, the edges between \(U_s'\) and \({\bf V} \setminus U_s\) remain unrevealed. This separation allows us to estimate the distance from \(u_s\) to any other vertex by counting edges from \(U_s'\), thereby avoiding the dependency issues discussed earlier.
\noindent
To continue our discussion, we introduce additional notation. Let
\[
    \mathbf{V}_{\mathrm{rest}}
    \;:=\;
    \mathbf{V}
    \;\setminus\;
    \Bigl(
        \bigcup_{s \in [\ell]} \bigl(U_s \;\cup\; U_{s'}\bigr)
    \Bigr).
\]
That is, \(\mathbf{V}_{\mathrm{rest}}\) is the set of all vertices in \(\mathbf{V}\) that do not belong to any \(U_s\) or \(U_{s'}\), for \(s \in [\ell]\).

It is important to note that applying Algorithm~\bnet~to \(G_{\mathbf{V}_1}\) depends only on \(\bigl(X_{\mathbf{V}_1}, \eU_{\mathbf{V}_1}\bigr)\), where \(\eU_{\mathbf{V}_1} = \{\eU_{i,j}\}_{i,j \in \mathbf{V}_1}\) are the uniform random variables used to generate edges according to the connection probabilities (see Definition~\ref{def: graphModel}). Moreover, for each \(s \in [\ell]\), constructing the set \(U_s'\) requires access only to \(\eU_{U_s, V_s}\) and \(X_{V_s}\).
In other words, the event \(\Omega_1\) is measurable with respect to the random variable
\[
    {\rm Z} := \Big( X_{\mathbf{V}_1}, \eU_{\mathbf{V}_1},\, (\eU_{U_s, V_s},\, X_{V_s})_{s \in [\ell]} \Big).
\]
In particular, the variables \(\eU_{U_s',\, \mathbf{V}_{\mathrm{rest}}}\) are not used in the construction of \(\Omega_1\) and remain conditionally independent of it.

Our next step is to estimate the distance from each \(X_v\) to the cluster center \(X_{u_s}\), which relies on part of the remaining unrevealed edge variables \(\eU_{U_s,V_s}\) and \(X_{V_s}\).
The main objective of this subsection is captured in the following proposition.

\begin{proposition}[Distance Approximation]
    \label{prop: GHdist}
    Define
    \[
      \Omega_{\dist}
      \;:=\;
      \Omega_{1}
      \;\cap\;
      \bigcap_{s \in [\ell]} \Enavi\bigl(U_{s}',\,\mathbf{V}_{\mathrm{rest}}\bigr).
    \]
    On this event, the following holds for every $s \in [\ell]$ and every vertex $v \in \mathbf{V}$:

    \begin{enumerate}
    \item For $v \notin U_{s} \cup U_{s}'$,

    \begin{align}
        \label{eq: GHdist00}
      \Big|\,
        \rp^{-1}\!\Bigl(\tfrac{|N(v)\,\cap\,U_{s}'|}{|U_{s}'|}\Bigr)
        \;-\;
        \|X_{v}\,-\,X_{u_{s}}\|
      \Big|
      \;\le\;
      0.01\,\delta.
    \end{align}

    \item For  $v \in U_{s} \cup U_{s}'$,

    \[
      \|X_{v}\;-\;X_{u_{s}}\|
      \;\le\;
      3\,\eta
      \;\le\;
      0.01\,\delta.
    \]
    \end{enumerate}

    Moreover, conditioning on any realization ${\rm z} \in \Omega_{1}$, the probability of $\Omega_{\dist}^c$ is
    \[
      \mathbb{P}
      \bigl[\,
        \Omega_{\dist}^c
        \,\bigm|\;
        {\rm Z} = {\rm z}
      \bigr]
      \;=\;
      n^{-\omega(1)},
    \]
    and therefore
    \[
      \mathbb{P}
      \bigl[\,
        \Omega_{\dist}^c
        \,\bigm|\;
        \Omega_{1}
      \bigr]
      \;=\;
      n^{-\omega(1)}.
    \]
    \end{proposition}

\noindent
Intuitively, the event \(\Omega_{\dist}\) guarantees that for each center \(u_s\), we can accurately approximate the distance \(\|X_{u_s} - X_v\|\) for any vertex \(v\), up to an additive error of \(0.01\,\delta\), by simply examining the proportion of edges from \(v\) to the cluster \(U_s'\). If \(v\) belongs to \(U_s \cup U_s'\), then by the definition of a cluster, it is even closer to its center \(u_s\), with distance at most \(3\,\eta \le 0.01\,\delta\).

To prove the Proposition \ref{prop: GHdist} and Lemma \ref{lem: clusterCopy},
we introduce the following notation, for each $s \in [\ell]$ we define
\[
    \rp_{s}(x)
:=
    \frac{\sum_{u \in U_s} \rp(\|X_u - x\|)}{|U_s|}\,.
\]

\begin{proof}[Proof of Lemma~\ref{lem: clusterCopy}]
    \noindent
    Under the event $\Enavi(U_s,V_s)$, for all $v \in V_s$ we have
    \begin{align}
        \label{eq: clusterCopy00}
        \bigg|
            \frac{|N(v)\cap U_s|}{|U_s|}
            \;-\;
            \rp_s(X_v)
        \bigg|
        \;\;\le\;\;
        n^{-\frac{1}{2}+\varsigma}.
    \end{align}
    Since $\Omega_{\rm netFound} \subseteq \Eclu(U_s,u_s)$, for $v \in V_s$ it follows that
    \[
        \rp_s(X_v)
        \;<\;
        \rp\bigl(\|X_v - X_{u_s}\| - \eta\bigr).
    \]
    For any $v$ satisfying $\|X_v - X_{u_s}\|\ge 3\eta$, we further have $\rp_s(X_v) < \rp(2\eta)$. Consequently,
    \[
        \frac{|N(v)\cap U_s|}{|U_s|}
        \;\;\le\;\;
            \rp_s(X_v) \;+\; n^{-\frac{1}{2} + \varsigma}
        \;<\;
            \rp(2\eta) \;+\;  n^{-\frac{1}{2} + \varsigma}
        \;\;\le\;\;
            \rp\bigl(2\eta \;-\; L_{\rp}\,n^{-\frac{1}{2}+\varsigma}\bigr)
        \;\;\overset{\eqref{eq: CondEta},\,\eqref{eq: CondVarepsilon}}{\le}\;\;
            \rp(1.5\eta).
    \]
    By definition, this implies $v \notin U_s'$. Therefore, we conclude that if $v \in U_s'$, then
    \[
        \|X_v - X_{u_s}\|
        \;<\;
        3\eta.
    \]

    \medskip

    \noindent
It remains to show that \( |U_s'| \ge n^{1-\varsigma} \). First, by the event
\( \Enet(V_s) \), we have
\begin{align}
    \label{eq: Us'00}
    \bigl|\bigl\{ v \in V_s : \|X_v - X_{u_s}\|\le \varepsilon \bigr\}\bigr|
    \;\;\ge\;\; n^{\,1 - \varsigma}.
\end{align}
Next, for each \(v \in B(X_{u_s}, \varepsilon) \cap V_s\),
\[
    \frac{|N(v)\,\cap\,U_s|}{|U_s|}
    \;
   \stackrel{\eqref{eq: clusterCopy00}}{\ge}
    \;
    \rp_s(X_v) \;-\; n^{-\tfrac{1}{2} + \varsigma}
    \;\;>\;\;
    \rp(\varepsilon + \eta) \;-\; n^{-\tfrac{1}{2} + \varsigma}
    \;\;\ge\;\;
    \rp\bigl(\varepsilon + \eta + L_{\rp}\,n^{-\tfrac{1}{2}+\varsigma}\bigr)
    \;\;\overset{\eqref{eq: CondEta}}{\ge}\;\;
    \rp(1.5\eta),\,
\]
which in term implies $v \in U_s$. In other words,
\[
    \bigl\{ v \in V_s : \|X_v - X_{u_s}\|\le \varepsilon \bigr\}
    \;\;\subseteq\;\;
    U_s'.
\]
Combining this inclusion with \eqref{eq: Us'00}, we conclude that
\[
    |U_s'| \;\;\ge\;\; n^{\,1 - \varsigma}.
\]

\end{proof}

\begin{proof}[Proof of Proposition~\ref{prop: GHdist}]

    \noindent
\underline{Probability Estimate:}
For any realization of $y \in \Omega_1$, we have
\begin{align*}
    \Prob \bigl\{
        \Omega_{\rm dist}^c
        \,\bigm\vert\,
        {\rm Z} = {\rm z}
    \bigr\}
    \;\le\;
    \sum_{s \in [\ell]} \Prob \Bigl\{
     \bigl(\Enavi\bigl( U_{s'}, {\bf V}_{\rm rest}\bigr)\bigr)^c \,\Bigm\vert\,
     {\rm Z} = {\rm z} \Bigr\}.
\end{align*}
Since $\Omega_1 \subseteq \Omega_{\rm netFound}$, we know from the definition of $\Omega_{\rm netFound}$ that $\ell \le n^{3\varsigma/4}$.
Hence, the probability estimate in the proposition follows if
\begin{align}
\label{eq: OmegaDist00}
\forall s \in [\ell],\quad
    \Prob \Bigl\{
    \bigl(\Enavi\bigl( U_{s'}, {\bf V}_{\rm rest}\bigr)\bigr)^c \,\Bigm\vert\,
    {\rm Z} = {\rm z} \Bigr\}
    \;=\;
    n^{-\omega(1)}.
\end{align}
Since the conditioning may not seem straightforward, we now describe it carefully.

First, fix a realization
\[
    \bigl(X_{{\bf V}_1}, \eU_{{\bf V}_1}\bigr)
    =
    \bigl(x_{{\bf V}_1}, \mathfrak{u}_{{\bf V}_1}\bigr) \;\in\; \Omega_{\rm netFound}.
\]
Then the set \(\{(U_s,u_s)\}_{s \in [\ell]}\) is determined as a function of \(\bigl(x_{{\bf V}_1}, \mathfrak{u}_{{\bf V}_1}\bigr)\).
Next, fix \(s \in [\ell]\) and a realization
\[
    \bigl(\eU_{U_s,V_s},\, X_{V_s}\bigr)
    \;=\;
    \bigl(\mathfrak{u}_{U_s,V_s}, x_{V_s}\bigr)
    \;\in\; \Enavi(U_s,V_s) \,\cap\, \Enet(V_s).
\]
Similarly, the set \(U_s'\) is then determined as a function of
\(\bigl(x_{{\bf V}_1}, \mathfrak{u}_{{\bf V}_1}, \mathfrak{u}_{U_s,V_s}, x_{V_s}\bigr)\).
Now, fix a realization of \({\rm Z} = {\rm z}\), as in the statement of the proposition.

Observe that the random variables \(\eU_{U_s', {\bf V}\setminus U_s}\) have not yet been revealed.
Applying Lemma~\ref{lem: Enavi}, we obtain
\begin{align*}
    \Prob_{\eU_{U_s,{\bf V}\setminus U_s}} \Bigl\{
        \bigl(\Enavi\bigl(U_{s'},{\bf V}\setminus U_s\bigr)\bigr)^c
        \,\Bigm\vert\,
        {\rm Z} = {\rm z}
    \Bigr\}
    \;=\;
    n^{-\omega(1)},
\end{align*}
which, by Fubini's theorem, implies
\begin{align*}
    \Prob \Bigl\{
        \bigl(\Enavi\bigl(U_{s'},{\bf V}\setminus U_s\bigr)\bigr)^c
        \,\Bigm\vert\,
        \Omega_1
    \Bigr\}
    \;=\;
    n^{-\omega(1)}.
\end{align*}
Since \(\ell \le n^{3\varsigma/4}\) by Theorem~\ref{thm: net}, this establishes \eqref{eq: OmegaDist00}.

    \medskip

\noindent
\underline{Distance Estimate:}
For the second statement, first note that \((U_s',u_s)\) is a \(3\eta\)-cluster by Lemma~\ref{lem: clusterCopy}. Thus, for any \(v \in {\bf V}\setminus U_s\), the event $\Enavi(U_s', {\bf V} \setminus _)$ implies that
\begin{align*}
    \rp\bigl(\|X_v - X_{u_s}\| - 3\eta\bigr)
    &\;\le\; \rp_s(X_v)
    \;
    \stackrel{\eqref{eq: E-navi}}{\le}
    \; \frac{|N(v) \cap U'_s|}{|U'_s|} \;+\; n^{-1/2+\varsigma} \\
    \Rightarrow\quad
    \rp \Bigl(\|X_v - X_{u_s}\| - 3\eta \;-\; \ell_\rp^{-1} n^{-1/2+\varsigma}\Bigr)
    &\;\le
        \; \frac{|N(v) \cap U'_s|}{|U'_s|}.
\end{align*}
Since \(\rp\) is monotone decreasing, applying \(\rp^{-1}\) to both sides yields
\[
    \rp^{-1}\biggl(\frac{|N(v) \cap U'_s|}{|U'_s|}\biggr)
    \;\ge\;
    \|X_v - X_{u_s}\| - 3\eta \;-\; \ell_\rp^{-1} n^{-1/2+\varsigma}
    \;\ge\;
    \|X_v - X_{u_s}\| -
    0.01\delta.
\]
A corresponding upper bound follows by a similar argument; we omit those details here.

\end{proof}

\subsection{Construction of the weighted graph $\Gamma$ and the metric $\de$}
Let us first layout the algorithm to construct the weighted graph described in Theorem \ref{thm: geodInformal}:\\

\newpage
\begin{algorithm}[H]
\caption{Weighted Graph $\Gamma = \Gamma(G,\rr)$ Construction}
\label{alg:weightedGraph}
\SetKwInOut{Input}{Input}
\SetKwInOut{Output}{Output}
\SetKwComment{tcc}{\hspace{0.25em}// }{}

\Input{
  Graph $G=(\mathbf{V},E)$ with vertex set $\mathbf{V}$ of size $2n\,(d+2)\,\lceil n^{\varsigma}\rceil$;\\
  parameter $\rr>0$
}
\Output{ $(\Gamma\,, {\rm d}_\Gamma)$}

\BlankLine
$m \;\leftarrow\; (d+2)\,\lceil n^{\varsigma}\rceil$\;
Partition $\mathbf{V}$ into two equal parts
$\mathbf{V}_1$ and $\mathbf{V}_2$, each of size $n\,m$\;
Partition $\mathbf{V}_2$ into $m$ batches of size $n$: $V_1,\dots,V_m$\;

\smallskip

$\{(U_s,u_s)\}_{s\in[\ell]} \;\leftarrow\; \bnet\bigl(G_{\mathbf{V}_1}\bigr)$
\tcc*[r]{Run \bnet\ on the first half}

\smallskip
\tcc{Construct $U_s'$}
\For{$s \leftarrow 1$ \KwTo $\ell$}{
  $U_s' \leftarrow \emptyset$\;
  \ForEach{$v \in V_s$}{
    \If{$|N(v)\cap U_s|\,/\, |U_s|
        \;\ge\; \rp(1.5\,\eta)$}{
        $U_s' \leftarrow U_s' \cup \{v\}$\;
    }
  }
}

\BlankLine
Initialize $\Gamma$ as an edgeless graph on vertex set $\mathbf{V}$\;

\smallskip
\tcc{ Step 1: Edges between cluster centers }
\For{$1\le s < t \le \ell$ }{
      $d_{st}\ \leftarrow\ \rp^{-1}\!\bigl(|N_G(u_s)\cap U'_t|\,/\,|U'_t|\bigr)$\;
      \If{$d_{st}\le \rr$}{
          ${\rm w}(u_s,u_t)\ \leftarrow\ d_{st}+0.04\,\delta$\;
          add edge $\{u_s,u_t\}$ with weight ${\rm w}(u_s,u_t)$ to $\Gamma$\;
  }
}

\smallskip
\tcc{Step 2: Edges from all other vertices }
\For(\tcc*[f]{for each non-center $v$, find its closest center $u_{s_v}$}){$v\in\mathbf{V}\setminus\{u_s\}_{s\in[\ell]}$}{
   \If{$v\in\bigcup_{s\in[\ell]}(U_s\cup U_s')$}{
        $s_v \gets$  unique $s$ such that $v\in U_{s}\cup U'_{s}$\;
   }
   \Else{
        $d_{\min}\leftarrow\infty,\ s_v\leftarrow \mathrm{null}$ \tcc*[r]{find the unique $u_s$ which minimizes the "distance"}
        \For{$s\gets 1$ \KwTo $\ell$}{
             $d_s\leftarrow \rp^{-1}\!\bigl(|N_G(v)\cap U'_s|\,/\,|U'_s|\bigr)$\;
             \If{$d_s<d_{\min}$}{
                 $d_{\min}\leftarrow d_s,\ s_v\leftarrow s$\;
             }
        }
   }
   ${\rm w}(v,u_{s_v})\ \leftarrow\ \delta$
   add edge $\{v,u_{s_v}\}$ with weight ${\rm w}(v,u_{s_v})$ to $\Gamma$\;
}
    ${\bf d}_{\Gamma} \gets$ shortest-path algorithm (i.e. Dijkstra's algorithm applied to every vertex) applied on $\Gamma$\;
   \Return{$(\Gamma, {\rm d}_\Gamma)$}
\end{algorithm}
First of all, to construct the graph we need to choose a suitable parameter $\rr>0$. Let us postpone the discussion of the choice of $\rr$ for now and proceed to the construction of the graph $\Gamma = \Gamma(G,\rr)$.
The graph includes edges between the cluster centers \(\{u_s\}_{s \in [\ell]}\), where the presence and weight of an edge between \(u_s\) and \(u_t\) depend on their estimated distance. Specifically, each such edge is assigned a weight approximating \(\|X_{u_s} - X_{u_t}\| + 0.04\delta\). Here we add an extra $0.04\delta$ to discourage overly long paths from being considered shortest-path distances. This modification prevents small gaps from accumulating when comparing the shortest path distance to the underlying geodesic distance.

In addition, for every non-cluster-center vertex \(v \in \mathbf{V} \setminus \{u_s\}_{s \in [\ell]}\), we find a unique index $s_v \in [\ell]$ in which we treat $u_{s_v}$ is the nearest cluster center to \(v\), and $v$
 only has one edge connecting to its nearest cluster center \(u_{s_v}\), determined using the distance approximation from Proposition~\ref{prop: GHdist}. The weight of this edge is set to \(\delta\).
Consequently, the total number of edges in the graph is at most
\begin{align}
    \label{eq: weightGraph_edgeCount}
    \ell^2 + |\mathbf{V}|  \le 2|\mathbf{V}|\,.
\end{align}

\begin{remark}[Running Time of Algorithm \wg]
Given a successful output $\{(U_s,u_s)\}_{s \in [\ell]} = \bnet(G_{{\bf V}_1})$, the total running time of Algorithm~\wg\ is bounded by
\begin{align}
    \label{eq: pathdistance_runningtime}
    O(n^{3(1+\varsigma)}) = O(|{\bf V}|^3),
\end{align}
where the dominant cost arises from executing \bnet\ on $G_{{\bf V}_1}$.

To explain this bound, let us briefly analyze the time complexity of Algorithm~\wg\ **after** \bnet\ has been executed. Constructing the graph $\Gamma(G,\rr)$ involves two primary tasks:
1. Extracting $U_s'$ for each $s \in [\ell]$, which takes $O(n \cdot \ell)$ time; and 2. Computing $|N(v) \cap U'_s|$ for each $v \in \mathbf{V}$ and $s \in [\ell]$ and find the maximium among them, which takes $O(\ell n^2)$ time overall. Therefore, the overall time to construct $\Gamma(G,\rr)$ is
\[
    O(\ell n^2) = O(n^{2 + 3\varsigma/4}),
\]
To compute all-pairs shortest path distances $\sd_{\Gamma(G,\rr)}(v,w)$ for $v,w \in \mathbf{V}$, we apply Dijkstra’s algorithm from each vertex. Since $\Gamma(G,\rr)$ has at most
\begin{align}
    \label{eq: weightGraph_edgeCount}
    |E| \le 2|\mathbf{V}| = O(n^{1+\varsigma})
\end{align}
edges, the total running time for this step is
\[
    O\big(|\mathbf{V}| \cdot (|\mathbf{V}| + |E|)\log |E|\big)
    \le O(n^{2(1+\varsigma)} \log n).
\]

In summary, the construction and shortest path computation in \wg\ takes time at most $O(n^{2(1+\varsigma)}\log n)$, which is dominated by the $O(n^{3(1+\varsigma)})$ cost of \bnet.
\end{remark}

\noindent
We will only consider the graph $\Gamma$ for two specific choices of the parameter $\rr$:

\begin{itemize}
    \item \underline{Geodesic-distance approximation.}
    Let $\Gamma = \Gamma(G, 2\delta^{1/3})$, i.e.\ $\rr = 2\delta^{1/3}$.
    Intuitively, the induced subgraph of $\Gamma$ restricted to $\{u_s\}_{s\in[\ell]}$ is formed by connecting $u_i$ and $u_j$ whenever $\|X_{u_i}-X_{u_j}\| \lesssim \delta^{1/3}$.
    Then, for any $v \notin \{u_s : s \in [\ell]\}$, vertex $v$ is connected to exactly one neighbor $u_{s_v}$, and the corresponding edge weight is set to $\delta$.

    \item \underline{Euclidean-distance approximation.}
    Define $\de$ as the path-distance metric on $\Gamma = \Gamma(G,\infty)$, noting that $\rr = \infty$.
    Admittedly, the graph $\Gamma(G,\infty)$ provides less metric structure than the geodesic setting, since its induced subgraph on $\{u_s\}_{s\in[\ell]}$ is a complete graph. We adopt $\Gamma(G,\infty)$ for defining $\de$ so that we can reuse arguments developed for the geodesic-distance setting.
\end{itemize}

\begin{definition}
    Given the event $\Omega_{\rm dist}$, we define the weighted graph $$\Gamma = \Gamma(G,2\delta^{1/3})$$ as in Algorithm \ref{alg:weightedGraph}.
    This will be the graph stated in Theorem~\ref{thm: geodInformal}. Furthermore, let $\Gamma_{\rm euc} = \Gamma(G,\infty)$ and define the metric $\de$ on ${\bf V}$ as
    $$
        \de(v,w) = \sd_{\Gamma_{\rm euc}}(v,w),
    $$
    the path metric on $\Gamma_{\rm euc}$.
    To avoid ambiguity, let ${\rm w}(v,w)$ denote the weight of the edge $\{v,w\}$ in $\Gamma$ and ${\rm w}_{\rm euc}(v,w)$ denote the weight of the edge $\{v,w\}$ in $\Gamma_{\rm euc}$.
\end{definition}

Let us wrap this subsection with the following observation on $\Gamma$:
\begin{lemma}
    \label{lem: 4i2034i2342}
Condition on $\Omega_{\rm dist}$.  The following holds.
\begin{itemize}
    \item For $ \forall 1 \le i < j \le \ell$,
\begin{align}
\nonumber
       &&& \rp^{-1}\left(\frac{|N(u_j) \cap U'_i|}{|U'_i|}\right)
    \le
        2\delta^{1/3}\\
\label{eq: wToDist}
\Rightarrow &&&
        \|X_{u_i} - X_{u_j}\| + 0.03\delta
    \le
        {\rm w}(u_i,u_j)
    \le
        \|X_{u_i} - X_{u_j}\| + 0.05\delta.
\end{align}
\item
For each $v \notin \bigcup_{s \in [\ell]}U_s\cup U'_s$,
\begin{align}
\label{eq: vusvdist}
    \|X_v -X_{u_{s_v}}\| \le 1.02\delta,
\end{align}
\end{itemize}
\end{lemma}

\begin{proof}

    \noindent
    \underline{First Statement:} From the assumption $\rp^{-1}\left(\frac{|N(u_j) \cap U'_i|}{|U'_i|}\right) \le  2\delta^{1/3}$, we know $u_j$ and $u_i$ are connected by an edge with weight ${\rm w}(u_i,u_j) = \rp^{-1}\left(\frac{|N(u_j) \cap U'_i|}{|U'_i|}\right) + 0.04\delta$. Then, by \eqref{eq: GHdist00} from Proposition \ref{prop: GHdist}, the first statement follows.

    \noindent
    \underline{Second Statement:}
    For the second statement, recall that $\Omega_{\rm dist} \subseteq \Omega_{\rm netFound}$, thus $\{u_s\}_{s\in [\ell]}$ is a $\delta$-net. In particular, there exists $s'_v$ such that
    $\|X_v - X_{u_{s'_v}}\| \le \delta$.
    Thus, by Proposition \ref{prop: GHdist},
    \begin{align*}
        \|X_v -X_{u_{s_v}}\|
    \le &
        \rp^{-1} \left(\frac{|N(v) \cap U'_{s_v}|}{|U'_{s_v}|}\right) + 0.01 \delta  \\
    & \phantom{AAA} \le
        \rp^{-1} \left(\frac{|N(v) \cap U'_{s_v'}|}{|U'_{s_v'}|}\right) + 0.01\delta
    \le
        \|X_v - X_{u_{s'_v}}\| + 0.02\delta
    \le 1.02\delta.
    \end{align*}
\end{proof}
An immediate analogue in the case when $\rr = \infty$ is the following:
\begin{lemma}
    \label{lem: 4i2034i2342euc}
Condition on $\Omega_{\rm dist}$.  If $\rr = \infty$, then the following holds.
\begin{itemize}
    \item For $ \forall 1 \le i < j \le \ell$,
\begin{align}
\label{eq: wToDist_euc}
        \|X_{u_i} - X_{u_j}\| + 0.03\delta
    \le
        {\rm w}_{\rm euc}(u_i,u_j)
    \le
        \|X_{u_i} - X_{u_j}\| + 0.05\delta.
\end{align}
\item
For each $v \notin \bigcup_{s \in [\ell]}U_s\cup U'_s$,
\begin{align}
\label{eq: vusvdist_euc}
    \|X_v -X_{u_{s_v}}\| \le 1.02\delta.
\end{align}
\end{itemize}
\end{lemma}
We omit the proof here since it is an analogue to that of Lemma \ref{lem: 4i2034i2342}.

\subsection{Proof of Theorem~\ref{thm: geodInformal}}
Let us now restate Theorem~\ref{thm: geodInformal} in the context of $\Gamma$ and $\Omega_{\rm dist}$ introduced in the previous subsection.
\begin{theor}
\label{thm: distTechnical}
    Suppose the event \(\Omega_{\rm dist}\) occurs, which happens with probability at least \(1 - n^{-\omega(1)}\). Then for all \(v, w \in \mathbf{V}\), we have
    $$
        \left|\sd_{\Gamma}(v,w) - \sd_{\rm gd}(X_v,X_w)\right| \le C{\rm diam}_{\rm gd}(M) r_M \delta^{2/3},
    $$
    where $C \ge 1$ is an universal constant, and
    $$
        \left| \de(v,w) - \|X_v - X_w\| \right| \le 4\delta .
    $$
\end{theor}

Before we proceed to the proof of the theorem, we need a comparison between Euclidean distance and geodeisc Distance:
\begin{lemma} \label{lem: EucDistGeoDist}
    For points $p,q \in M$ with $\|p-q\|\le r_M$, we have
    $$
        \sd_{\rm gd}(p,q) \le   \|p-q\| (1+ \kappa \|p-q\|^2/2).
    $$
\end{lemma}
\begin{proof}
   Fix points $p,q$ described in the Lemma. Consider Proposition~\ref{prop: SecondFundamentalForm} with $H=T_pM$.

   By definition of $r_M$, the set $B(p,r_M) \cap M$ is connected.
   By $(e)$ of Proposition~\ref{prop: SecondFundamentalForm}, the orthogonal projection $P$ from $M$ to $T_pM$ is a bijection when restricted to $B(p,r_M) \cap M$. Denote its inverse by $\phi$. Then there exists $v \in T_pM \cap B(p,r_M)$ such that $\phi(v) = q$. By $(d)$ of Proposition~\ref{prop: SecondFundamentalForm} applied to $v$,
   we obtain
   $$
        \sd_{\rm gd}(p,q) \le \|p-q\| (1+ \kappa \|p-q\|^2/2)\,,
   $$
   completing the proof.

\end{proof}

\subsection{Proof for the Geodesic Distance Setting}
Let us begin the proof of Theorem~\ref{thm: distTechnical} in the geodesic setting. To do this, we split the comparison between $\sd_{\Gamma}$ and $\sd_{\rm gd}$ into two lemmas, each of which addresses one side of the inequality.
\begin{lemma} \label{lem: GammaGH1}
Condition on $\Omega_{\dist}$. With $\Gamma = \Gamma(G, 2\delta^{1/3})$,
for every pair of vertices $v,w \in {\bf V}$, we have
$$
    \sd_{\Gamma}(v,w)
\le
    \sd_{\rm gd}(X_v,X_w) ( 1 + 20\delta^{2/3})
\le
    \sd_{\rm gd}(X_v,X_w) +
    20{\rm diam}_{\rm gd}(M) \delta^{2/3}.
$$
\end{lemma}
\begin{proof}

Fix $v,w \in {\bf V}$. We will derive the bound depending whether or not $v,w \in \{u_s\}_{s \in [\ell]}$.

\noindent
\underline{Case: $\|X_v-X_w\| \le \delta^{1/3}$.}
Assume that both $v,w \notin \bigcup_{s \in [\ell]} U_s$. Then,
$$
    \|X_{u_{s_v}} - X_{u_{s_w}} \|
\le
    \|X_{v} - X_{u_{s_v}} \| + \|X_{v} - X_{w}\| +  \|X_{w}- X_{u_{s_w}}\|
\le
    1.02\delta + \delta^{1/3} + 1.02\delta
\le
   1.1 \delta^{1/3}\,,
$$
where the second to last inequality follows from Lemma \ref{lem: 4i2034i2342}.
Particularly, according to the event $\Omega_{\rm dist}$, we have
$$
    \max
    \left\{
        \rp^{-1}\left(\frac{|N(u_{s_v}) \cap U'_{s_w}|}{|U'_{s_w}|}\right),\,
        \rp^{-1}\left(\frac{|N(u_{s_w}) \cap U'_{s_v}|}{|U'_{s_v}|}\right)
    \right\}
\le
    \|X_{u_{s_v}} - X_{u_{s_w}}\| + 0.01 \delta  \le 1.2\delta^{1/3}.
$$
Hence, by \eqref{eq: wToDist}, the pair $\{u_{s_v}, u_{s_w}\}$ is an edge with weight
$$
    {\rm w}(u_{s_v}, u_{s_w}) \le \|X_{u_{s_v}} - X_{u_{s_w}}\| + 0.05\delta.
$$
Combining these, we obtain
\begin{align*}
    \sd_{\Gamma}(v,w)
\le &
    {\rm w}(v,u_{s_v})
+
    {\rm w}(u_{s_v},u_{s_w})
+
    {\rm w}(u_{s_w},w)\\
\le &
    \delta + \|X_{u_{s_v}} - X_{u_{s_w}}\| +0.05\delta + \delta\\
\le &
    3\delta +
     \|X_{v} - X_{u_{s_v}} \| + \|X_{v} - X_{w}\| +  \|X_{w}- X_{u_{s_w}}\|\\
\overset{\eqref{eq: vusvdist}}{\le} &
    6\delta + \|X_v-X_w\| \\
\le &
    \sd_{\rm gd}(X_v,X_w) + 6\delta.
\end{align*}
Therefore, $$
    \sd_{\rm \Gamma}(v,w) \le \sd_{\rm gd}(X_v,X_w) + 6\delta.
$$
When $v$ or $w$ is in $\bigcup_{s \in [\ell]} U_s$,
the argument is analogous, and we omit the details here.

\noindent
\underline{Case: $\|X_v-X_w\| \ge \delta^{1/3}$.}
First, observe that $$ \sd_{\rm gd}(X_v,X_w) \ge \delta^{1/3}.$$
Let $\gamma$  be a geodesic from $X_v$ to $X_w$ in $M$.
Define
$$t \;:=\; \left\lceil \frac{\sd_{\mathrm{gd}}(X_v, X_w)}{0.5\,\delta^{1/3}} \right\rceil.$$
Along this geodesic, choose a sequence of points in $M$
$$( X_v = q_0, q_1, q_2,\dots, q_{t}=X_w)$$
such that for $i \in [t-1]$,
$$
    \sd_{\rm gd}(q_i,q_{i-1}) = \frac{1}{2}\delta^{1/3}
$$
and
$\sd_{\rm gd}(q_{t-1},q_{t}) \le \frac{1}{2}\delta^{1/3}$.

Since $\Omega_{\rm netFound}\supseteq \Omega_{\rm dist}$, we know that the points $\{u_s\}_{s\in [\ell]}$ form a $\delta$-net in $M$. Hence, there exists a sequence of vertices
$$(v=v_0, v_1,v_2,\dots, v_{ t}=w)$$
with $\{ v_i \}_{i \in [ t-1]}$ is a multiset of  $\{u_s\}_{s\in [\ell]}$ such that for each $i \in [ t -1]$,
$$
    \|X_{v_i} - q_{i}\| \le \delta.
$$
In particular, for each $i \in [ t -1]$,
$$
    \|X_{v_{i-1}} - X_{v_{i}}\| \le \frac{1}{2}\delta^{1/3} + 2\delta \le \delta^{1/3}\,.
$$
By the argument from the previous case (where the Euclidean distance is at msot $\delta^{1/3}$), it follows that
$$
    \sd_{\rm \Gamma}(v_{i-1},v_i) \le \sd_{\rm gd}(X_{v_{i-1}},X_{v_i}) + 6\delta.
$$

We now sum these distances to bound \(\sd_{\Gamma}(v, w)\):
\begin{align*}
    \sd_{\rm \Gamma}(v,w)
\le &
    \sum_{i \in [t]}
    \sd_{\rm \Gamma}(v_{i-1},v_i)\\
\le &
    \sum_{i \in [t]}\left(
    \sd_{\rm gd}(X_{v_{i-1}},X_{v_i}) + 6\delta \right) \\
\le &
    \sum_{i \in [t]}\left(
    \sd_{\rm gd}(q_{i-1},q_{i})
    + \sd_{\rm gd}(q_{i-1},X_{v_{i-1}})
    + \sd_{\rm gd}(q_{i},X_{v_i})
    + 6\delta \right).
\end{align*}
Applying Lemma~\ref{lem: EucDistGeoDist}
(which bounds geodesic distance by Euclidean dsitance, plus a small correction) gives
\begin{align*}
    (*)
\le &
    \sum_{i \in [t]}\left(
    \sd_{\rm gd}(q_{i-1},q_{i})
    +   \delta ( 1+ \kappa \delta^2/2)
    +   \delta ( 1+ \kappa \delta^2/2)
    + 6\delta \right)\,.
\end{align*}
Because $t$ is of order $\delta^{-1/3}$, it implies
\begin{align*}
(*)
\le &
    \sd_{\rm gd}(X_v,X_w) + 2\frac{\sd_{\rm gd}(X_v,X_w)}{\delta^{1/3}} (6 \delta + 2\delta ( 1+ \kappa \delta^2/2)) \\
\le &
    \sd_{\rm gd}(X_v,X_w) (1+ 20 \delta^{2/3}),
\end{align*}
where the last inequality follows from $\kappa \ge r_M$ and $\delta^2 \le r_M$.
\end{proof}

\begin{lemma}
    \label{lem: GammaGH2}
Suppose $n$ is large enough so that
$\delta^{1/3} < 0.1 r_M$.
Condition on $\Omega_{\dist}$.
With $\Gamma = \Gamma(G, 2\delta^{1/3})$,
for every pair of vertices $v,w \in {\bf V}$, we have
$$
    \sd_{\Gamma}(v,w)
\ge
    \sd_{\rm gd}(X_v,X_w)
-
    C {\rm diam}_{\rm gd}(M)r_M \delta^{2/3},
$$
for some universal constant $C\ge 1$.
\end{lemma}
\begin{proof}
Consider a shortest path in $\Gamma$ from $v$ to $w$, denoted
   $$
        (v=v_0, v_1,v_2,\dots, v_{t}=w)\,.
   $$
By definition,
   \begin{align*}
   \sd_{\Gamma}(v,w)
=
    \sum_{ i \in [t]} {\rm w}(v_{i-1},v_i).
\end{align*}

Our main technical objective is to show $t$, the number of edges in the path, is at most of order $\delta^{-1/3}$.
For now, assume that neither $v$ nor $w$ is one of the special vertices $\{u_s\}_{s\in [\ell]}$. We will first derive two claims about properties of the path.\\

\noindent
{\bf Claim 1:} $\{v_i\}_{i \in [t-1]} \subseteq \{u_s\}_{s\in [\ell]}$.
Let us prove by contraction.
Given that $v$ has only one neighbor $u_{s_v}$, $v_1 \in \{u_s\}_{s\in [\ell]}$.
For the same reason, $v_{t-1} \in \{u_s\}_{s\in [\ell]}$ as well.
Suppose $v_i \notin \{u_s\}_{s \in [\ell]}$ for some $1 < i < t-1$.
Then,
$$v_{i-1} = v_{i+1} = u_{s_{v_i}},$$
which makes $(v_0,v_1,\dots, v_{i-1},v_{i+1},\dots, v_t)$ a shorter path, and this is a contradiction. \\

\noindent
{\bf Claim 2:} For $1 \le i < j \le t - 1$ with $j-i>1$, then
$$
    \|X_{v_i} - X_{v_j}\| >  \delta^{1/3}.
$$
Let us assume the claim fails
that $\|X_{v_i} - X_{v_j}\| \le  \delta^{1/3}$.
Without lose of generality, from the first Claim we may assume
$ (v_i,v_{i+1},\dots, v_{j}) = (u_i,u_{i+1},\dots, u_j)$.
(The path $(v_i,v_{i+1},\dots, v_j)$ must have distinct vertices, otherwise it is not a shortest path.)
Then, given  $\Omega_{\rm dist}$,
$$
    \rp^{-1}\left(\frac{|N(u_i) \cap U'_j|}{|U'_j|}\right)
\le
    \delta^{1/3} +  0.01\delta
\le
    2\delta^{1/3}.
$$
Hence, by \eqref{eq: wToDist} we have
$$
    {\rm w}(u_i,u_j)
\le
    \|X_{u_i}-X_{u_j}\| + 0.05\delta.
$$
On the other hand,
\begin{align*}
   \sum_{i' \in [i+1,j]} {\rm w}(u_{i'-1},u_{i'})
= &
    \sum_{i' \in [i+1,j]} \left(\rp^{-1}\left(\frac{|N(u_{i'-1}) \cap U'_{i'}|}{|U'_{i'}|}\right)+ 0.04\delta\right) \\
\ge &
    \sum_{i' \in [i+1,j]} \left(\|X_{u_{i'-1}} -  X_{u_{i'}}\| -0.01\delta + 0.04\delta\right) \\
\ge &
    \|X_{u_i}-X_{u_j}\| + (j-i)0.03\delta
\ge
    \|X_{u_i}-X_{u_j}\| + 0.06\delta
>
    {\rm w}(u_i,u_j),
\end{align*}
which implies that
$$
    (v_0,v_1,\dots, v_i,v_j,v_{j+1},\dots, v_t)
$$
is a shorter path, a contradiction to the assumption of the path.
Thus, the claim holds.

\medskip

With the two claims been proved, we can bound $t$ from above.
By the second claim, for any $i \in [1, t -3]$,
$$
    {\rm w}(v_i,v_{i+1}) +  {\rm w}(v_{i+1},v_{i+2})
\ge
    \|X_{v_i}-X_{v_{i+1}}\| +0.03\delta
    +
     \|X_{v_{i+1}}-X_{v_{i+2}}\| +0.03\delta
\ge
    \|X_{v_{i+2}}-X_{v_i}\|
\ge
    \delta^{1/3}.
$$
Hence,
$$
    \sd_{\Gamma}(v,w)
\ge
    \sum_{i \in [t-2]} {\rm w}(v_i,v_{i+1})
\ge
    \max\left\{ \left\lfloor \frac{t-2}{2}\right\rfloor, 0 \right\}  \delta^{1/3}
\ge
    \left( \frac{t-4}{2}\right) \delta^{1/3}.
$$

On the other hand, by Lemma \ref{lem: GammaGH1},
\begin{align*}
    \sd_{\rm \Gamma}(v,w) \le 2{\rm diam}_{\rm gd}(M).
\end{align*}
Together we conclude that
$$
    5{\rm diam}_{\rm gd}(M)\delta^{-1/3}
\ge
    4{\rm diam}_{\rm gd}(M)\delta^{-1/3} +4
\ge
    t.
$$
The case when $v$ or $w$ is contained in $\{u_s\}_{s\in [\ell]}$ is the same and simpler, and the same upper bound for $t$ follows. We omit the proof here.

Now we are ready to prove the lemma.
Given the boundedness of $t$, we can show the following.
For each $i \in [t]$,
$$
    {\rm w}(v_i,v_{i-1})
\ge
    \|X_{v_i}-X_{v_{i-1}}\| - 0.01\delta.
$$
Further, from definition of $\Gamma(G)$, any weight does not exceed
$ 2\delta^{1/3} +  0.04 \delta \le 3 \delta^{1/3}$.

Finally, relying on the additional assumption that $\delta^{1/3} \le 0.1r_M$,
we can apply Lemma \ref{lem: EucDistGeoDist} to get, for $ i \in [t]$,
$$
    \sd_{\rm gd}(X_{v_i}, X_{v_{i-1}})
\le
    \|X_{v_i}-X_{v_{i-1}}\|
    (1+ \kappa \|X_{v_i}-X_{v_{i-1}}\|^2/2)
\le
    \|X_{v_i}-X_{v_{i-1}}\|
    + \frac{27}{2}r_M\delta.
$$
Then, we conclude that
\begin{align*}
   \sd_{\Gamma}(v,w)
\ge &
    \|X_{v_1} -X_{v_0}\| - \delta
    + \|X_{v_t}-X_{v_{t-1}}\| -\delta
    +
    \sum_{i \in [t-1]} \|X_{v_i}-X_{v_{i-1}}\|  \\
\ge &
    \sum_{i \in [t]}
    \left( \sd_{\rm gd}(X_{v_i}, X_{v_{i-1}}) -  \frac{27}{2}r_M \delta \right) - 2\delta\\
\ge &
    \sd_{\rm gd}(X_v,X_w) - {\rm diam}_{\rm gd}(M) r_M \delta^{2/3}
    -2\delta.
\end{align*}

\end{proof}

\begin{proof}[Proof of the geodesic distance statement in Theorem~\ref{thm: distTechnical}:]
    The proof follows from Lemma~\ref{lem: GammaGH1} and Lemma~\ref{lem: GammaGH2}.
\end{proof}

\subsection{Proof for the Euclidean Distance Setting}

\begin{proof}[Proof of the Euclidean distance statement in Theorem~\ref{thm: distTechnical}:]

\noindent
{\bf Claim 1:} For $1 \le i < j \le \ell$, the shortest path in $\Gamma_{\rm euc}$ connecting $u_i$ and $u_j$ is $(u_i,u_j)$.

First, the path $(u_i,u_j)$ has length $ {\rm w}(u_i,u_j) \le \|X_{u_i} - X_{u_j}\| +0.05 \delta$ by Lemma \ref{lem: 4i2034i2342euc}.

Consider any shortest path $(u_i = v_0 ,v_1,\dots, v_t =u_j)$ in $\Gamma_{\rm euc}$ connecting $u_i$ and $u_j$.
Then, as illustrated in the proof of 1st Claim of Lemma \ref{lem: GammaGH2},
all vertices of this path are contained in $\{u_s\}_{s\in [\ell]}$.
Now suppose $t \ge 2$. Then, for $1 \le t' \le t$, we have ${\rm w}(v_{t'},v_{t'-1}) \ge \|X_{v_{t'}} -X_{v_{t'-1}}\| + 0.03\delta$. Hence, by Lemma \ref{lem: 4i2034i2342euc},
$$
    \mbox{length of }(u_i = v_0 ,v_1,\dots, v_t =u_j)
\ge
    \sum_{t' \in [t]}\|X_{v_{t'}} -X_{v_{t'-1}}\| + 0.03\delta
\ge
    \|X_{u_i} - X_{u_j}\| + 0.06\delta,
$$
which is strictly greater than the length of $(u_i,u_j)$. Thus, a contradiction follows and the claim holds.

\noindent
{\bf Claim 2:} In general, for any distinct $v,w \in {\bf V}$, the shortest path in $\Gamma_{\rm euc}$ connecting $v$ and $w$ is
$$
    \begin{cases}
        (v,u_{s_v}, u_{s_w}, w) & \mbox{if } v,w \notin \{u_s\}_{s\in [\ell]},\\
        (v, u_{s_w},w) & \mbox{if } v \notin \{u_s\}_{s\in [\ell]}, w \in \{u_s\}_{s\in [\ell]},\\
        (v, u_{s_v},w) & \mbox{if } v \in \{u_s\}_{s\in [\ell]}, w \notin \{u_s\}_{s\in [\ell]},\\
        (v,w) & \mbox{if } v,w \in \{u_s\}_{s\in [\ell]}.
    \end{cases}
$$
The last case was shown in the first claim.
It suffices to show the first case, as we will see soon that the second and third cases are similar.
Consider the shortest path $(v = v_0,v_1,\dots, v_t = w)$ in $\Gamma_{\rm euc}$ connecting $v$ and $w$.
Then, immediately we have $v_1 = u_{s_v}$ and $v_{t-1}=u_{s_w}$ since $v$ and $w$ are only connected to $u_{s_v}$ and $u_{s_w}$, respectively.
Next, $(u_{s_v}=v_1,v_2,\dots, v_{t-1}=u_{s_w})$ must also be the shortest path in $\Gamma_{\rm euc}$ connecting $u_{s_v}$ and $u_{s_w}$, as a subpath of $(v_0,v_1,\dots,v_t)$.
From the first claim, we know $(u_{s_v}=v_1,v_2,\dots, v_{t-1}=u_{s_w}) = (u_{s_v}, u_{s_w})$. Hence, the first case follows. The claim holds.

Finally, by Lemma \ref{lem: 4i2034i2342euc},  for $v,w \notin \{u_s\}_{s\in [\ell]}$,
\begin{align*}
    \de(v,w)=\sd_{\Gamma_{\rm euc}}(v,w)  \le&  \delta +  \|X_{u_{s_v}} - X_{u_{s_w}}\| + 0.05\delta + \delta
    \le \|X_{u_{s_v}} - X_{u_{s_w}}\| + 2\delta \\
    &\le \|X_v - X_w\| + \|X_v - X_{u_{s_v}}\| + \|X_w - X_{u_{s_w}}\| + 2\delta
    \le \|X_v -X_w\| + 4\delta.
\end{align*}
Also, the lower bound can be establish in the same way:
\begin{align*}
    \de(v,w)=\sd_{\Gamma_{\rm euc}}(v,&w)  \ge {\rm w}(u_{s_v},u_{s_w})
    \ge \|X_{u_{s_v}} - X_{u_{s_w}}\| + 0.03\delta \\
    &\ge \|X_v - X_w\| - \|X_v - X_{u_{s_v}}\| - \|X_w - X_{u_{s_w}}\| + 0.03\delta
    \ge  \|X_v - X_w\| - 2\delta.
\end{align*}
The same estimate when $v \in \{u_s\}_{s\in [\ell]}$ or $w \in \{u_s\}_{s\in [\ell]}$
can be established in the same way, we will omit the proof here. Therefore, the theorem follows.
\end{proof}

\subsection{Gromov-Hausdorff Distance: Proof of Theorem~\ref{thm: geodGHInformal}}
Recall that $\Omega_1$ is an event of
$$
    {\rm Z} := (X_{{\bf V}},\eU_{{\bf V}_1},  \{X_{V_s}, \eU_{U_s,V_s}\}_{s \in [\ell]}).
$$
{\bf The discussion in this subsection is always conditioned on a sample ${\rm Z}={\rm z} \in \Omega_1$.}
It is worth to make a few remarks:
\begin{itemize}

    \item Recall that $V_0 \subseteq {\bf V}_2$ is a vertex set size $n$ and
$\Omega_1$ is not an event of $(X_{V_0}, \eU_{V_0,\bigcup U_s'})$.
\item While we have not condition on $\Omega_{\rm dist} \subseteq \Omega_1$, the graph $\Gamma(G,\rr)$ is well-defined for any $\rr \ge 0$.
\item For $v \in V_0$, recall the definition of $s_v$ for $v \in V_0 $ from the definition of $\Gamma(G,\rr)$ (regardless of the choice of $\rr$):
$$
    s_v = {\rm argmax}_{s \in [\ell]} \frac{|N(v) \cap U_s'|}{|U_s'|}.
$$
\end{itemize}

Now, we define a (random) measure $\nu$ on $\{u_s\}_{s\in [\ell]}$ as follows:
$$
    \nu(\{u_s\})
=
    \frac{ |\{ v \in V_0\,:\, s_v =s \}|}{ |V_0|}.
$$
Fix any $v \in V_0$, given the fact that $(X_w, \eU_{\{w\},\bigcup_{s \in [\ell]} U_s'})$ for $w \in V_0 \setminus \{v\}$
are i.i.d. copies of $(X_v, \eU_{\{v\},\bigcup_{s \in [\ell]} U_s'})$, which in turn implies that the conditional random variables
$
    \{s_w\}_{w \in V_0 }
$
are i.i.d. copies. Hence, $\nu$ is the empirical measure of the measure $\nu_0$ defined by
$$
    \nu_0(\{u_s\})
=
    \mathbb{P}\{ s_v = s\}.
$$

Before we proceed, let us emphasize that both $\nu$ and $\nu_0$ are random measures, due the fact that $\{U_s,u_s\}_{s \in [\ell]}$ are random. Though, to avoid the complexity of notation, we will not explicitly write the dependence on the sample ${\rm z} \in \Omega_1$.

\begin{lemma}
    \label{lem: nu0nu}
    Fix any sample ${\rm z} \in \Omega_1$.
    With probability $1-n^{-\omega(1)}$ (on $(X_{V_0}, \eU_{V_0, \bigcup_{s \in [\ell]}U_s'})$), the total variation distance of $\nu$ and $\nu_0$ is bounded by $n^{-1/4}$. That is, for any $U \subseteq \{u_s\}_{s \in [\ell]}$,
    $$
        |\nu_0(U) - \nu(U)| < n^{-1/4}.
    $$
    We denote the above event by $\Omega_{\nu,\nu_0} \subseteq \Omega_1$.
\end{lemma}

\begin{proof}
    Notice that $n\nu(\{u_s\})$  is the sum of $n$ i.i.d. Bernoulli random variables with probability $\nu_0(\{u_s\})$.
    For each $s \in [\ell]$, we can apply the Hoeffding's inequality,
    for $t \ge 0$,
    $$
       \mathbb{P}\left\{
           \big| \nu(\{u_s\}) - \nu_0(\{u_s\})\big| \ge t
        \right\}
    \le
        2\exp\left( - 2\frac{ (tn)^2}{n}\right).
    $$
    By taking
    $
        t =  \frac{\log(n)}{\sqrt{n}}
    $
     and apply the union bound argument,  we have

    $$
       \mathbb{P}\left\{  \exists s \in [\ell],\,
           \big| \nu(\{u_s\}) - \nu_0(\{u_s\})\big| \ge  \frac{\log(n)}{\sqrt{n}}
        \right\}
    =
        n^{-\omega(1)},
    $$
    where we relied on $\ell \le n^{3\varsigma/4}$ from $\Omega_{\rm netFound} \supseteq \Omega_1$.

    Finally, for $U' \subseteq \{u_s\}_{s\in [\ell]}$,
    \begin{align*}
        |\nu(U') - \nu_0(U')|
    &\le
        \sum_{s \in [\ell]}
        \frac{\log(n)}{\sqrt{n}}
    \le
        \frac{\log(n)}{\sqrt{n}}
        \ell
    \le
        \log(n) n^{-1/2} n^{3\varsigma/4}
    <
        n^{-1/4}.
    \end{align*}

\end{proof}

\begin{proof}[Proof of Theorem~\ref{thm: geodGHInformal}]
    Recall the event $\Omega_1$ introduced in the last section, which happens with probability $1-n^{-\omega(1)}$.
    Now we fix a realization ${\rm z} \in \Omega_1$. We will show that the statements of the Theorem hold within the realization.

    We define the graph $\tilde \Gamma$ to be the induced subgraph of $\Gamma$ with vertex set $ V'=\{u_s\}_{s \in [\ell]}$.

    Notice that there is a natural coupling $\pi_{\mu,\nu_0}$ of  $\mu$ and $\nu_0$
    in the generation of the random graph $\Gamma(G)$:
    Fix $v \in V_0$, consider the $(X_v,\eU_{\{v\},\bigcup_{s \in [\ell]} U_s'})$. We have $X_v \sim \mu$ and $s_v = s_v(X_v,\eU_{\{v\},\bigcup_{s \in [\ell]} U_s'}) \sim \nu_0$.
    Now, let $v'$ be another vertex of $V_0$.
    Then, $(X,u) = X(v),u_{s_v})$ and $(X',u') = (X(v'),u_{s_{v'}})$ are two independent copies of random pairs with distribution $\pi_{\mu,\nu_0}$. \\

   Consider the event $\Omega_{\rm dist} \subseteq \Omega_1$ introduced in the last section. Within the event, we have
    $\|X_v - X_{u_{s_v}}\| \le 1.01\delta$. Further, from Proposition \ref{prop: GHdist},
    \begin{align}
    \label{eq: distGH01}
        \mathbb{P}\{\Omega_{\rm dist}^c \,\vert\, {\rm Z}={\rm z}\}
    =
        n^{-\omega(1)}.
    \end{align}

    Then, by Theorem \ref{thm: distTechnical}, within the event $\Omega_{\rm dist}$, we have
    \begin{align}
    \nonumber
         | \sd_{\rm gd}(X,X') - \sd_{\tilde \Gamma}(u,u') |
    \le &
        \sd_{\rm gd}(X_v,X_{u_{s_v}}) +  \sd_{\rm gd}(X_{v'},X_{u_{s_{v'}}})
        +
        | \sd_{\rm gd}(X_{u_{s_v}},X_{u_{s_{v'}}}) - \sd_{\Gamma}(X_{u_{s_v}},X_{u_{s_{v'}}})| \\
    \label{eq: distGH00}
    \le &
        \delta + \delta + C {\rm diam}_{\rm gd}(M)r_M \delta^{2/3}
    \le
        C' {\rm diam}_{\rm gd}(M)r_M \delta^{2/3}.
    \end{align}

    \medskip

    Next, given the total variation distance between $\nu_0$ and $\nu$ is almost $n^{-1/4}$ from Lemma \ref{lem: nu0nu},
    we also know there exists a coupling $\pi_{\nu_0,\nu}$ such that for $(u,w) \sim \pi_{\nu_0,\nu}$,
    $$
        \mathbb{P}\{ u \neq w\} < n^{-1/4}.
    $$

    \medskip

    We can further couple $(\mu,\nu,\nu_0)$ together to get a measure $\pi_{\mu,\nu_0,\nu}$,
    with the marginal distribution corresponding to $(\mu,\nu_0)$ and $(\mu,\nu)$ being $\pi_{\mu,\nu_0}$ and $\pi_{\nu_0,\nu}$, respectively.
    While such a coupling is not unique, for any $\pi_{\mu,\nu_0,\nu}$, the marginal distribution corresponding to $(\mu,\nu)$ is a probability measure $\pi$ such that for a pair of independent copies of $(X,u,w), (X,u',w') \sim \pi$,
    \begin{align*}
        \mathbb{P} \big\{
            | \sd_{\rm gd}(X,X') - \sd_{\Gamma}(w,w') |  \ge 1.01\delta
        \big\}
    &\le
        \mathbb{P} \big\{
            | \sd_{\rm gd}(X,X') - \sd_{\Gamma}(u,u') |  \ge 1.01\delta
        \big\}
    +
        \mathbb{P} \big\{
            u=w
        \big\}
    +
    \mathbb{P} \big\{
            u'=w'
        \big\} \\
    &\le
        n^{-\omega(1)}+2n^{-1/4},
    \end{align*}
    and the first statement about the coupling in the geodesic distance setting follows.
    \medskip

    Now we move to the proof of the second statement.
    Same to the derivation of  \eqref{eq: distGH00}, we can apply Theorem \ref{thm: distTechnical} to show that: Given the event $\Omega_{\rm dist} \subseteq \Omega_1$,
    \begin{align*}
       \forall \tilde u \in \{ u_s\}_{s\in [\ell]},\,
        | \sd_{\rm gd}(X,X_{\tilde u}) - \sd_{\Gamma}(u,\tilde u) |
    \le
        C' {\rm diam}_{\rm gd}(M)r_M \delta^{2/3}.
    \end{align*}
    Thus, for any $\tilde u \in \{ u_s\}_{s\in [\ell]}$ and $t >0$, we have
    \begin{align*}
        & \mathbb{P} \big\{
            X \in B_{\rm gd}(X_{\tilde u}, t + C' {\rm diam}_{\rm gd}(M)r_M \delta^{2/3})\,\vert {\rm Z} = {\rm z}
        \} - \mathbb{P}\{\Omega_{\rm dist}^c \,\vert\, {\rm Z} = {\rm z}\}\\
    \le &
        \mathbb{P} \big\{
            u \in B_{\Gamma}(\tilde u, t)
        \big\}\\
    & \le
        \mathbb{P} \big\{
            X \in B_{\rm gd}(X_{\tilde u}, t + C' {\rm diam}_{\rm gd}(M)r_M \delta^{2/3})
        \,\vert\, {\rm Z} = {\rm z}\} + \mathbb{P}\{\Omega_{\rm dist}^c \,\vert\, {\rm Z} = {\rm z}\}.
    \end{align*}
    Now, if we combine the above inequality with \eqref{eq: distGH01} and
    $$
        \big| \mathbb{P}\{ w \in B_{\Gamma}(\tilde u, t) \} -
\mathbb{P} \big\{  u \in B_{\Gamma}(\tilde u, t) \big\} \big| < n^{-1/4}
    $$
    from the total variation distance estimate Lemma \ref{lem: nu0nu}, the second statement follows.

    It remains to these two statements for the coupling with respect to Euclidean distance. However, the proof is identical to the one in the geodesic distance setting, with the difference been replacing \eqref{eq: distGH00} by the corresponding estimate given Theorem~\ref{thm: distTechnical} for the Euclidean distance setting. We omit the proof here.
\end{proof}

\section{Open Problems}
Here we give a list of open problems.
\begin{enumerate}
    \item \textbf{Efficiency:} To approximate the manifold $M$ within a distance $\delta$, it's necessary and sufficient to construct a net of $M$ with a mesh size no greater than $\delta$. A simple volumetric argument indicates that the size of the net is of order $\delta^{1/d}$, where $d$ is the dimension of $M$.
    In our setting, while we do not know the underlying distance since only a graph is given, our result indicates that this is possibly true since we can produce a net of $M$ with mesh size at most $n^{-c/d}$, where $n$ is the number of vertices.
    Then, our algorithm or perhaps our analysis on the performance is suboptimal in two primary aspects:
    \begin{itemize}
        \item Our proof can only guarantee the algorithm to produce the desired result when the number of vertices $n$ is at least larger than $r_M^{-d^2\log(d)}$, where $r_M$ is the radius within which $M$ appears as a flat space of dimension $d$, centered at any point on $M$. Ideally, the order should be $r_M^{-d}$. This discrepancy primarily arises from the difficulty in comparison between $\max_{v,w \in W}\rK(X_v,X_w)$ and $\max_{v,w \in W} \frac{|N(v)\cap N(w) \cap W|}{|W|}$ for a typical batch $W$, which can be reduced to finding good upper and lower estimates of $|\rK(x,x) - \rK(x,y)|$ when $y$ is not far from $x$.
        \item Furthermore, even when the vertex count is sufficiently large, our algorithm yields a net with a mesh size of at most $n^{-c/d}$, rather than the desired $n^{-1/d}$.
    \end{itemize}

    \item \textbf{Generalization on the Manifold Constraints:} The algorithm seems adaptable to scenarios where the manifold is not connected, featuring several components of potentially varying dimensions. Detecting the local dimension at a given point is feasible by continuously identifying clusters, which then serve to construct ``orthogonal vectors,'' as discussed in the proof ideas. However, we have not considered the case where the manifold has a boundary.

    \item \textbf{Sparse Random Geometric Graphs:} One can naturally extend the definition of the random geometric graph to a sparse setting, namely connecting an edge $\{v,w\}$ with probability $\lambda(n)\rp(\|X_v - X_w\|)$, where $\lambda(n)$ is a sparse parameter. It seems promising that our result can be adapted to this setting when $\lambda(n) = n^{-c}$ with a sufficiently small $c>0$ without much modification and can produce a net at the cost of a larger mesh size. In general, it is interesting to understand how this question can be answered in a sparse setting.

    \item \textbf{Noisy Setting:} In this work, we assume that the vertices are sampled exactly from the manifold. However, in manifold learning it is often more realistic to assume that the samples are drawn from a distribution concentrated in a neighborhood of the manifold (see, e.g., \cite{JMLR:v26:25-0183,aizenbud2021non}, where the authors study estimation of the projection of a query point onto the manifold). An interesting direction for future work is to investigate how our results extend to this noisy setting.

\end{enumerate}

\newpage

\bibliographystyle{alpha}
\bibliography{email}

\newpage
\appendix

\section{Embedded Riemannian Manifold}
\label{sec: ERM}
\subsection{Second fundamental form and distance between subspaces}

The goal of this subsection is to establish Lemma~\ref{lem: IItod}, which gives an upper bound for the distance between two subspaces in terms of the second fundamental form. We begin by recalling the definition of the second fundamental form. We then show Lemma~\ref{lem: 22d}, and use it to prove Lemma~\ref{lem: IItod}.

\smallskip

Let $M$ be a $d$-dimensional compact complete connected smooth manifold embedded in $\R^{N}$. Let $D_X$ be the standard directional derivative with respect to a smooth vector field $X$ and $P_{T_pM}$ be the orthogonal projection from $T_p\R^N$ to $T_pM$ (with respect to the Euclidean structure).

\begin{definition}\label{def:II}
The \defn{second fundamental form} of $M$ is
\begin{equation} \label{eq:II}
{\rm II}(X,Y)_p := D_{X_p}Y - P_{T_pM} D_{X_p}Y = P_{(T_pM)^\perp}(D_{X_p} Y).
\end{equation}
for smooth vector fields $X,Y$ in $M$ and $p \in M$. Let us remark that ${\rm II}(X,Y)_p$ is defined pointwise, since ${\rm II}(X,fY) = f{\rm II}(X,Y)$ for every smooth function $f$ on $M$.
\end{definition}

Define
\[
\kappa = \kappa_M := \max_{p \in M} \, \max_{u,v \in T_pM \cap \mathbb{S}^{N-1}} \|{\rm II}(u,v)_p\| > 0.
\]

For the rest of this subsection, let $I \subseteq \mathbb{R}$ be a nonempty interval. Let $\gamma: I \to M$ be a unit-speed smooth curve. For each $t \in I$, define $H_t := T_{\gamma(t)} M$. Note that we think of each tangent space $T_p M$ as the linear subspace of dimension $d$ in $\mathbb{R}^N$ containing the tangent vectors at $p \in M$. In particular, the zero vector $\vec{0}$ is always in $T_p M$.

\begin{lemma} \label{lem: 22d}
For any interior point $s \in I$, and for any $\vartheta > 0$, there exists $\varrho > 0$ such that for any $t \in (s-\varrho, s+\varrho) \cap I$, we have $\sd(H_s, H_t) \le (\kappa + \vartheta) \cdot |t-s|$.
\end{lemma}
\begin{proof}
Without loss of generality, let us assume $s=0$. Take any $\vartheta > 0$.

For each $t$ in an open neighborhood of $0$, let $X_1(t), \ldots, X_d(t)$ be smooth vector fields along $\gamma$ which form an orthonormal basis of $H_t$. Since $M$ is embedded in $\R^N$, they are smooth vector fields in $\R^N$ along $\gamma$. Hence, we can define $G(t) := P_{H_0^\perp}[X_1(t),\dots, X_d(t)] \in \mathbb{R}^{(N-d) \times d}$ where $P_{H_0^\perp}: \R^N \to H_0^\perp$ is the orthogonal projection to $H_0^\perp$. Observe that
\begin{align}
\|G(t)\|_{\rm op} &= \max_{w \in \mathbb{S}^{d-1}} \|G(t) \cdot w\| \notag \\
&= \max \left\{ \big\| P_{H_0^\perp}\big(w_1 X_1(t) + \cdots + w_d X_d(t)\big) \big\| \, : \, [w_1, \ldots, w_d]^\tp \in \mathbb{S}^{d-1} \right\} \notag \\
&= \max \left\{ \| P_{H_0^\perp} v\| \, : \, v \in H_t, \|v\| = 1 \right\} \notag \\
&= \sd(H_0, H_t). \label{eq:d-H0-Ht}
\end{align}

Notice that $G(0) \in \mathbb{R}^{(N-d) \times d}$ is the zero matrix. Consider the first-degree Taylor approximation $G(t)=A \cdot t+ B(t)$ where $A, B(t) \in \mathbb{R}^{(N-d) \times d}$ are matrices, where each entry $(B(t))_{ij}$ is of order $O(t^2)$. Therefore, for each $(i,j) \in [N-d] \times [d]$, there exists $\widetilde{\varrho}_{ij} > 0$ such that for any $t$ with $|t| < \widetilde{\varrho}_{ij}$, we have
\[
|(B(t))_{ij}| < \frac{\vartheta}{N} \cdot |t|.
\]

Now we take $\varrho := \min \{ \widetilde{\varrho}_{ij} \, | \, (i,j) \in [N-d] \times [d] \} > 0$. Thus, for any $t \in \mathbb{R}$ with $|t| < \varrho$, we have
\[
\|B(t)\|_{\rm op} \le \sqrt{(N-d)d} \cdot \frac{\vartheta}{N} \cdot |t| < \vartheta \cdot |t|.
\]
Therefore, for any $v \in \mathbb{R}^d$, we have by the triangle inequality that
\[
\|(G(t))(v)\| = \|(A\cdot t + B(t))(v)\|\le \|Atv\| + \|(B(t))(v)\|\le (\|A\|_{\rm op} + \vartheta) \cdot |t| \cdot \|v\|,
\]
whence
\begin{equation}\label{ineq:A-op-theta}
\sd(H_0, H_t) \overset{\eqref{eq:d-H0-Ht}}{=} \|G(t)\|_{\rm op} \le (\|A\|_{\rm op} + \vartheta) \cdot |t|,
\end{equation}
for any $t \in (-\varrho, \varrho)$.

Now let $\alpha = [\alpha_1, \ldots, \alpha_d]^\tp \in \mathbb{S}^{d-1}$ be a unit vector such that $\|A\alpha\| = \|A\|_{\rm op}$. Then
\[
Y(t) := \sum_{i=1}^d \alpha_i X_i(t)
\]
is a smooth unit vector field along the curve $\gamma$ in $M$, defined in an open neighborhood of $0$. Observe that
\[
A \alpha = G'(0)\alpha = P_{H_0^\perp} (D_{\gamma'(t)} \widetilde{Y})\Big|_{t=0},
\]
where $\widetilde{Y}$ is the vector field along $\gamma$ in $M$ given by $\widetilde{Y}_{\gamma(t)} := Y(t)$. Therefore, we have
\begin{equation}\label{ineq:A-op-kappa}
\|A\|_{\rm op} = \|A\alpha\| = \| {\rm II}(\gamma', \widetilde{Y})_{\gamma(0)} \| \le \kappa.
\end{equation}
Note that the inequality above is valid since both $\gamma'(0)$ and $\widetilde{Y}_{\gamma(0)} = Y(0)$ are unit vectors. We finish by combining \eqref{ineq:A-op-theta} and \eqref{ineq:A-op-kappa}.
\end{proof}

\begin{lemma} \label{lem: IItod}
For any interior points $a, b \in I$, we have $\sd(H_a,H_b) \le \kappa \cdot |b-a|$.
\end{lemma}

\begin{proof}
Without loss of generality, assume $a \le b$. Take an arbitrary $\vartheta > 0$. We define a non-decreasing sequence $\{a_i\}_{i=0}^{\infty}$ of real numbers as follows. Take $a_0 := a$. For each $i \ge 0$, define
\begin{equation}\label{eq:def-a-i+1}
a_{i+1} := \max \left\{ x \in [a_i, b] \, | \, \sd(H_{a_i}, H_x) \le (\kappa + \vartheta)(x-a_i) \right\}.
\end{equation}
We give two remarks about \eqref{eq:def-a-i+1}. First, we do not require that $\sd(H_{a_i}, H_t) \le (\kappa + \vartheta)(t-a_i)$ for all $a_i \le t \le x$. We only require the inequality to be true when $t = x$. Second, $a_{i+1}$ is well-defined since $x \mapsto \sd(H_{a_i}, H_x) - (\kappa + \vartheta)(x-a_i)$ is a continuous function, and hence the set in \eqref{eq:def-a-i+1} is a closed set.

Lemma~\ref{lem: 22d} implies that $a_{i+1} > a_i$ unless $a_i = b$. We argue that there exists $T < \infty$ such that $a_T = b$. Suppose otherwise for the sake of contradiction. Then $\{a_i\}_{i=0}^{\infty}$ is a {\it strictly} increasing sequence, and hence it converges to a limit $L \in (a,b]$.

Using Lemma~\ref{lem: 22d} again, we find that there exists $\varrho > 0$ such that for any $t \in (L-\varrho, L + \varrho) \cap I$, we have
\begin{equation}\label{ineq:HL-Ht}
\sd(H_L, H_t) \le (\kappa + \vartheta) \cdot |t-L|.
\end{equation}
Since $\{a_i\}_{i=0}^{\infty}$ converges to $L$, there exists $k \in \mathbb{Z}_{\ge 0}$ such that $L-\varrho < a_k < L$, and so \eqref{ineq:HL-Ht} gives
\[
\sd(H_{a_k}, H_L) \le (\kappa + \vartheta) \cdot |L - a_k|,
\]
which, by \eqref{eq:def-a-i+1}, implies that $a_{k+1} \ge L$. However, this means $a_{k+1} = a_{k+2} = \cdots = L$, which is a contradiction.

Therefore, there exists an index $T$ for which $a_T = b$. By the triangle inequality, we have
\[
\sd(H_a, H_b) \le \sum_{i=0}^{T-1} \sd(H_{a_i}, H_{a_{i+1}}) \le \sum_{i=0}^{T-1} (\kappa + \vartheta)(a_{i+1} - a_i) = (\kappa + \vartheta)(b-a).
\]
Since $\vartheta > 0$ was arbitrary, by taking $\vartheta \searrow 0$, the lemma follows.
\end{proof}

\subsection{Projection as a local diffeomorphism}
This subsection collects some useful geometric results for us to use in the paper. Let $M$ and $\kappa$ be as described in the previous subsection. Let $\sd_{\rm gd}(p,q)$ denote the geodesic distance between points $p, q \in M$. We use the notation $B_{\rm gd}(p,r) \subseteq M$ to denote the set of all points in $M$ with geodesic distance strictly less than $r$ from $p \in M$, while $B(p,r) \subseteq \mathbb{R}^N$ denotes the Euclidean open ball of radius $r$ centered at $p$.

\begin{proposition}\label{prop: SecondFundamentalForm}
Fix $p \in M$. Suppose $H$ is a $d$-dimensional affine subspace of $\mathbb{R}^N$ through $p$ such that $1- \sd(H,T_p M) =: \zeta > 0$ and $P_H: \mathbb{R}^N \to H$ is the orthogonal projection to $H$. Then
\begin{enumerate}
\item[(a)] $P_H$ is a diffeomorphism from $B_{\rm gd}(p,\zeta/10\kappa)$ to its image.
\item[(b)] $B(p, 0.09\zeta^2/\kappa)\cap H \subseteq P_H(B_{\rm gd}(p,\zeta/10\kappa))$.
\item[(c)] Let $\phi$ be the inverse of $P_H|_{B_{\rm gd}(p,\zeta/10\kappa)}$ so that $\phi(P_H(q)) = q$, for every $q \in B_{\rm gd}(p,\zeta/10\kappa)$. For any $v\in B(p, 0.09\zeta^2/\kappa)\cap H$, we have $\|\phi(v)-p\| \le \sd_{\rm gd}(\phi(v),p) \le \frac{1}{0.9\zeta}\|v - p\|$. In particular,
$$
    \sd_{\rm gd}(\phi(v),p)
\le
    \|v-p\|(1+\kappa^2\|v-p\|^2/2)
\le
    \|\phi(v)-p\| (1+\kappa^2\|\phi(v)-p\|^2/2)
$$
\item[(d)] If $H = T_p M$, then for any $v \in B(p,0.09/\kappa) \cap T_p M$, we have $\| P_{H^\perp}\phi(v) \| \le \kappa \|v-p\|^2$  and $\sd_{\rm gd}(\phi(v),p) \le  \|v-p\|(1+ \kappa^2\|v-p\|^2/2)$.

\item[(e)] If $M \cap B(p, 0.09 \zeta^2/\kappa)$ is connected, then $M \cap B(p, 0.09 \zeta^2/\kappa) \subseteq \phi( B(p, 0.09 \zeta^2/\kappa) \cap H)$.
\end{enumerate}
\end{proposition}

\begin{proof}
With a translation, we assume $p = \vec{0} \in \mathbb{R}^N$ throughout this proof.

{\bf (a)}
For any point $q \in M$ with $\sd_{\rm gd}(\vec{0}, q) < \zeta/10\kappa$, the triangle inequality and Lemma~\ref{lem: IItod} give
\[
1 - \sd(H,T_qM) \ge 1 - \sd(H,T_{\vec{0}}M) - \sd(T_{\vec{0}}M,T_qM) > \zeta - \kappa \cdot \frac{\zeta}{10\kappa} = 0.9 \zeta.
\]
For any $v \in T_qM$ with $\|v\|=1$, we have
\begin{equation}\label{ineq:0.9-zeta}
\|P_Hv\| = \sqrt{ 1- \|P_{H^\perp}v\|^2} \ge \sqrt{1-\sd(H,T_qM)^2} \ge 1-\sd(H,T_qM) \ge 0.9\zeta.
\end{equation}
Therefore, when we restrict the domain of $P_H$ to $M$, we find that the least singular value of the linear map
\[
(dP_H)_q\big|_{T_qM} = P_H\big|_{T_qM}: T_q M \to T_{P_H(q)} H
\]
is bounded below by $0.9\zeta > 0$. Therefore, at any point $q \in {B_{\rm gd}(\vec{0}, \zeta/10\kappa)}$, the map $P_H\big|_M$ is a local diffeomorphism.

The next step is to show that $P_H$ is a diffeomorphism when restricted to $B_{\rm gd}(\vec{0}, \zeta/10 \kappa)$. It suffices to show that $P_H\big|_{B_{\rm gd}(\vec{0}, \zeta/10 \kappa)}$ is injective. To that end, consider any $q,q' \in B_{\rm gd}(\vec{0},\zeta/10\kappa)$ with $q \neq q'$. We claim that $P_Hq \neq P_Hq'$.

Let $\gamma:[0,t_0] \to M$ be a shortest unit-speed geodesic connecting $q$ and $q'$ with $\gamma(0)=q$ and $\gamma(t_0)=q'$. Observe that $\gamma'(0)$ is a unit vector in $T_q M$, and therefore by \eqref{ineq:0.9-zeta}, we find that
\begin{equation}\label{ineq:PH-gamma}
\|P_H \gamma'(0)\| \ge 0.9 \zeta > 0.
\end{equation}
Hence, we can consider $v := \frac{P_H\gamma'(0)}{\|P_H\gamma'(0)\|}$.

We will show that $\langle P_H\gamma'(t) ,v \rangle >0$ for $t \in [0,t_0]$, which implies
\[
\langle P_H q' - P_H q, v \rangle = \langle P_H\gamma(t_0) - P_H\gamma(0), v \rangle >0,
\]
and hence $P_Hq' \neq P_Hq$.

To achieve that, we take any $t \in [0,t_0]$ and compare $\gamma'(t)$ and $\gamma'(0)$. Observe that
\[
\|D_{\gamma'(t)}\gamma'(t)\| = \sqrt{\|(D_{\gamma'(t)}\gamma'(t))^{\perp}\|^2 + \|(D_{\gamma'(t)}\gamma'(t))^{\top}\|^2}.
\]
Since $\gamma$ is a unit speed geodesic, the second summand inside the square root is $0$, and thus
\[
\|D_{\gamma'(t)}\gamma'(t)\| =\|(D_{\gamma'(t)}\gamma'(t))^{\perp}\|
= \|{\rm II}(\gamma'(t),\gamma'(t))\| \le \kappa.
\]
Therefore, we have
\begin{equation}\label{ineq:kappa-t}
\|\gamma'(t) - \gamma'(0)\| \le \kappa t.
\end{equation}

Notice that since $q,q' \in B_{\rm gd}(\vec{0}, \zeta/10\kappa)$, we have
\begin{equation}\label{ineq:0.2-zeta-kappa}
t \le t_0 = \sd_{\rm gd}(q,q') \le 0.2 \zeta/\kappa,
\end{equation}
by the triangle inequality.

Hence,
\begin{align*}
\langle P_H \gamma'(t), v \rangle &= \|P_H \gamma'(0)\| + \langle P_H (\gamma'(t) - \gamma'(0)), v \rangle \\
&\ge \| P_H \gamma'(0) \| - \| \gamma'(t) - \gamma'(0) \| \\
&\overset{\eqref{ineq:PH-gamma}, \eqref{ineq:kappa-t}}{\ge} 0.9 \zeta - \kappa t \\
&\overset{\eqref{ineq:0.2-zeta-kappa}}{\ge} 0.9 \zeta - \kappa \cdot 0.2 \zeta/\kappa = 0.7 \zeta > 0.
\end{align*}
We have finished the proof of part~(a).

\medskip

{\bf (b)}
Let $U := P_H(B_{\rm gd}(\vec{0},\zeta/10\kappa))$. From part~(a), $P_H: B_{\rm gd}(\vec{0},\zeta/10\kappa) \to U$ is a diffeomorphism. Let $\phi$ denote the inverse of $P_H$. Suppose, for the sake of contradiction, that there exists $z \in B(\vec{0}, 0.09 \zeta^2/\kappa) \cap H$ such that $z \notin U$. Since $\vec{0} \in U$, we have $z \neq \vec{0}$, and it makes sense to consider $u := \frac{z}{\|z\|}$, and let us define $S:=\{t \in \mathbb{R} \, : \, tu \in U\}$.

By construction, observe the following: (i) $S$ is an open set containing $0 \in \mathbb{R}$, (ii) $S$ is bounded, since every element $x \in S$ satisfies $|x| < \zeta/10\kappa$, and (iii) $\|z\| \notin S$. Therefore, the set
\[
S' := [0, \zeta/10\kappa] \setminus S
\]
is a compact set which contains $\|z\|$. We define $s_0 := \min S'$. Thus, we have
\begin{equation}\label{ineq:0.09-zeta-kappa}
0 < s_0 \le \|z\| < 0.09 \zeta^2/\kappa,
\end{equation}
and for any $s \in [0,s_0)$, we have $su \in U$, while $s_0 u \notin U$.

Consider the curve $\widetilde{\gamma}:[0,s_0) \to B_{\rm gd}(\vec{0},\zeta/10\kappa)$, given by $\widetilde{\gamma}(s) := \phi(su)$. Let $\gamma:[0,t_0) \to B_{\rm gd}(\vec{0},\zeta/10\kappa)$ be the unit-speed reparametrization of $\widetilde{\gamma}(s(t)) = \gamma(t)$, for a certain monotonically increasing function $s(t)$ with $s(0) = 0$. Note that $P_H \gamma(t) = s(t) \cdot u$, and so we have $s'(t) = \|P_H \gamma'(t)\|$.

For each $t \in [0,t_0)$, since $\gamma(t) \in B_{\rm gd}(\vec{0},\zeta/10\kappa)$ and $\|\gamma'(t)\| = 1$, we can use \eqref{ineq:0.9-zeta} to obtain $\|P_H \gamma'(t)\| \ge 0.9 \zeta$, and therefore
\begin{equation}\label{ineq:int-0-t0}
s_0 = \int_0^{t_0} s'(t) \, \sd t \ge 0.9 \zeta t_0.
\end{equation}
Combining \eqref{ineq:0.09-zeta-kappa} and \eqref{ineq:int-0-t0}, we find $t_0 < 0.1 \zeta/\kappa$. This implies that $\lim_{t \to t_0} \gamma(t) \in B_{\rm gd}(\vec{0},\zeta/10\kappa)$. However, this gives $s_0 u = \lim_{t \to t_0} P_H \gamma(t) \in U$, a contradiction.

\medskip

{\bf (c)} Each point $v \in B(\vec{0},0.09\zeta^2/\kappa) \cap H$ can be expressed as $v = s_0u$ where $s_0\in [0,0.09\zeta^2/\kappa)$ and $u\in H$ with $\|u\|=1$. Consider the same construction, as in part~(b), of a curve $\widetilde{\gamma}(s)= \phi(su)$ for $s\in [0,s_0)$ with the unit-speed reparametrization $\gamma(t) = \widetilde{\gamma}(s(t))$. The same argument from part~(b) implies that $\sd_{\rm gd}(\phi(s_0u),\vec{0}) \le t_0 \le \frac{s_0}{0.9\zeta} = \frac{\|v\|}{0.9 \zeta}$. The other inequality is clear, since $\sd_{\rm gd}(\phi(v),\vec{0})$ is the length of a geodesic joining $\phi(v)$ and $\vec{0}$, which must be bounded below by the Euclidean distance between the two points.

\medskip

{\bf (d)} Take any $v \in B(\vec{0}, 0.09/\kappa) \cap H$. Consider the same construction as in the previous steps. In particular, we have
\begin{itemize}
    \item the expression $v = s_0 u$ and so $\|v\| = s_0$,
    \item the curve $\widetilde{\gamma}:[0,s_0) \to M$, given by $\widetilde{\gamma}(s) = \phi(su)$, and
    \item the unit-speed reparametrization $\gamma:[0,t_0) \to M$ so that $\gamma(t) = \widetilde{\gamma}(s(t))$.
\end{itemize}
Note that we have $s_0 \ge 0.9 t_0$.

For each $t \in [0,t_0)$, using Lemma~\ref{lem: IItod}, we have
\begin{equation}\label{ineq:P-H-gamma}
\|P_{H^\perp} \gamma'(t)\| \le \sd(H, T_{\gamma(t)} M) = \sd(T_{\gamma(0)} M, T_{\gamma(t)} M) \le \kappa t,
\end{equation}
which by integration implies
\begin{equation}\label{ineq:kappa-t0-2}
\|P_{H^\perp} \phi(v)\| \le \frac{1}{2} \kappa t_0^2 \le \kappa s_0^2 = \kappa \|v\|^2.
\end{equation}
From \eqref{ineq:P-H-gamma}, we also obtain
\[
s'(t) = \|P_H \gamma'(t)\| \ge \sqrt{1-\kappa^2 t^2} \ge 1 - 0.51 \kappa^2 t^2,
\]
where we use $\kappa t \le \kappa t_0 \le \frac{\kappa s_0}{0.9} < 0.1$. Therefore, we find by integrating that
\[
s_0 = \int_0^{t_0} s'(t) \, {\rm d} t \ge t_0 - 0.17 \kappa^2 t_0^3 \ge t_0 \cdot \left( 1 - 0.3 \kappa^2 s_0^2 \right),
\]
and hence
\[
\sd_{\rm gd}(\phi(v), \vec{0}) \le t_0 \le \frac{s_0}{1-0.3 \kappa^2 s_0^2} \le s_0 \cdot \left(1+ \frac{\kappa^2 s_0^2}{2} \right).
\]

\medskip

{\bf (e)} Suppose that $N := M \cap B(\vec{0}, 0.09 \zeta^2/\kappa)$ is connected. Take an arbitrary point $q \in N$. We want to show that $q \in \phi(B(\vec{0}, 0.09 \zeta^2/\kappa) \cap H)$. Note that this is clear when $q = \vec{0}$, so for the rest of this proof let us assume $q \neq \vec{0}$. Because $N$ is connected, we can take a unit-speed path $\gamma:[0, t_0] \to N$ with $\gamma(0) = \vec{0}$ and $\gamma(t_0) = q$, for some $t_0 > 0$.

Since $P_H$ is a contraction, using part~(b), we find that for each $t \in [0,t_0]$,
\begin{align}
P_H(\gamma(t)) &\in P_H(M \cap B(\vec{0}, 0.09 \zeta^2/\kappa)) \label{in:PH-gamma-t} \\
&\subseteq P_H M \cap P_H(B(\vec{0}, 0.09 \zeta^2/\kappa)) \notag \\
&\subseteq H \cap B(\vec{0}, 0.09 \zeta^2/\kappa) \notag \\
&\subseteq P_H(B_{\rm gd}(\vec{0}, \zeta/10\kappa)), \notag
\end{align}
which shows that $P_H(\gamma(t))$ is in the domain of $\phi$. Hence, it makes sense to define $\widetilde{\gamma}:[0,t_0] \to M$ by $\widetilde{\gamma}(t) := \phi(P_H(\gamma(t)))$, for each $t \in [0,t_0]$. We claim that $\gamma(t) = \widetilde{\gamma}(t)$, for every $t \in [0,t_0]$.

Consider the set
\[
S'' := \left\{ t \in [0, t_0] \, : \, \gamma(t) = \widetilde{\gamma}(t) \right\}.
\]
For each $t \in S''$, from \eqref{in:PH-gamma-t}, we have
\[
\gamma(t) = \widetilde{\gamma}(t) \in \phi(H \cap B(\vec{0}, 0.09 \zeta^2/\kappa)) =: N'.
\]
Since $N'$ is an open set in $M$, there exists $\varrho > 0$ such that for each $t' \in (t-\varrho, t+\varrho) \cap [0,t_0]$, we have $\gamma(t') \in N'$. From parts~(a) and (b), we know that $\phi \circ P_H|_{N'}$ is the identity map on $N'$. Thus, $\widetilde{\gamma}(t') = \phi(P_H(\gamma(t'))) = \gamma(t')$, which implies that $t' \in S''$.

The argument from the previous paragraph shows that $S''$ is an open set in $[0,t_0]$. On the other hand, since the function $t \mapsto \gamma(t) - \widetilde{\gamma}(t)$ is continuous, $S''$ is also a closed set in $[0,t_0]$. We find that $S''$ is either $[0,t_0]$ or $\varnothing$. Because $0 \in S''$, we conclude that $S'' = [0,t_0]$, and in particular, $q = \gamma(t_0) = \widetilde{\gamma}(t_0)$.

Finally, using \eqref{in:PH-gamma-t} again, we find
\[
q = \widetilde{\gamma}(t_0) = \phi(P_H(\gamma(t_0))) \in \phi(B(\vec{0}, 0.09 \zeta^2/\kappa) \cap H),
\]
as desired.
\end{proof}

\begin{corollary}\label{cor:p-p'}
If $p, p' \in M$ satisfy $\|p-p'\| \le 0.01/\kappa$, then $\sd(T_p M, T_{p'} M) \le 2 \kappa \cdot \|p-p'\|$.
\end{corollary}
\begin{proof}
Let $H:=T_p M$, and let $P_H: \mathbb{R}^N \to H$ denote the orthogonal projection to $T_p M$. Let $v := P_H(p')$. Note that $\|v-p\| \le \|p'-p\|$. Proposition~\ref{prop: SecondFundamentalForm}(d) gives
\[
\sd_{\rm gd}(p',p) \le \|v-p\|(1+\kappa^2\|v-p\|^2/2) \le \|p'-p\|(1+\kappa^2\|p'-p\|^2/2) \le 2 \|p'-p\|,
\]
and thus Lemma~\ref{lem: IItod} yields $\sd(T_p M, T_{p'} M) \le \kappa \cdot \sd_{\rm gd}(p,p') \le 2 \kappa \cdot \|p-p'\|$.
\end{proof}

\begin{corollary}\label{cor:PH-diffeo}
In the setting of Proposition~\ref{prop: SecondFundamentalForm}, if $M' := M \cap B(p, 0.09 \zeta^2 / \kappa)$ is connected, then the projection $P_H$ is a diffeomorphism from $M'$ to its image.
\end{corollary}
\begin{proof}
Using parts~(a), (b) and (e) of Proposition~\ref{prop: SecondFundamentalForm}, we find that
\[
M \cap B(p, 0.09 \zeta^2/\kappa) \overset{{\rm (e)}}{\subseteq} \phi(B(p,0.09 \zeta^2/\kappa)\cap H) \overset{{\rm (b)}}{\subseteq} \phi(P_H(B_{\rm gd}(p, \zeta/10\kappa))) \overset{{\rm (a)}}{=} B_{\rm gd}(p, \zeta/10\kappa),
\]
and therefore, by part~(a) again, we conclude that $P_H|_{M \cap B(p, 0.09 \zeta^2 / \kappa)}$ is a diffeomorphism.
\end{proof}

\section{Proof of standard Geometric Lemmas and Probability Lemmas }\label{appx: basic result proof}
\subsection{Geometric Lemmas}

\begin{proof}[Proof of lem: sminB]

We can express
\[
B = r^2 Q^\tp Q + \widetilde B,
\]
where $Q = (q_{ij})_{i,j \in [0,k]} \in \mathbb{R}^{(k+1) \times (k+1)}$ is the upper-triangular matrix such that for $i,j \in [0,k]$,
\[
q_{ij} = \begin{cases}
    1 & \text{ if } i = j \text{ or } i = 0, \\
    0 & \text{ otherwise},
\end{cases}
\]
and $\widetilde B \in \mathbb{R}^{(k+1) \times (k+1)}$ is a matrix whose entries are uniformly bounded above by $15 r \delta$ in absolute value.

Notice that $Q^{-1} = (\widetilde q_{ij})_{i,j \in [0,k]}$ is simply the matrix such that for $i,j \in [0,k]$,
\[
\widetilde{q}_{ij} = \begin{cases}
1 & \text{ if } i = j, \\
-1 & \text{ if } i = 0 \text{ and } j \ge 1, \\
0 & \text{ otherwise}.
\end{cases}
\]
For example, when $k=3$,
\[
    Q = \begin{pmatrix} 1 & 1 & 1 & 1 \\ 0&1&0&0 \\ 0&0&1&0 \\ 0&0&0&1\end{pmatrix}
\qquad
\text{and}
\qquad
    Q^{-1} = \begin{pmatrix} 1 & -1 & -1 & -1 \\ 0&1&0&0 \\ 0&0&1&0 \\ 0&0&0&1\end{pmatrix}.
\]

Thus, we have the following upper bound on the operator norm\footnote{Because the matrix $Q$ is particularly nice, we can actually compute this operator norm explicitly. A more careful analysis yields
\[
\|Q^{-1}\|_{\rm op} = \sqrt{\frac{k+2 + \sqrt{k^2+4k}}{2}}.
\]
This shows that our upper bound is at most a multiplicative factor of $\sqrt{2}$ away from the actual value, which is good enough for our purpose. Thus instead of showing tedious calculation details, we decided to simply use this bound.} of $Q^{-1}$:
\[
\|Q^{-1}\|_{\rm op} \le \sqrt{\sum_{i,j \in [0,k]}\widetilde q_{ij}^2} = \sqrt{2k+1} < \sqrt{2(k+1)}.
\]

Therefore,
\[
s_{\min}(r^2 Q^\tp Q) = r^2 \cdot s_{\min}(Q)^2 \ge \frac{r^2}{2(k+1)}.
\]

Since we have the following bound on the operator norm of $\widetilde{B}$:
\[
\|\widetilde B\|_{\rm op}
\le \sqrt{\sum_{i,j \in [0,k]} \widetilde B_{ij}^2}
\le (k+1) \cdot 15 r \delta
= \frac{15(k+1)}{ \cdr d^2 } r^2
\le \frac{r^2}{4(k+1)},
\]
we conclude that
\[
    {\rm s}_{\min}(B) \ge \frac{r^2}{2(k+1)} - \frac{r^2}{4(k+1)} = \frac{r^2}{4(k+1)}.
\]

\end{proof}

\begin{proof} [Proof of \ref{lem:dHTX0M}]

{{\bf (a)}}  First, ${\rm s}_{\rm min}(Z^\tp Z) = ({\rm s}_{\rm min}(Z))^2$ since $N \ge d$.  Notice that, for $\alpha, \beta \in [d]$,
$$
    (Z^\tp Z)_{\alpha,\beta}
=
    \langle Z_\alpha, Z_\beta\rangle
=
    -\frac{1}{2} (\|Z_\alpha - Z_\beta\|^2 - \|Z_\alpha\|^2-\|Z_\beta\|^2).
$$

Now we shall estimate the summands. First,
\begin{align}
\Big| \|Z_\alpha\| - r \Big| &= \Big| \| X_i - X_{i_\alpha} \| - r \Big| \notag \\
&\le
    \Big| \|X_i-X_{i_\alpha}\| - \|X_{\inx}-X_{i_\alpha}\| \Big|
    + \Big| \|X_{\inx}-X_{i_\alpha}\| - \|X_{i_0}-X_{i_\alpha}\| \Big| + \Big| \|X_{i_0} - X_{i_\alpha}\| - r \Big| \notag \\
&\le
    \|X_i-X_{\inx}\|
    + \underbrace{\|X_{\inx} - X_{i_0}\|}_{{\cal E}_{\rm dist}(\inx,i_0)}
    + \underbrace{\Big| \|X_{i_0} - X_{i_\alpha}\| - r \Big|}_{{\cal E}_{\rm ao}(i_0, i_1, \ldots, i_d)} \notag \\
&\le
    \delta + 0.6 \delta + \delta < 3 \delta \label{ineq:wi-r-3},
\end{align}
which in turn implies
$$
    \big| \|Z_\alpha\|^2 -r^2 \big|
\le
    2r\cdot 3\delta + (3\delta)^2
\le
    7r\delta.
$$
Second,
\begin{align}
\label{ineq:wi-wj-sqrt-2-r}
   &  \Big| \|Z_\alpha - Z_\beta\| - \sqrt{2} r{\bf 1}(\alpha = \beta) \Big|
=
    \underbrace{\Big| \|-X_{i_\alpha} +X_{i_\beta}\| - \sqrt{2}r{\bf 1}(\alpha = \beta) \Big|}_{{\cal E}_{\rm ao}(i_0, i_1, \ldots, i_d)}
\le
    \delta \\
\nonumber
\Rightarrow &
    \big| \|Z_\alpha - Z_\beta\|^2 - 2r^2 {\bf 1}(\alpha = \beta) \big|
\le
    2r\delta + \delta^2
\le
    4r\delta.
\end{align}
Substituting these two bounds into \eqref{eq: dHTX0M00} we obtain
\begin{align*}
    \big| (Z^\tp Z)_{\alpha,\beta}  - r^2 {\bf 1}(\alpha = \beta)\big|
\le
    9r\delta.
\end{align*}
Hence, we can express
$$
    Z^\tp Z = r^2 {\rm Id} + B,
$$
where ${\rm Id}$ is the $d$ by $d$ identity matrix, and $B = (b_{\alpha,\beta})_{\alpha,\beta \in [d]}$ is a $d$ by $d$ matrix whose entries are bounded by $9r\delta$. Given this,  we can show
\begin{align}
\label{eq: dHTX0M01}
    {\rm s}_{\rm min}(Z^\tp Z) \ge r^2 - \| B\|_{\rm op}
    \ge r^2 -  \sqrt{ \sum_{\alpha,\beta \in [0,d]} b_{\alpha,\beta}^2}
    \ge r^2 - 9r\delta d
    \ge r^2/4,
\end{align}
and ${\bf (a)}$ follows.

{\bf (b)}
Similar to \eqref{eq: Hflat00}, we have
\begin{align}
\label{eq: dHTX0M00}
    \sd(H_Z,\TM)
=
    \max_{u \in H_Z\,:\, \|u\|=1} \| P_{\TM^\perp}u\|
\le
    \frac{ \| P_{\TM^\perp}Z\|_{\rm op}}{{\rm s}_{\rm min}(Z)}
\overset{\eqref{eq: dHTX0M01}}{\le}
    \frac{2\| P_{\TM^\perp}Z\|_{\rm op}}{r}.
\end{align}

Now we will establish an upper bound for $\|P_{\TM^\perp}Z\|_{\rm op}$.
The estimate in \eqref{ineq:wi-r-3} gives $\|Z_\alpha\| < r + 3\delta < r_M$. Thus, we can apply \eqref{eq: phiNorm} and obtain
\begin{align} \label{ineq:P-T-perp-16}
    \| P_{\TM^\perp} Z\|_{\rm op}
\le &
    \sqrt{ \sum_{\alpha \in [d]} \|P_{\TM^\perp} Z_\alpha\|^2}
=
    \sqrt{ \sum_{\alpha \in [d]} \|Z_\alpha - P Z_\alpha\|^2}
\overset{\eqref{eq: phiNorm}}{\le}
    \sqrt{ \sum_{\alpha \in [d]} (\kappa \|PZ_\alpha\|^2)^2}  \\
\nonumber
&\phantom{AAA AAA AAA}
\le
    \sqrt{ \sum_{\alpha \in [d]} (\kappa \|Z_\alpha\|^2)^2}
\le
   \kappa (2r)^2 \sqrt{d}
\le
   \frac{0.01}{r_M}\cdot 4r^2 \sqrt{d}
   \le
    \frac{r}{2^4\cdr}.
\end{align}

Substituting the bound for $\| P_{\TM^\perp} Z\|_{\rm op}$ into \eqref{eq: dHTX0M00} we conclude
$$
    \sd(H_Z,\TM)
\le
    \frac{r}{2^4\cdr} \cdot \frac{2}{r}
\le
    \frac{2^{-3}}{\cdr}.
$$

{\bf (c)}
By Corollary~\ref{cor:p-p'}, we have
\begin{align}\label{ineq:2-0.6}
    \sd(\TM, H_i)
\le
    2\kappa  \|X_{i_0} - X_{i}\|
\le
    2\kappa  \big(
        \underbrace{\|X_{i_0} - X_{\inx}\|}_{{\cal E}_{\rm dist}(\inx,i_0)}
    + \|X_{\inx}-X_i\|\big) \\
\nonumber
\le
    2 \kappa  (\delta + 0.6 \delta )
\le
    2\frac{0.01}{r_M}(\delta + 0.6 \delta )
<
     2^{-4}/\cdr.
\end{align}

By the triangle inequality, we have
\[
    \sd(H_Z, H_i)
\le
    \sd(H_Z, H_0)
    + \sd(H_0, H_i)
\overset{\eqref{ineq:2-0.6}, {\bf (b)}}{\le}
    2^{-2}/\cdr.
\]

\end{proof}

\subsection{Probability Estimates}
\begin{proof}[Proof of Lemma \ref{lem: cn}]
Recall that the adjacency matrix $A$ of $G$ is given by
\begin{align}
\label{eq: adjacencyMatrix2}
    a_{i,j}
=
    {\bf 1}( \eU_{i,j} \le \rp(\|X_i-X_j\|)),
\end{align}
where $ \{X_i\}_{i \in {\bf V}} \cup \{\eU_{i,j}\}_{i,j \in {\bf V}}$ are jointly independent up to the symmetric restriction $\eU_{i,j}=\eU_{j,i}$.

Write
\[
(\Ecn(W))^c = \bigcup_{\{i,j\} \in \binom{W}{2}} O_{\{i,j\}},
\]
where
\[
O_{\{i,j\}} := \left\{ \left| \frac{|N_W(i) \cap N_W(j)|}{n} - \rK(X_i, X_j) \right| > n^{-\frac{1}{2} + \varsigma} \right\}.
\]

Note that for any $\{i,j\} \in \binom{W}{2}$, we have
\begin{align*}
\left| \frac{|N_W(i) \cap N_W(j)|}{n} - \frac{|N_{W \setminus\{i,j\}}(i) \cap N_{W \setminus\{i,j\}}(j)|}{n-2} \right| &= \left( \frac{1}{n-2} - \frac{1}{n} \right) \cdot |N_W(i) \cap N_W(j)| \\
&= \frac{2}{n(n-2)} \cdot |N_W(i) \cap N_W(j)| \le \frac{2}{n}.
\end{align*}
Therefore, by the triangle inequality, the event $O_{\{i,j\}}$ is a subevent of the event
\[
O'_{\{i,j\}} := \left\{ \left| \frac{|N_{W \setminus\{i,j\}}(i) \cap N_{W \setminus\{i,j\}}(j)|}{n-2} - \rK(X_i,X_j) \right| \ge n^{-\frac{1}{2} + \varsigma} - \frac{2}{n} \right\}.
\]

Now, let us fix a pair $\{i,j\} \in \binom{W}{2}$. For each $k \in W \setminus \{i,j\}$, let $Z_k$ be the indicator that $k \in N_W(i) \cap N_W(j)$. From \eqref{eq: adjacencyMatrix2}, we have
$$
    Z_k
=
    {\bf 1}\big( \eU_{i,k} \le \rp(\|X_i- X_k\|) \big)
    {\bf 1}\big( \eU_{j,k} \le \rp(\|X_j- X_k\|) \big).
$$

According to the above expression and $ \{X_i\}_{i \in {\bf V}} \cup \{\eU_{i,j}\}_{i,j \in {\bf V}}$ are jointly independent,
we know that conditioning on $X_i=x_i$ and $X_j=x_j$, $\{Z_k\}_{k \in W \setminus \{i,j\}}$ are i.i.d. Bernoulli random variables with
$$
    \mathbb{E} \big[Z_k\, \big| \, X_i=x_i, X_j = x_j \big]
=
    \mathbb{E}_{X_k}\rp( \|X_k - x_i\|) \rp(\|X_k-x_j\|)
=
    \rK(x_i,x_j).
$$

Now we apply Hoeffding's inequality,
\begin{align*}
    & \mathbb{P} \big\{E'_{\{i,j\}}
    \,\big|\,  X_i = x_i,\, X_j=x_j\big\} \\
=&
     \mathbb{P} \bigg\{
        \Big|\sum_{k \in W \setminus \{i,j\}}Z_k - (n-2) \rK(X_i,X_j)\Big| \ge
        (n-2)\Big( n^{-\frac{1}{2} + \varsigma} - \frac{2}{n}\Big)
    \,\bigg|\,  X_i = x_i,\, X_j=x_j \bigg\} \\
\le &
    2 \cdot \exp\left( - 2(n-2) \left( n^{-\frac{1}{2} + \varsigma} - \frac{2}{n} \right)^2 \right)
=
    n^{-\omega(1)}.
\end{align*}
By Fubini's Theorem,

\[
    \mathbb{P}\{ O_{i,j}\}
\le
    \mathbb{P}\{ O'_{i,j}\}
=
    n^{-\omega(1)}.
\]

Now we use the union bound:
\begin{align*}
\mathbb{P}\{({\cal E}_{cn}(W))^c\} &\le \sum_{\{i,j\} \in \binom{W}{2}} \mathbb{P}\{O_{\{i,j\}}\}
\\
\le n^2 \cdot n^{-\omega(1)} = n^{-\omega(1)}.
\end{align*}

\end{proof}

\begin{proof}[Proof of Lemma \ref{lem: epsilonNetEvent}]

Consider any subset $\cal M \subseteq M$ such that every pair of points $p,q \in M$ satisfies $\|p-q\|\ge \frac{\varepsilon}{3}$.
Since the collection of open balls $\{B(p,\varepsilon/6)\}_{p\in \cal M}$
are pairwise disjoint,

\begin{align*}
   & 1
=
    \mu(M)
\ge
    \mu\Big( \bigcup_{p \in \cal M} B(p,\varepsilon/6)\Big)
=
    \sum_{p \in \cal M} \mu\big( B(p,\varepsilon/6)\big)
\ge
    |\cal M| \mu_{\rm min}(\varepsilon/6) \\
\Rightarrow \phantom{AAA AAA}&
    |\cal M|
\le
    \frac{1}{\mu_{\rm min}(\varepsilon/6)}
\le
    \frac{n^{\varsigma}}{2}.
\end{align*}

Now, we start with an empty set $\cal M$ and keep adding point from $M$ into $\cal M$ with the restriction that any pair of points within $\cal M$ has distance at least $\frac{\varepsilon}{3}$ until it is not possible to proceed.
This process must terminate at finite step due to the above bound. The resulting set $\cal M$ is an $\frac{\varepsilon}{3}$-net. That is, every point $q$ in $M$ is within distance $\frac{\varepsilon}{3}$ to some point $p \in \cal M$. From now on, we fix the set $\cal M$.

For each \(q \in {\cal M}\), by Hoeffding's inequality,
\begin{align*}
& \mathbb{P} \Big\{\Big| \left\{ i \in W \, : \, X_i \in B(q, 2\varepsilon/3) \right\} \Big|
\le
   \rv( 2 \varepsilon/3 )
  \cdot \frac{n}{2}\Big\}\\
& \le
\mathbb{P} \Big\{\Big| \left\{ i \in W \, : \, X_i \in B(q, 2\varepsilon/3) \right\} \Big|
\le
\mathbb{E} \Big| \left\{ i \in W \, : \, X_i \in B(q, 2\varepsilon/3) \right\} \Big|
      - n^{1-\varsigma}\Big\}\\
& \le
\exp\!\Big( - 2\frac{n^{2(1-\varsigma)}}{n}\Big)
      = \exp( - 2n^{1-2\varsigma}).
     \end{align*}

Taking a union bound, we have
\begin{align}
\label{eq:eventNet00}
\mathbb{P} \Big\{ \exists q \in {\cal M} \Big| \left\{ i \in W \, : \, X_i \in B(q, 2\varepsilon/3) \right\} \Big|
  \le
   \rv( 2 \varepsilon/3 )
  \cdot \frac{n}{2}\Big\}
& \le  \frac{n^\varsigma}{2} \exp( - 2n^{1-2\varsigma}) \\
& \le \exp(-n^{1/2}). \nonumber
\end{align}

Within the complement of the event $\Enet( W)$, there exists $p \in M$ such that
\[
\big| \left\{ i \in W \, : \, X_i \in B(p,\varepsilon) \right\} \big| < \rv(2\varepsilon/3) \cdot \frac{n}{2}.
\]
Let $q \in {\cal M}$ be a point such that $\|p-q\| < \varepsilon/3$. Then,

$$
 \big| \left\{ i \in W \, : \, X_i \in B(q, 2\varepsilon/3) \right\} \big|
\le
\big| \left\{ i \in W \, : \, X_i \in B(p, \varepsilon) \right\}\big| < \rv(2\varepsilon/3) \cdot \frac{n}{2}.
$$

This shows that the event in the estimate \eqref{eq:eventNet00} contains the event $(\Enet( W))^c$, and hence
\[
\mathbb{P} \{ (\Enet(W))^c \} \le \exp(-n^{1/2}).
\]
\end{proof}

\begin{proof}[Proof of Lemma \ref{lem: Enavi}]

We claim that
\[
    \mathbb{P} \big\{
        (\Enavi(\{V_\alpha\}_{\alpha \in [0,k]}, W))^c
    \,\big|\,
        X_{V}=x_{V}, \eU_{V}={\rm u}_{V}
    \big\}
\le
    2d \cdot n \cdot \exp\!\left( -2 n^{\varsigma} \right).
\]
Observe that the event $(\Enavi(\{V_\alpha\}_{\alpha \in [0,k]}, W))^c$ can be expressed as
\[
(\Enavi(\{V_\alpha\}_{\alpha \in [0,k]}, W))^c = \bigcup_{i \in W} \bigcup_{\alpha = 0}^k O_{i,\alpha},
\]
where
\[
O_{i,\alpha} := \left\{ \left| \frac{|N(i) \cap V_{\alpha}|}{|V_{\alpha}|} - \sum_{j \in V_{\alpha}} \frac{\rp(\|X_i - X_j\|)}{|V_{\alpha}|} \right| > n^{-1/2+\varsigma} \right\}.
\]

Let us fix a pair $(i,\alpha)$ with $i \in W$ and $\alpha \in [0,k]$.

For each $j \in V$, let $Z_j$ be the indicator of the edge $\{i,j\}$, which can be expressed as
$$
    Z_j := {\bf 1}( \eU_{i,j} \le \rp(\|X_i-X_j\|)).
$$
With this notation, we have
$$
    |N(i) \cap V_\alpha| = \sum_{j \in V_\alpha} Z_j.
$$
On the other hand, conditioning on $X_{V}=x_{V}$, $\eU_{V}={\rm u}_{V}$, and
$X_W = x_w$, we have $\{Z_j\}_{j \in V }$ are independent Bernoulli random variables with $\mathbb{E}Z_j=\rp(\|x_i-x_j\|)$ for each $j \in V$. Thus, we can apply Hoeffding's inequality, to obtain
\begin{align*}
   \mathbb{P} \big\{ O_{i,\alpha} \, \big| \, X_{V}=x_{V}, \eU_{V}= {\rm u}_{V}, X_W=x_w \big\}
\le
    2\exp( -2n^{\varsigma} ),
\end{align*}
where we rely on the condition $|V_\alpha| \ge n^{1-\varsigma}$
from the event $\Eclu(\{(V_\alpha, i_\alpha)\}_{\alpha \in [0,k]})$.

Hence, by the union bound with $|W| \le |{\bf V}| \le n^2$, we obtain
\[
\mathbb{P}\{\Enavi^c \,|\,
X_{V}=x_{V}, \eU_{V}=\eU_{V}, X_W=x_w\big\}
\le n^2 \cdot (k+1) \cdot 2 \exp\!\left( -2 n^{\varsigma} \right)
\le 2d \cdot n^2 \cdot \exp\!\left( -2 n^{\varsigma} \right)
= n^{-\omega(1)}.
\]

The second statement of the lemma simply follows by taking expectation with respect to $X_W$ on both sides of the above inequality.

\end{proof}

\section{The Graph Observer}\label{sec: GO}
The goal of the graph observer is to reconstruct geometric information of the underlying manifold $M$ from the random geometric graph $G$. In this subsection, we describe explicitly what are given to the graph observer. Proposition~\ref{prop: graph-obs}, a crucial practical result of this subsection, says how the graph observer may ``use'' the results in this paper. First, the proposition says there are ``graph-observer versions'' of parameters which the graph observer can compute exactly. Then the proposition gives a test which the graph observer can run on these parameters. If the parameters pass the test, then the graph observer may plug in these graph-observer versions of parameters as various parameters throughout the paper, and the parameters will be feasible.

\medskip

The graph observer is given the following information:
\begin{itemize}
    \item[(i)] an instance of the random graph $G$---for example, in terms of the adjacency matrix---together with the description of how the random graph is constructed,
    \item[(ii)] the property of the manifold $M$ that it is compact, complete, connected, and smooth, and that it is embedded in some Euclidean space $\mathbb{R}^N$ (even though $N$ is not given explicitly to the graph observer),
    \item[(iii)] the dimension $d \in \mathbb{Z}_{\ge 1}$ of the manifold $M$,
    \item[(iv)] the function $\rp:[0,\infty) \to [0,1]$,
    \item[(v)] an upper bound\footnote{As a rule of thumb, we use the superscript $^\circ$ to indicate parameters which the graph observer is given or can compute exactly. This circle superscript $^\circ$ is a mnemonic: it has the shape of the letter ``o'', the first letter in the word ``observer.''} $D^\circ > 0$, together with the information that $\diam(M) \le D^\circ$,
    \item[(vi)] an upper bound $\kappa^\circ > 0$, together with the information that $\kappa \le \kappa^\circ$,
    \item[(vii)] a lower bound $r_{M,0}^\circ > 0$, together with the information that $r_{M,0} \ge r_{M,0}^\circ$, and
    \item[(viii)] a constant $C^\circ > 0$, together with the information that
    \[
    \mu_{\min}(x) \ge C^\circ \cdot x^d,
    \]
    for every $x \in [0,r_M]$ (even though $r_M$ is not given explicitly to the graph observer).
\end{itemize}
The graph observer may take any $\varsigma \in (0,1/4)$ and take $\cdr = 2^{10}$. These two parameters, $\varsigma$ and $\cdr$, are considered known and need not be given to the graph observer.

Next, the graph observer computes:
\begin{itemize}
    \item[(i)] the positive integer $n = \left\lfloor |{\bf V}|^{1-2\varsigma} \right\rfloor$,
    \item[(ii)] $r_M^\circ := 0.01 \cdot \min\{1/\kappa^\circ, r_{M,0}^\circ \}$,
    \item[(iii)] the Lipschitz constant $L_\rp$,
    \item[(iv)]
    \[
    \ell_\rp^\circ := \min_{x \in [0, D^\circ]} |\rp'(x)|,
    \]
    \item[(v)]
    \[
    \varepsilon^\circ := \max\left\{ 6 \cdot (2/C^\circ)^{1/d} n^{-\varsigma/d}, \frac{n^{-1/2+\varsigma}}{\rp(D^\circ) \cdot L_\rp} \right\},
    \]
    \item[(vi)]
    \[
    c_1^\circ := \frac{1}{800} (\ell_\rp^\circ)^2 C^\circ (r_M^\circ/4)^d,
    \]
    \item[(vii)]
    \[
    C_2^\circ := 4 \cdot \frac{C_{\rm gap}^{1/2} d^{1/2} L_\rp}{(\ell_\rp^\circ)^{1/2}(c_1^\circ)^{1/2}},
    \]
    \item[(viii)]
    \[
    c_3^\circ := \frac{\ell^\circ_\rp}{\cdr \sqrt{d}},
    \]
    \item[(ix)]
    \[
    \eta^\circ := \max\left\{ C_2^\circ \cdot (\varepsilon^\circ)^{1/2}, \frac{L_\rp^2}{c^\circ_3 \ell^\circ_\rp \sqrt{d}} \cdot \varepsilon^\circ \right\},
    \]
    \item[(x)] $\delta^\circ := \cdr \sqrt{d} \cdot \eta^\circ$, and
    \item[(xi)] $r^\circ := \cdr d^2 \cdot \delta^\circ$.
\end{itemize}
It is not hard to see that all the eleven numbers computed above are strictly positive (and finite).

We note from the definition of $r_M^\circ$ that
\begin{equation}\label{ineq:r-M-circ}
r_M^\circ = 0.01 \cdot \min\{1/\kappa^\circ, r_{M,0}^\circ \} \le 0.01 \cdot \min\{1/\kappa, r_{M,0} \} = r_M.
\end{equation}
Moreover, we deduce from Corollary~\ref{cor:PH-diffeo} in the appendix that
\[
D^\circ \ge \diam(M) > 2 r_M,
\]
and therefore, by the definition of $\ell_\rp^\circ$, we obtain
\begin{equation}\label{ineq:ell-rp-circ}
\ell_\rp^\circ = \min_{x \in [0,D^\circ]} |\rp'(x)| \le \min_{x \in [0,2r_M]} |\rp'(x)| = \ell_\rp.
\end{equation}
Hence, both parameters $r_M^\circ$ and $\ell_\rp^\circ$ which the graph observer can compute exactly are lower bounds for their corresponding variables.

\begin{proposition}\label{prop: graph-obs}
We have the following items.
\begin{itemize}
\item[(a)] The parameters $\varepsilon^\circ$, $c_1^\circ$, $C_2^\circ$, $c_3^\circ$, $\eta^\circ$, $\delta^\circ$, $r^\circ$ can be computed exactly by the graph observer.
\item[(b)] {\rm (``the test'')} The two inequalities
\begin{equation}\label{ineq:feas-test}
r_M^\circ \ge 2^4 \cdr d^2 r^\circ \qquad \text{and} \qquad n \ge 100
\end{equation}
can be checked by the graph observer.
\item[(c)] If both inequalities in part~(b) hold, then the parameters $\varepsilon^\circ$, $c_1^\circ$, $C_2^\circ$, $c_3^\circ$, $\eta^\circ$, $\delta^\circ$, $r^\circ$ may play the roles of $\varepsilon$, $c_1$, $C_2$, $c_3$, $\eta$, $\delta$, $r$, respectively, throughout the paper, as they satisfy the parameter assumptions, and furthermore, these parameters are feasible.
\end{itemize}
\end{proposition}

\begin{proof}
The first two parts, (a) and (b), follow by design. Let us prove (c). Suppose that both inequalities in \eqref{ineq:feas-test} hold. We now check the parameter assumptions. First, we have
\[
r_M \overset{\eqref{ineq:r-M-circ}}{\ge} r_M^\circ \overset{\eqref{ineq:feas-test}}{\ge} 2^4 \cdr d^2 r^\circ > r^\circ > \delta^\circ > \eta^\circ > \varepsilon^\circ > \varepsilon^\circ/6.
\]
This shows that $\varepsilon^\circ/6 \in [0,r_M]$, and thus by the definitions of $C^\circ$ and $\varepsilon^\circ$, we have
\[
\mu_{\min}(\varepsilon^\circ/6) \ge C^\circ \cdot (\varepsilon^\circ/6)^d \ge C^\circ \cdot \frac{2}{C^\circ} \cdot n^{-\varsigma} = 2n^{-\varsigma}.
\]
We also have
\[
\varepsilon^\circ \ge \frac{n^{-1/2+\varsigma}}{\rp(D^\circ) L_\rp} \ge \frac{1}{\sqrt{\|\rK\|_{\infty}} L_\rp} \cdot n^{-1/2+\varsigma},
\]
where in the last inequality, we use the bound from Remark~\ref{rem: Lp-rK-gtr-0}. This shows that the condition for $\varepsilon$ in Subsection~2.3 is satisfied.

Second, the condition for $c_1$ in Subsection~2.3 is satisfied, since
\[
0 < c_1^\circ = \frac{1}{800} (\ell_\rp^\circ)^2 C^\circ (r_M^\circ/4)^d
\le \frac{1}{800} \ell_\rp^2 C^\circ (r_M/4)^d
\le \frac{1}{800} \ell_\rp^2 \mu_{\min}(r_M/4),
\]
where we use the definition of $C^\circ$, noting that $r_M/4 \in [0,r_M]$.

Third, the condition for $C_2$ in Subsection~2.3 is satisfied, since
\[
C_2^\circ = 4 \cdot \frac{C_{\rm gap}^{1/2} d^{1/2} L_\rp}{(\ell_\rp^\circ)^{1/2}(c_1^\circ)^{1/2}} \ge 4 \cdot \frac{\|\rK\|_{\infty}^{1/4} C_{\rm gap}^{1/2} d^{1/2} L_\rp}{(\ell_\rp)^{1/2}(c_1^\circ)^{1/2}},
\]
where we use the observation that $\|\rK\|_{\infty} \le 1$ from the definition of $\rK$ (see Definition~\ref{def:K}).

Next, the conditions for $c_3$ and $\eta$ in Subsection~2.3 follow from the definitions of $c^\circ_3$ and $\eta^\circ$ and the inequality $\ell^\circ_\rp \le \ell_\rp$ from \eqref{ineq:ell-rp-circ}. Finally, the condition for $\delta$ in Subsection~2.3 is immediate from the definitions and $\delta^\circ$ and $r^\circ$.

Having shown that $\varepsilon^\circ$, $c_1^\circ$, $C_2^\circ$, $c_3^\circ$, $\eta^\circ$, $\delta^\circ$, $r^\circ$ may play the roles of $\varepsilon$, $c_1$, $C_2$, $c_3$, $\eta$, $\delta$, $r$, we proceed to show that the parameters are feasible. This final step is easy. From \eqref{ineq:r-M-circ}, we know that $r_M \ge r_M^\circ$. Thus, with $r^\circ$ playing the role of $r$, if both inequalities in \eqref{ineq:feas-test} hold, then
\[
r_M \ge r_M^\circ \ge 2^4 \cdr d^2 r^\circ,
\]
and $n \ge 100$, implying that the parameters are feasible.
\end{proof}

\newpage

\section{An example where the step function \texorpdfstring{$\rp$}{p} cannot distinguish two different embedded manifolds}
\label{appx: step-function}
In this appendix, we construct an explicit family of examples of two embedded manifolds with different intrinsic geometry for which the step-function distance-probability function $\rp(x) := p \cdot {\bf 1}_{[0,r]}(x)$, as given in the introduction of the present article, fails to distinguish the two manifolds. (See Proposition~\ref{prop:two-helices-m-n} below.) Throughout this appendix, let $r > 0$ and $p \in (0,1)$.

For each positive integer $n$, let us define the function $\psi_n: \bbR \to \bbR$ by
\[
\psi_n(\theta) := 4 - 2 \cos(\theta) - 2 \cos(n \theta),
\]
for every $\theta \in \bbR$.

For each positive real number $R > 0$ and for each positive integer $n$, define the function $F_{R,n}: \bbR \to \bbR^4$ by
\[
F_{R,n}(\theta) := \left( R \cos \theta, R \sin \theta, R \cos(n\theta), R \sin(n \theta) \right), \quad \forall \theta \in \bbR.
\]
We let $\calM(R,n) \subseteq \bbR^4$ denote the $1$-dimensional manifold that is the image of $F_{R,n}$:
\[
\calM(R,n) = F_{R,n}(\bbR) = F_{R,n}\!\left( [0,2\pi) \right).
\]

The goal of this appendix is to prove the following proposition.

\begin{proposition}\label{prop:two-helices-m-n}
Let $r > 0$ and $p \in (0,1)$ be positive real numbers. Let $m > n$ be positive integers. Let $\rp:[0,\infty) \to [0,1]$ denote the distance-probability function given by $\rp(x) := p \cdot {\bf 1}_{[0,r]}(x)$, for every $x \ge 0$.

Let $\xi$ be any positive real number such that $0 < \xi < \frac{1}{m^2}$. Define
\[
R_m := \frac{r}{\sqrt{\psi_m(\xi)}}
\qquad \text{ and } \qquad
R_n := \frac{r}{\sqrt{\psi_n(\xi)}}.
\]
Let $\mu_m$ and $\mu_n$ denote the uniform measures on $\calM(R_m,m)$ and $\calM(R_n,n)$, respectively. Then for any positive integer $N$, the two random geometric graphs
\[
G(N, \calM(R_m,m), \mu_m, \rp)
\qquad \text{ and } \qquad
G(N, \calM(R_n,n), \mu_n, \rp)
\]
have the same distribution, while
\[
\vol(\calM(R_m,m)) > \vol(\calM(R_n,n)).
\]
Here, $\vol$ denotes the intrinsic volume of the manifold, which in this case is the total arc length.
\end{proposition}

In other words, the two random graph distribution {\em cannot} distinguish the two manifolds $\calM(R_m,m)$ and $\calM(R_n,n)$, even though they are different with respect to both their extrinsic and intrinsic geometry.

\begin{figure}
\includegraphics[scale = 1.0]{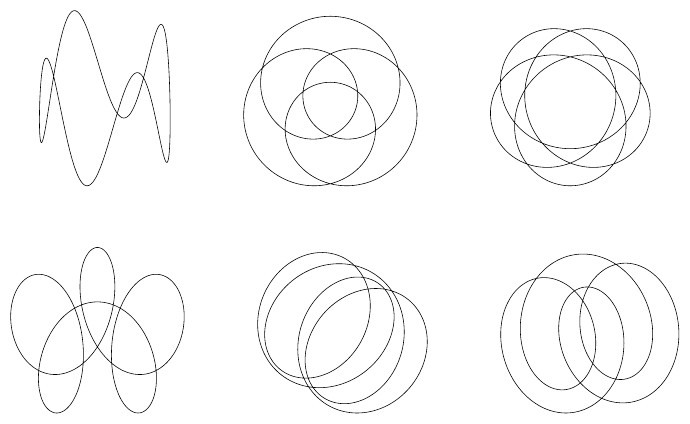}
    \caption{Six different projections of the curve $\calM(1,4) \subseteq \bbR^4$.}
    \label{fig:4-dim-helix-t-4t}
\end{figure}

In Figure~\ref{fig:4-dim-helix-t-4t}, we show six different projections of the curve $\calM(1,4) \subseteq \bbR^4$ onto a two-dimensional space.

For the rest of this appendix, we let $r, p, \rp, \xi, m, n, R_m, R_n, N, \mu_m, \mu_n$ be as in the assumptions of Proposition~\ref{prop:two-helices-m-n}. We would like to give arguments with explicit exact bounds. To that end, we record here a useful exact estimate for the cosine function, which may be proved using single-variable calculus.

\begin{lemma}\label{lem: exact-bounds-for-cosine-720-900}
For every real number $\theta \in [-\pi,\pi]$, we have the following exact bounds:
\[
1 - \frac{\theta^2}{2} + \frac{\theta^4}{24} - \frac{\theta^6}{720} \le \cos \theta \le 1 - \frac{\theta^2}{2} + \frac{\theta^4}{24} - \frac{\theta^6}{900}.
\]
\end{lemma}

The next lemma gives a property of the function $\psi_m$.

\begin{lemma}\label{lem: ball-r-intersects-nicely-in-helices-xi-tau-xi}
For any real number $\tau \in [-\pi, \pi]$, we have $\psi_m(\tau) \le \psi_m(\xi)$ if and only if $-\xi \le \tau \le \xi$.
\end{lemma}
\begin{proof}
($\Leftarrow$) Suppose that $-\xi \le \tau \le \xi$. Note that the derivative $\psi_m'$ is $\psi'_m(\theta) = 2 \sin \theta + 2m \sin(m \theta)$, from which it is immediate that $\psi'$ is strictly decreasing in $[-\pi/m,0]$, and strictly increasing in $[0,\pi/m]$. Since $\xi < \frac{1}{m^2} < \frac{\pi}{m}$, we obtain $\psi_m(\tau) \le \psi_m(\xi)$.

($\Rightarrow$) We prove the contrapositive. Suppose that $|\tau| > \xi$. We want to show that $\psi_m(\tau) > \psi_m(\xi)$. Since $\psi_m$ is an even function ($\psi_m(-\tau) = \psi_m(\tau)$), we may assume that $\tau > \xi$. Since $\psi_m$ is strictly increasing on $[\xi,\pi/m]$, if $\xi < \tau \le \pi/m$, we finish.

Now assume $\tau > \pi/m$. In this case, we have
\[
\psi_m(\tau) = 4 - 2 \cos(\tau) - 2 \cos(m\tau) \ge 2 - 2 \cos(\tau) > 2 - 2 \cos\!\left( \frac{\pi}{m} \right),
\]
and
\[
\psi_m(\xi) = 4 - 2 \cos(\xi) - 2 \cos(m \xi) \le 4 - 2 \cos\!\left( \frac{1}{m^2} \right) - 2 \cos\!\left( \frac{1}{m} \right).
\]
It is an exercise in single-variable calculus to show that
\[
2 - 2 \cos\!\left( \frac{\pi}{m} \right) \ge 4 - 2 \cos\!\left( \frac{1}{m^2} \right) - 2 \cos\!\left( \frac{1}{m} \right)
\]
holds for every positive integer $m \ge 1$. For example, one can use the estimates provided in Lemma~\ref{lem: exact-bounds-for-cosine-720-900}. Hence, we obtain $\psi_m(\tau) > \psi_m(\xi)$. We have finished the proof.
\end{proof}

Note that since $m > n$, we also have that $\xi < \frac{1}{n^2}$ as well. This shows that the same argument as in the proof of Lemma~\ref{lem: ball-r-intersects-nicely-in-helices-xi-tau-xi} implies that for any real number $\tau \in [-\pi,\pi]$, we have $\psi_n(\tau) \le \psi_n(\xi)$ if and only if $-\xi \le \tau \le \xi$.

A geometric consequence of Lemma~\ref{lem: ball-r-intersects-nicely-in-helices-xi-tau-xi} is that every open ball of radius $r$ centered at a point on $\calM(R_m,m)$ (or $\calM(R_n,n)$) intersects ``nicely'' with the manifold. We present this as the following lemma.

\begin{lemma}\label{lem: intersect-nicely-B-F-theta-r-int-M}
For every real number $\theta \in \bbR$, we have
\begin{itemize}
    \item[(a)] $\overline{B}(F_{R_m,m}(\theta),r) \cap \calM(R_m,m) = F_{R_m,m}\!\left( [\theta - \xi, \theta + \xi] \right)$, and
    \item[(b)] $\overline{B}(F_{R_n,n}(\theta),r) \cap \calM(R_n,n) = F_{R_n,n}\!\left( [\theta - \xi, \theta + \xi] \right)$.
\end{itemize}
Here, $\overline{B}(x,r) \subseteq \bbR^4$ denotes the closed ball of radius $r$ centered at $x \in \bbR^4$.
\end{lemma}

\begin{proof}
Let us show~(a). The proof for~(b) is analogous.

For every $\alpha \in \bbR$, the orthogonal matrix
\[
T_\alpha :=
\begin{bmatrix}
    \cos\alpha & -\sin\alpha & 0 & 0 \\
    \sin\alpha & \cos\alpha & 0 & 0 \\
    0 & 0 & \cos(m\alpha) & -\sin(m\alpha) \\
    0 & 0 & \sin(m\alpha) & \cos(m\alpha)
\end{bmatrix}
\]
is an isometry $T_\alpha: \calM(R_m,m) \to \calM(R_m,m)$ such that
\[
T_\alpha F_{R_m,m}(\theta) = F_{R_m,m}(\theta + \alpha),
\]
for every $\theta \in \bbR$. Using this symmetry, it suffices to show that
\[
    B(F_{R_m,m}(0),r) \cap \calM(R_m,m) = F_{R_m,m}\!\left( (-\xi, \xi) \right).
\]
Observe that for any $\tau \in (-\pi,\pi]$, we have
\[
    \| F_{R_m,m}(\tau) - F_{R_m,m}(0) \| = R_m \sqrt{\psi_m(\tau)}.
\]
Hence,
\[
F_{R_m,m}(\tau) \in B(F_{R_m,m}(0),r) \cap \calM(R_m,m)
\]
if and only if $R_m \sqrt{\psi_m(\tau)} < r = R_m \sqrt{\psi_m(\xi)}$, which is if and only if $\psi_m(\tau) < \psi_m(\xi)$. Thus, we finish by invoking Lemma~\ref{lem: ball-r-intersects-nicely-in-helices-xi-tau-xi}.
\end{proof}

\begin{lemma}
The two random geometric graphs
\[
G(N, \calM(R_m,m), \mu_m, \rp)
\qquad \text{ and } \qquad
G(N, \calM(R_n,n), \mu_n, \rp)
\]
have the same distribution.
\end{lemma}
\begin{proof}
We construct an explicit coupling between the two random geometric graphs. Let $\mu_0$ denote the uniform measure on the real interval $[0, 2\pi)$. The pushforward measures under $F_{R_m,m}$ and $F_{R_n,n}$ are the uniform measures $\mu_m$ and $\mu_n$ on $\calM(R_m,m)$ and $\calM(R_n,n)$, respectively. Now for each pair $(i,j) \in [N]^2$ with $i < j$, let $U_{i,j}$ be an independent uniform distribution on $[0,1]$.

Take $X_1, X_2, \ldots, X_N \overset{\text{i.i.d.}}\sim \mu_0$. For each $i \in [N]$, let $Y_i := F_{R_m,m}(X_i)$ and $Z_i := F_{R_n,n}(X_i)$ to be the latent points on $\calM(R_m,m)$ and $\calM(R_n,n)$, respectively. For each $i < j$, connect $Y_i$ and $Y_j$ in the random geometric graph if and only if
\[
\|Y_i - Y_j\| \le r
\qquad \text{ and } \qquad
U_{i,j} \le p.
\]
Analogously, connect $Z_i$ and $Z_j$ if and only if
\[
\|Z_i - Z_j\| \le r
\qquad \text{ and } \qquad
U_{i,j} \le p.
\]
Now by Lemma~\ref{lem: intersect-nicely-B-F-theta-r-int-M}, we have that for each $i, j$ with $i < j$,
\[
\|Y_i - Y_j\| \le r
\qquad \text{ if and only if } \qquad
|X_i - X_j| \le \xi
\qquad \text{ if and only if } \qquad
\|Z_i - Z_j\| \le r.
\]
Hence, $G(N, \calM(R_m,m), \mu_m, \rp)$ and $G(N, \calM(R_n,n), \mu_n, \rp)$ are identically realized under the coupling we constructed.
\end{proof}

On the other hand, we can show that the two manifolds are {\em intrinsically different}. The following lemma says that the total arc length of $\calM(R_m,m)$ is strictly higher than that of $\calM(R_n,n)$.

\begin{lemma}
We have $\vol(\calM(R_m,m)) > \vol(\calM(R_n,n))$.
\end{lemma}
\begin{proof}
The total arc length of $\calM(R_m,m)$ can be computed directly:
\[
\vol(\calM(R_m,m)) = \int_0^{2\pi} \| F'_{R_m,m}(\theta) \| {\rm d} \theta  = \sqrt{m^2+1} \cdot 2 \pi R_m.
\]
Similarly, we have the analogous formula for $\vol(\calM(R_n,n))$. Using the definitions of $R_m$ and $R_n$, we find that it suffices to show
\begin{equation}\label{eq: psi-n-xi-m2-1-psi-m-xi-n2-1-gtr-0}
\psi_n(\xi) \cdot (m^2+1) - \psi_m(\xi) \cdot (n^2+1) > 0.
\end{equation}
Applying the exact bounds for the cosine function from Lemma~\ref{lem: exact-bounds-for-cosine-720-900}, we find
\begin{equation}\label{en: LHS-eq-psi-ge-1-12-1-360-xi-6}
{\rm LHS}_{\eqref{eq: psi-n-xi-m2-1-psi-m-xi-n2-1-gtr-0}}
\ge
\frac{1}{12} \left( (n^2+1)m^4 - (n^4+1)m^2 - (n^4 - n^2) \right) \xi^4 - \frac{n^2+1}{360} m^6 \xi^6.
\end{equation}
Since $0 < \xi < 1/m^2$ and $m \ge n+1$, we obtain
\begin{align*}
{\rm RHS}_{\eqref{en: LHS-eq-psi-ge-1-12-1-360-xi-6}}
& \ge \frac{\xi^6}{360} \! \left( 30(n^2+1)(n+1)^2 m^6 - 30(n^4+1) m^6 - (n^2+1) m^6 - (n^4 - n^2) m^4 \right) \\
&\ge \frac{\xi^6}{360} \! \left( 59 n^3 m^6 - n^4 m^4 \right) = \frac{\xi^6}{360} n^3 m^4 (59m^2 -n) > 0.
\end{align*}
We have finished the proof.
\end{proof}

\bigskip

\bigskip

\newpage

\section{Main parameters and events}
\label{sec: CS}

\noindent
\underline{Parameter $\varsigma$ \eqref{eq: varsigma}:}
\begin{align*}
    0 < \varsigma < \frac{1}{4}.
\end{align*}

\noindent
\underline{Parameter $\varepsilon$ \eqref{eq: CondVarepsilon}:}
\begin{align*}
\rv ( \varepsilon / 6 ) \ge 2n^{-\varsigma}
\qquad \mbox{and} \qquad
\varepsilon \ge \frac{1}{\sqrt{\| \rK \|_\infty} L_\rp} n^{-1/2+ \varsigma},
\end{align*}
where
$\|\rK\|_{\infty} := \max_{p,q \in M} \rK(p,q)$. When $n$ is large enough,
$$
    \varepsilon = O(n^{-\varsigma/d})\,.
$$

\smallskip

\noindent
\underline{Constants \eqref{ineq:Cond-c1}, \eqref{ineq:Cond-C}, and \eqref{ineq:Cond-c3}:}
\begin{align*}
   \cdr =& 2^{10}\,, &
    c_1 = & \frac{1}{800} \cdot \ell_\rp^2 \cdot \mu_{\min}(r_M/4)\,, &
    C_2 = &4 \cdot \frac{\|\rK\|_{\infty}^{1/4} C_{\rm gap}^{1/2} d^{1/2} L_{\rp}}{\ell_\rp^{1/2} c_1^{1/2}},\,& \mbox{ and } \qquad
    c_3 =& \frac{\ell_\rp}{\cdr \sqrt{d}}.
\end{align*}

\noindent
\underline{Parameters $\eta$, $\delta$, and $r$ \eqref{eq: CondEta}, \eqref{eq: CondDeltaR}:}
\begin{align*}
\eta &= \max\Big\{ C_2 \cdot \varepsilon^{1/2}, \frac{L_\rp^2}{c_3 \ell_\rp \sqrt{d}} \cdot \varepsilon \Big\}\,, &
\delta &= \cdr \sqrt{d} \cdot \eta &
\mbox{and} \qquad
r := & \cdr d^2 \cdot \delta.
\end{align*}
Asymptotically,
$$
    \eta = O(n^{-\varsigma/2d})\,.
$$

\noindent
\underline{Feasibility Assumption \eqref{eq: parameterFeasible}:} We require that $n$ should be large enough so that
\begin{align*}
r_M \ge 2^4 \cdr d^2 r
\qquad \mbox{and} \qquad
n \ge 100.
\end{align*}

\noindent
\underline{Common Neighbor Event $\Ecn$ \eqref{eq: Ecommonneighbor}:}
For any $W \subseteq {\bf V}$ with $|W|=n$,
\begin{equation*}
    \Ecn(W) := \left\{\forall \{i,j\} \in {W \choose 2},\, \Big| |N_W(i) \cap N_W(j)|/n - \rK(X_i,X_j) \Big| \le n^{-1/2+\varsigma}\right\},
\end{equation*}

\noindent
\underline{Net event $\Enet$ \eqref{eq:Enet}:}
For any $W \subseteq {\bf V}$ with $|W|=n$,
\begin{equation*}
\Enet(W) :=
\bigg\{
\forall p \in M, \, \big| \left\{ i \in W \, : \, X_i \in B(p, \varepsilon) \right\} \big| \ge \rv(2\varepsilon/3) \cdot \frac{n}{2}
\bigg\}\,.
\end{equation*}

\noindent
\underline{Cluster event $\Eclu$ \eqref{defi clu}:}
For a finite index set $A$ and a collection of pairs $\{(V_\alpha, i_\alpha)\}_{\alpha \in A}$ with $i_\alpha \in V_\alpha \subseteq {\bf V}$,
    \begin{align*}
    \Eclu(\{(V_\alpha, i_\alpha)\}_{\alpha \in A}) := \big\{ \text{for each } \alpha \in A, \, |V_\alpha| \ge n^{1-\varsigma} \mbox{ and for each } j \in V_\alpha \, |X_j - X_{i_\alpha}| < \eta.
    \big\}.
    \end{align*}

\noindent
\underline{Navigation Event $\Enavi$ \eqref{eq: E-navi}:}
    Suppose that $A$ is an index set. For a collection $\{V_\alpha\}_{\alpha \in A}$ of nonempty subsets of ${\bf V}$ and for a subset $W \subseteq {\bf V}$,
    \begin{equation*}
    \Enavi(\{V_\alpha\}_{\alpha \in A}, W) := \bigg\{
    \forall i \in W,
    \forall \alpha \in A,
    \bigg| \frac{|N(i) \cap V_\alpha|}{ |V_\alpha|} -
    \sum_{j\in V_\alpha} \frac{\rp(\|X_i-X_j\|)}{|V_\alpha|}
    \bigg| \le n^{-1/2 + \varsigma}
    \bigg\}.
    \end{equation*}

\noindent
\underline{Almost Orthogonal Event \eqref{eq: ao-event}} For a non-negative inteer $k$ and $i_0,\dots, i_k \in {\bf V}$,
\begin{align*}
    {\cal E}_{\rm ao}(i_0, \ldots, i_k)
= &
    \bigg\{
        \forall \alpha \in [k]\,, \Big| \|X_{i_\alpha} - X_{i_0}\| - r \Big| \le \delta \quad \mbox{and}  \quad
    \forall \{\alpha, \beta\} \in {[k] \choose 2},\,
        \Big| \|X_{i_\alpha}-X_{i_\beta}\| - \sqrt{2}r \Big| \le \delta
    \bigg\}\,.
\end{align*}

\noindent
\underline{Repulsion Event \eqref{eq: eventRps}}: For a finite set of vertices $\{u_\alpha\}_{\alpha \in [\ell]}$:
\begin{align*}
    {\cal E}_{\rm rps}(\{u_\alpha\}_{\alpha \in [\ell]}) := \bigg\{
    \forall \{\alpha, \beta\} \in \binom{[\ell]}{2}, \, \|X_{u_\alpha} - X_{u_\beta}\| \ge 0.3 \delta
    \bigg\}.
\end{align*}

\end{document}